# Nonconvex Statistical Optimization: Minimax-Optimal Sparse PCA in Polynomial Time

Zhaoran Wang* and Huanran Lu† and Han Liu‡

**Abstract**

Sparse principal component analysis (PCA) involves nonconvex optimization for which the global solution is hard to obtain. To address this issue, one popular approach is convex relaxation. However, such an approach may produce suboptimal estimators due to the relaxation effect. To optimally estimate sparse principal subspaces, we propose a two-stage computational framework named "tighten after relax": Within the "relax" stage, we approximately solve a convex relaxation of sparse PCA with early stopping to obtain a desired initial estimator; For the "tighten" stage, we propose a novel algorithm called sparse orthogonal iteration pursuit (SOAP), which iteratively refines the initial estimator by directly solving the underlying nonconvex problem. A key concept of this two-stage framework is the basin of attraction. It represents a local region within which the "tighten" stage has desired computational and statistical guarantees. We prove that, the initial estimator obtained from the "relax" stage falls into such a region, and hence SOAP geometrically converges to a principal subspace estimator which is minimax-optimal within a certain model class. Unlike most existing sparse PCA estimators, our approach applies to the non-spiked covariance models, and adapts to non-Gaussianity as well as dependent data settings. Moreover, through analyzing the computational complexity of the two stages, we illustrate an interesting phenomenon: Larger sample size can reduce the total iteration complexity. Our framework motivates a general paradigm for solving many complex statistical problems which involve nonconvex optimization with provable guarantees.

## 1 Introduction

Let $\mathbf{x}_1, \ldots, \mathbf{x}_n$ be $n$ observations of a random vector $\boldsymbol{X} \in \mathbb{R}^d$ with population covariance matrix $\boldsymbol{\Sigma} \in \mathbb{R}^{d \times d}$. Principal component analysis (PCA) estimates the top $k$ leading eigenvectors $\mathbf{u}_1^*, \ldots, \mathbf{u}_k^*$ of $\mathbf{\Sigma}$. In high dimensional regimes where $d \gg n$, Johnstone and Lu (2009); Paul (2007); Nadler (2008) show that the classical PCA may be inconsistent in estimating $\mathbf{u}_1^*$. To handle this problem, a popular approach is to assume that $\mathbf{u}_1^*$ is sparse, i.e., the number of nonzero entries of $\mathbf{u}_1^*$, denoted by $s^*$, is much smaller than $n$. Under such a sparsity assumption, significant progress has been made on the methodological development (Jolliffe et al., 2003; d'Aspremont et al., 2007; Zou et al., 2006; Shen and Huang, 2008; d'Aspremont et al., 2008; Witten et al., 2009; Journée et al., 2010; Yuan and Zhang, 2013; Ma, 2013; Vu et al., 2013) as well as theoretical understanding of sparse PCA (Johnstone and Lu, 2009; Paul and Johnstone, 2012; Nadler, 2008; Amini and Wainwright, 2009; Vu and Lei, 2012; Shen et al., 2013; Birnbaum et al., 2013; Vu and Lei, 2013; Cai et al., 2013; Berthet and Rigollet, 2013b,a; Lounici, 2013; Krauthgamer et al., 2013; Lei and Vu, 2014; Yang et al., 2014).

*Department of Operations Research and Financial Engineering, Princeton University, Princeton, NJ 08544, USA; e-mail: zhaoran@princeton.edu.

†Department of Operations Research and Financial Engineering, Princeton University, Princeton, NJ 08544, USA; e-mail: huanranl@princeton.edu.

‡Department of Operations Research and Financial Engineering, Princeton University, Princeton, NJ 08544, USA; e-mail: hanliu@princeton.edu.

However, there remains a significant gap between the computational method and statistical theory of sparse PCA: No tractable algorithm is available to achieve the minimax-optimal principal component estimator for general covariance structures. This gap arises from the nonconvex optimization formulation of sparse PCA. In particular, the estimator of the first leading eigenvector $\mathbf{u}_1^*$ of $\mathbf{\Sigma}$ can be formulated as

$$\widehat{\mathbf{u}}_1 = \operatorname*{argmin}_{\|\mathbf{v}\|_2=1} -\mathbf{v}^T\widehat{\mathbf{\Sigma}}\mathbf{v}, \quad \text{subject to } \|\mathbf{v}\|_0 \le s^*, \tag{1.1}$$

where $\widehat{\mathbf{\Sigma}}$ is the sample covariance estimator, $\|\cdot\|_2$ is the Euclidean norm, $\|\cdot\|_0$ denotes the number of nonzero coordinates, and $s^*$ is the sparsity level of $\mathbf{u}_1^*$. Though this estimator attains the optimal statistical rate of convergence (Vu and Lei, 2012, 2013), it is NP-hard to compute (1.1) since it requires minimizing a concave function over a cardinality constraint (Moghaddam et al., 2006a). It is even more challenging to estimate the top $k$ leading eigenvectors because of the extra orthogonality constraint on $\widehat{\mathbf{u}}_1, \ldots, \widehat{\mathbf{u}}_2$.

To address such a computational issue, d'Aspremont et al. (2007) propose a convex relaxation approach named DSPCA for estimating $\mathbf{u}_1^*$. Vu et al. (2013) further generalize DSPCA to estimate the principal subspace spanned by the top $k$ leading eigenvectors. Nevertheless, Vu et al. (2013) show that the obtained estimator only attains the suboptimal $s^*\sqrt{\log d/n}$ statistical rate of convergence. Meanwhile, several methods have been proposed to directly solve the underlying nonconvex problem, including variants of power methods and iterative thresholding methods (Journée et al., 2010; Yuan and Zhang, 2013; Ma, 2013), greedy method (d'Aspremont et al., 2008), as well as regression-type methods (Jolliffe et al., 2003; Zou et al., 2006; Shen and Huang, 2008; Cai et al., 2013). However, most of these methods are lack of statistical guarantees. There exist several exceptions:

- Yuan and Zhang (2013) propose the truncated power method. To estimate $\mathbf{u}_1^*$, their estimator attains the optimal $\sqrt{s^* \log d/n}$ rate. Nevertheless, this result requires the assumption that the initial estimator $\mathbf{u}_1^{(0)}$ satisfies $\big|\sin\angle(\mathbf{u}_1^{(0)}, \mathbf{u}_1^*)\big| \le 1 - C$, where $C \in (0,1)$ is a constant, which is hard to be satisfied when $d \to \infty$ (Ball, 1997). In addition, their method only estimates $\mathbf{u}_1^*$, and employs the deflation method (Mackey, 2009) to estimate $\mathbf{u}_2^*, \ldots, \mathbf{u}_k^*$, which leads to identifiability and orthogonality issues when the top $k$ eigenvalues of $\mathbf{\Sigma}$ are not distinct.

- Ma (2013) proposes an iterative thresholding method, which achieves a near optimal statistical rate of convergence when estimating several individual leading eigenvectors. Cai et al. (2013) propose a regression-type method that attains an optimal principal subspace estimator. These two methods rely on the rather restrictive spiked covariance assumption and require the data to be Gaussian (sub-Gaussian not included).

To close the gap between the computational and statistical aspects of sparse PCA, we propose a two-stage framework named "tighten after relax" for estimating the $k$-dimensional principal subspace $\mathcal{U}^*$ spanned by the top $k$ leading eigenvectors $\mathbf{u}_1^*, \ldots, \mathbf{u}_k^*$. The details are as follows:

- Within the "relax" stage, we approximately solve a convex relaxation of sparse PCA (see §3.2 for details) with the alternating direction method of multipliers (ADMM) (Boyd et al., 2011). Unlike d'Aspremont et al. (2007); Vu et al. (2013), which aim to compute exact minimizers of the convex relaxation, we early stop ADMM as soon as the iterative sequence enters the basin of attraction for the "tighten" stage, i.e., the region where the "tighten" stage enjoys desired computational and statistical guarantees.

- For the "tighten" stage, we propose a novel algorithm called sparse orthogonal iteration pursuit (SOAP) to refine the initial estimator obtained from the "relax" stage. This algorithm extends the orthogonal iteration algorithm (Golub and Van Loan, 2012) with a sparsification step. As will be shown in §3.1, it can also be viewed as a generalization of projected gradient descent applied on the nonconvex optimization formulation of sparse PCA.

This two-stage strategy is illustrated in Figure 1. Under a unified analytic framework, we provide simultaneous statistical and computational guarantees for the two stages. Suppose that the sample size $n$ is sufficiently large and the eigengap between the $k$-th and $(k+1)$-th largest eigenvalues of $\mathbf{\Sigma}$ is nonzero, we prove that:

- The final principal subspace estimator $\widehat{\mathcal{U}}$ obtained from our two-stage framework attains the minimax-optimal $\sqrt{s^* \log d/n}$ statistical rate of convergence.

- At the "relax" stage, the iterative sequence of principal subspace estimators $\{\mathcal{U}^{(t)}\}_{t=1}^{T}$ (at the $T$-th iteration we early stop the "relax" stage) satisfies

$$D(\mathcal{U}^*, \mathcal{U}^{(t)}) \leq \underbrace{\delta_1(\mathbf{\Sigma}) \cdot s^* \sqrt{\frac{\log d}{n}}}_{\text{Statistical Error}} + \underbrace{\delta_2(k, d, n) \cdot \frac{1}{\sqrt{t}}}_{\text{Optimization Error}}, \quad \text{for } t = 1, \ldots, T \tag{1.2}$$

  with high probability, where $D(\cdot, \cdot)$ is the subspace distance and $s^*$ is the sparsity level of $\mathcal{U}^*$, both of which will be rigorously defined in §2. Here $\delta_1(\mathbf{\Sigma})$ is a quantity which depends on the population covariance matrix $\mathbf{\Sigma}$, while $\delta_2(k, d, n)$ depends on $k$, $d$ and $n$ (see §4 for details).

- At the "tighten" stage, the iterative sequence $\{\mathcal{U}^{(t)}\}_{t=T+1}^{T+\widetilde{T}}$ (where $\widetilde{T}$ denotes the total number of iterations of SOAP) satisfies

$$D(\mathcal{U}^*,\mathcal{U}^{(t)}) \le \underbrace{\delta_3(\mathbf{\Sigma},k)\cdot \overbrace{\sqrt{\frac{s^*\log d}{n}}}^{\text{Optimal Rate}}}_{\text{Statistical Error}} + \underbrace{\gamma(\mathbf{\Sigma})^{(t-T-1)/4}\cdot D(\mathcal{U}^*,\mathcal{U}^{(T+1)})}_{\text{Optimization Error}}, \quad \text{for} \quad t = T+1,\ldots,T+\widetilde{T} \tag{1.3}$$

with high probability, where $\delta_3(\mathbf{\Sigma},k)$ is a quantity that depends on $\mathbf{\Sigma}$ and $k$, and

$$\gamma(\mathbf{\Sigma}) = [3\lambda_{k+1}(\mathbf{\Sigma}) + \lambda_k(\mathbf{\Sigma})]/[\lambda_{k+1}(\mathbf{\Sigma}) + 3\lambda_k(\mathbf{\Sigma})] < 1. \tag{1.4}$$

Here $\lambda_k(\mathbf{\Sigma})$ and $\lambda_{k+1}(\mathbf{\Sigma})$ are the $k$-th and $(k+1)$-th eigenvalues of $\mathbf{\Sigma}$ (see §4 for details).

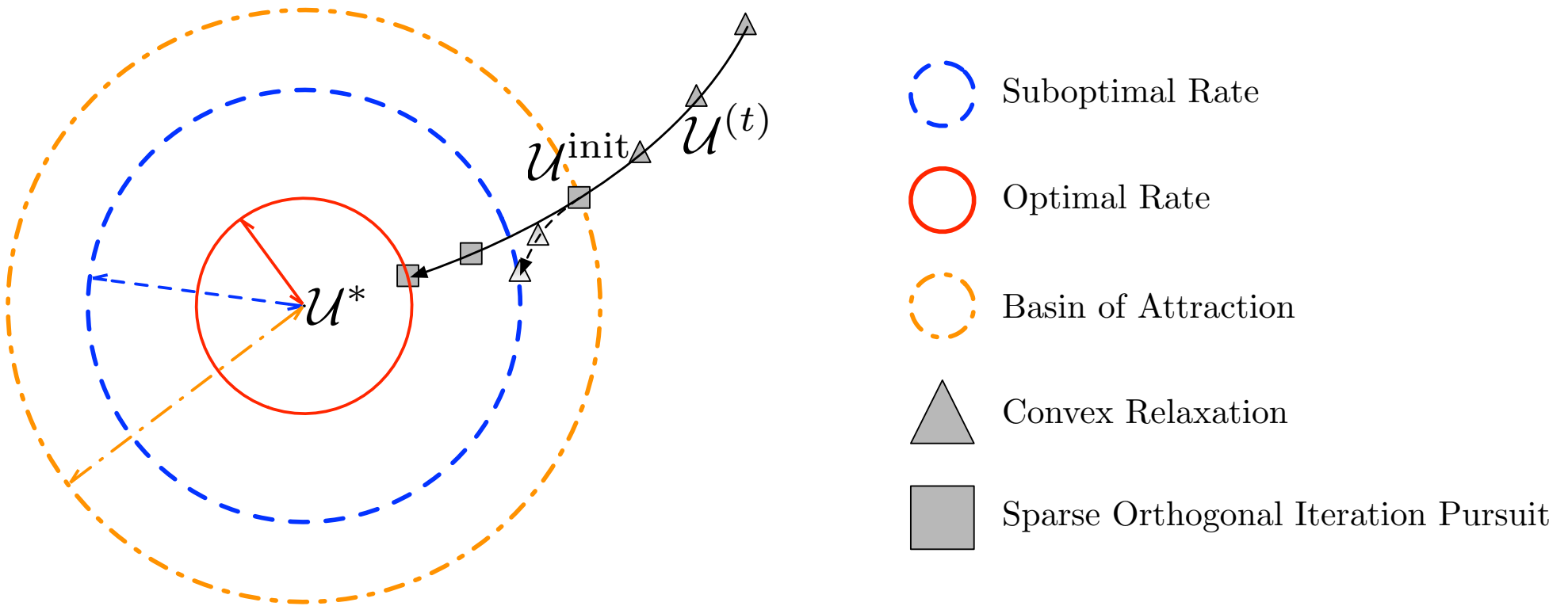

Figure 1: An illustration of the "tighten after relax" framework. Within the "relax" stage, we solve a convex relaxation of sparse PCA with ADMM. Correspondingly, the iterative sequence of principal subspace estimators is plotted with solid triangles. The "tighten" stage employs the proposed SOAP algorithm, and its iterative sequence is denoted by solid squares. Note that we early stop the "relax" stage as soon as the corresponding iterative sequence enters the basin of attraction for the "tighten" stage. The rationale is that, assuming we exactly solve the convex relaxation, the iterative sequence (hollow triangles) still converges to a suboptimal estimator (on the blue dash circle), while incurring extra computational cost (see §4 for details).

Our analysis shows that, within the "relax" stage, the optimization error term in (1.2) converges to zero at the rate of $1/\sqrt{t}$. However, the upper bound of the subspace estimation error $D(\mathcal{U}^*,\mathcal{U}^{(t)})$ in (1.2) can't be faster than the suboptimal $s^*\sqrt{\log d/n}$ rate of convergence, even if the "relax" stage takes an infinite number of iterations. This phenomenon, as illustrated in Figure 1, is because of the relaxation effect of the convex programming approach. Within the "tighten" stage, as the optimization error term in (1.3) decays to zero geometrically, the rate of convergence of $D(\mathcal{U}^*,\mathcal{U}^{(t)})$ improves to be $\sqrt{s^*\log d/n}$, which is optimal with respect to the sparsity level $s^*$, dimension $d$ and sample size $n$ (Vu and Lei, 2013). Compared with most existing works, our theory and method neither rely on the spiked covariance structure, nor the Gaussian assumption. Meanwhile, we need no assumption on the initial estimator. Furthermore, we provide generalizations to data distributions that are not sub-Gaussian, and data that are not independently distributed.

Moreover, in Theorem 4.2 we will show that, the basin of attraction for the "tighten" stage can be characterized as

$$\mathcal{U} : D(\mathcal{U}^*,\mathcal{U}) \le R = \min\left\{ \sqrt{\frac{k\gamma(\mathbf{\Sigma})\left[1-\gamma(\mathbf{\Sigma})^{1/2}\right]}{2}}, \frac{\sqrt{2\gamma(\mathbf{\Sigma})}}{4} \right\}, \tag{1.5}$$

where $R$ represents its radius and $\gamma(\mathbf{\Sigma})$ is defined in (1.4). Based upon this characterization, through analyzing the computational complexity of the two stages we illustrate an interesting phenomenon: The total number of operations needed to compute the optimal principal subspace estimator can be reduced if we have access to data with larger sample size.

This paper has three major contributions:

- We propose the first tractable procedure that provably attains the principal subspace estimator with optimal statistical rate of convergence with respect to the sparsity level $s^*$, dimension $d$ and sample size $n$, without assuming spiked covariance model or Gaussian data distribution.
- We propose a novel sparse PCA algorithm named sparse orthogonal iteration pursuit (SOAP), which converges to a minimax-optimal subspace estimator at a geometric rate. The computation within each iteration is highly efficient.
- We establish a unified analytic framework that characterizes both the statistical and computational properties of sparse PCA. Under this framework, we sharply characterize the basin of attraction phenomenon for the SOAP algorithm. We also develop novel variational inequality-based proof technique for analyzing the statistical performance of all the intermediate solutions obtained from ADMM (Boyd et al., 2011).

The rest of this paper is organized as follows. In §2 we briefly introduce the background. In §3 we present our two-stage framework. In §4 we lay out the main results. In §5 we provide the proof of the results in §4. In §6 we discuss the implications of the theoretical results. Numerical results are presented in §7. §8 concludes the paper with some more discussions.

**Notation:** Let $\mathbf{A} = [\mathbf{A}_{i,j}] \in \mathbb{R}^{d\times d}$ and $\mathbf{v} = (v_1, \ldots, v_d)^T \in \mathbb{R}^d$. The $\ell_q$ norm ($q \ge 1$) of vector $\mathbf{v}$ is defined as $\|\mathbf{v}\|_q = \big(\sum_{j=1}^d |v_j|^q\big)^{1/q}$. Specifically, $\|\mathbf{v}\|_0$ is defined as the number of nonzero entries of $\mathbf{v}$. We define $\lambda_i(\mathbf{A})$ and $\sigma_i(\mathbf{A})$ as the $i$-th largest eigenvalue and singular value of matrix $\mathbf{A}$. For $q \ge 1$, we denote $\|\mathbf{A}\|_q$ to be the operator norm, e.g., for $q = 2$, it satisfies $\|\mathbf{A}\|_2 = \sigma_1(\mathbf{A})$. The Frobenius norm of $\mathbf{A}$ is defined as $\|\mathbf{A}\|_{\mathrm{F}} = \big(\sum_{i=1}^d \sum_{j=1}^d |\mathbf{A}_{i,j}|^2\big)^{1/2}$. For matrices $\mathbf{A}_1, \mathbf{A}_2$, we define their inner product as $\langle \mathbf{A}_1, \mathbf{A}_2\rangle = \mathrm{Tr}(\mathbf{A}_1^T \mathbf{A}_2)$. For a set $\mathcal{S}$, $|\mathcal{S}|$ is its cardinality. The identity matrix in $\mathbb{R}^{d\times d}$ is denoted with $\mathbf{I}_d$. For index sets $\mathcal{I}, \mathcal{J} \subseteq \{1, \ldots, d\}$, we denote with $\mathbf{A}_{\mathcal{I},\mathcal{J}} \in \mathbb{R}^{d\times d}$ the corresponding matrix whose $(i,j)$-th entry equals $\mathbf{A}_{i,j}$ if $i \in \mathcal{I}$ and $j \in \mathcal{J}$, and zero otherwise. If $\mathcal{I} = \mathcal{J}$, we simplify the notation to be $\mathbf{A}_{\mathcal{I}}$. If $\mathcal{I}$ or $\mathcal{J}$ is $\{1, \ldots, d\}$, we replace it with a dot, e.g., $\mathbf{A}_{\mathcal{I},\cdot}$. Also, we denote by $\mathbf{A}_{i,*} \in \mathbb{R}^d$ the $i$-th row vector of $\mathbf{A}$. A matrix is orthonormal if its columns are unit length orthogonal vectors. The $(p, q)$-norm of a matrix, denoted as $\|\mathbf{A}\|_{p,q}$, is computed by first taking the $\ell_p$ norm of each row and then taking $\ell_q$ norm of these row norms. We denote diag($\mathbf{A}$) to be the vector consisting of the diagonal entries of $\mathbf{A}$. With a slight abuse of notation, we define diag($\mathbf{v}$) to be the diagonal matrix with $\mathbf{v} = (v_1, \ldots, v_d)^T$ on its diagonal. Throughout this paper, we employ generic numerical constants such as $C$ or $C'$, whose value may change from line to line.

## 2 Background

We consider the problem of estimating the leading $k$-dimensional principal subspace $\mathcal{U}^*$ spanned by the top $k$ eigenvectors $\mathbf{u}_1^*, \ldots, \mathbf{u}_k^*$ of the population covariance matrix $\mathbf{\Sigma}$. In the sequel, we first define the distance between subspaces and the notion of sparsity for subspace. We then overview the optimization formulation of sparse principal subspace estimation. More details of these concepts can be found in Vu and Lei (2013).

**Subspace Distance:** To evaluate the statistical accuracy of principal subspace estimators, we first need to define the distance between subspaces. Let $\mathcal{U}$ and $\mathcal{U}'$ be two $k$-dimensional subspaces of $\mathbb{R}^d$, We denote the projection matrices onto them by $\mathbf{\Pi}$ and $\mathbf{\Pi}'$ respectively, and the orthonormal bases that span them by $\mathbf{u}_1, \ldots, \mathbf{u}_k$ and $\mathbf{u}'_1, \ldots, \mathbf{u}'_k$. One definition of the distance between $\mathcal{U}$ and $\mathcal{U}'$ is

$$D(\mathcal{U}, \mathcal{U}') = \|\mathbf{\Pi} - \mathbf{\Pi}'\|_{\mathrm{F}}. \tag{2.1}$$

Let $\mathbf{U}^{\perp} = \big\{\mathbf{u}_{k+1}|\ldots|\mathbf{u}_d\big\}$ be an orthonormal matrix whose columns are orthogonal to the columns of $\mathbf{U}$. Its column space is hence the orthogonal subspace of $\mathcal{U}$, which is denoted as $\mathcal{U}^{\perp}$. Correspondingly, the projection matrix onto $\mathcal{U}^{\perp}$ is $\mathbf{\Pi}^{\perp} = \mathbf{I}_d - \mathbf{\Pi}$. The following lemma characterizes useful properties of the distance between subspaces.

**Lemma 2.1.** Let $\mathcal{U}$ and $\mathcal{U}'$ be two $k$-dimensional subspaces of $\mathbb{R}^d$. Using the above notation, we have

$$D(\mathcal{U}, \mathcal{U}') = \sqrt{2} \cdot \big\|\mathbf{U}^T \mathbf{U}'^{\perp}\big\|_{\mathrm{F}} = \sqrt{2} \cdot \big\|(\mathbf{U}^{\perp})^T \mathbf{U}'\big\|_{\mathrm{F}} = \sqrt{2} \cdot \big\|\mathbf{\Pi}^{\perp}\mathbf{\Pi}'\big\|_{\mathrm{F}} = \sqrt{2} \cdot \big\|\mathbf{\Pi}\mathbf{\Pi}'^{\perp}\big\|_{\mathrm{F}} \leq \sqrt{2k},$$

and

$$D(\mathcal{U}, \mathcal{U}') = \sqrt{2} \cdot \big\|\mathbf{U}^T \mathbf{U}'^{\perp}\big\|_{\mathrm{F}} = \sqrt{2} \cdot \Big[k - \big\|\mathbf{U}^T \mathbf{U}'\big\|_{\mathrm{F}}^2\Big]^{1/2} = \sqrt{2} \cdot \Big[k - 1/2 \cdot D(\mathcal{U}, \mathcal{U}'^{\perp})^2\Big]^{1/2}.$$

*Proof.* See Stewart and Sun (1990) and Bhatia (1997) for details. □

**Subspace Sparsity:** If the leading eigenvector $\mathbf{u}_1^*$ of $\mathbf{\Sigma}$ is of interest, then the sparsity level $s^*$ is defined to be the number of nonzero coefficients of $\mathbf{u}_1^*$. However, when the $k$-dimensional principal subspace $\mathcal{U}^*$ of $\mathbf{\Sigma}$ is of interest, the definition of sparsity level is less straightforward. In particular, when the top $k$ eigenvalues of $\mathbf{\Sigma}$ are not distinct, there exist multiple groups of leading eigenvectors that are equivalent up to rotations. Therefore, our notion of subspace sparsity must be invariant to rotation, i.e., the sparsity level of a subspace should not change if we choose two different bases of the subspace. Note that the (orthogonal) projection matrix onto subspace $\mathcal{U}^*$ is uniquely defined as $\mathbf{\Pi}^* = \mathbf{U}^*(\mathbf{U}^*)^T$ even though $\mathbf{U}^*$ is not unique. Hence, we define the sparsity level of $\mathcal{U}^*$ as the total number of nonzero coefficients on the diagonal of $\mathbf{\Pi}^*$, i.e., $s^* = \big|\mathrm{supp}[\mathrm{diag}(\mathbf{\Pi}^*)]\big|$. Such a notion of subspace sparsity can be interpreted in two ways:

- Since $\mathbf{\Pi}^*_{i,i} = \sum_{j=1}^{k} (\mathbf{U}^*_{i,j})^2$, if $\mathbf{\Pi}^*_{i,i} = 0$, then we have $\mathbf{U}^*_{i,j} = 0$ for any $j \in \{1, \ldots, k\}$. In other words, whichever basis $\{\mathbf{u}^*_1 | \ldots | \mathbf{u}^*_k\} = \mathbf{U}^*$ we choose, the $i$-th row of $\mathbf{U}^*$ is zero when $\mathbf{\Pi}^*_{i,i} = 0$. Therefore, the sparsity of $\mathcal{U}^*$ is equivalent to the row sparsity of the corresponding orthonormal matrix $\mathbf{U}^*$, whose columns form a basis of $\mathcal{U}^*$. Therefore we have

$$s^* = \big|\text{supp}[\text{diag}(\mathbf{\Pi}^*)]\big| = \|\mathbf{U}^*\|_{2,0}, \tag{2.2}$$

  where $\|\cdot\|_{2,0}$ is by definition the level of row-sparsity, i.e., the number of nonzero rows. The index set of the nonzero rows of $\mathbf{U}^*$ is named as its row-support, and is denoted by $\mathcal{S}^*$ hereafter. For $k = 1$, this definition of subspace sparsity reduces to the usual definition of vector sparsity, i.e., $s^* = \|\mathbf{u}^*_1\|_0$ and $\mathcal{S}^* = \text{supp}(\mathbf{u}^*_1)$.

- Since $\mathbf{\Pi}^*_{i,i} = 0$ implies $\mathbf{U}^*_{i,l} = 0$ for all $l \in \{1, \ldots, k\}$, we have

$$\mathbf{\Pi}^*_{i,j} = \mathbf{\Pi}^*_{j,i} = \sum_{l=1}^{k} \mathbf{U}^*_{i,l} \cdot \mathbf{U}^*_{j,l} = 0, \quad \text{for all } \ j \in \{1, \ldots, d\}. \tag{2.3}$$

  Consequently, $\mathbf{\Pi}^*_{i,i} = 0$ implies $\mathbf{\Pi}^* \mathbf{e}_i = \mathbf{0}$, where $\mathbf{e}_i$ is the vector that is all-zero except that its $i$-th coordinate is one. Since $\mathbf{\Pi}^*$ is the projection matrix of $\mathcal{U}^*$, it follows that $\mathcal{U}^*$ is orthogonal to $\mathbf{e}_i$. In other words, $\mathcal{U}^*$ is irrelevant to the $i$-th coordinate of $\mathbb{R}^d$.

Corresponding to this notion of subspace sparsity, our approach directly estimates the $k$-dimensional sparse principal subspace $\mathcal{U}^*$ of $\mathbf{\Sigma}$ rather than the individual leading eigenvectors. In this way, our approach easily handles the situation where at least two of the top $k$ leading eigenvalues of $\mathbf{\Sigma}$ are identical. In this situation, the individual leading eigenvectors corresponding to the same eigenvalue are not even identifiable. Also, previous methods which only recover the first leading eigenvector and rely on deflation method to recover the rest leading eigenvectors could yield inconsistent estimators.

**Subspace Estimation:** For the $k$-dimensional principal subspace $\mathcal{U}^*$ of $\mathbf{\Sigma}$ with sparsity level $s^*$, Vu and Lei (2013) consider the following estimator for the orthonormal matrix $\mathbf{U}^*$ consisting of the basis of $\mathcal{U}^*$,

$$\breve{\mathbf{U}} = \underset{\mathbf{U} \in \mathbb{R}^{d \times k}}{\text{argmin}} -\big\langle \widehat{\mathbf{\Sigma}}, \mathbf{U}\mathbf{U}^T \big\rangle, \quad \text{subject to } \mathbf{U} \text{ orthonormal, and } \|\mathbf{U}\|_{2,0} \leq s^*, \tag{2.4}$$

where $\widehat{\mathbf{\Sigma}}$ is an estimator of $\mathbf{\Sigma}$. Then the principal subspace estimator $\breve{\mathcal{U}}$ is the column space of $\breve{\mathbf{U}}$. Vu and Lei (2013) prove that, when $\widehat{\mathbf{\Sigma}}$ is the sample covariance estimator, i.e., $\widehat{\mathbf{\Sigma}} = 1/n \cdot \sum_{i=1}^{n} \mathbf{x}_i \mathbf{x}_i^T$, and $\mathbf{x}_1, \ldots, \mathbf{x}_n$ are independently drawn from a sub-Gaussian distribution, $\breve{\mathcal{U}}$ achieves the minimax-optimal statistical rate of convergence. However, the computation of this estimator is NP-hard even when $k = 1$ (Moghaddam et al., 2006b).

# 3 "Tighten after Relax" for Sparse PCA

In this section, we present a two-stage "tighten after relax" computational framework for sparse principal subspace estimation. We first present the "tighten" stage, for which we propose the SOAP algorithm. To obtain a desired initial estimator that falls within the basin of attraction for SOAP, we then introduce the "relax" stage, which approximately solves a convex relaxation of sparse PCA.

### 3.1 "Tighten" Stage: Sparse Orthogonal Iteration Pursuit (SOAP)

For the "tighten" stage, we propose the SOAP algorithm (Algorithm 1), which approximately solves the nonconvex optimization formulation of sparse PCA in (2.4). The $\mathsf{Truncate}(\cdot,\cdot)$ function in Algorithm 1 (Line 8) is described by Algorithm 2. In Lines 7 and 9 of Algorithm 1, $\mathsf{Thin_QR}(\cdot)$ denotes the thin QR decomposition (Golub and Van Loan, 2012). In detail, $\mathbf{V}^{(t+1)}, \mathbf{U}^{(t+1)} \in \mathbb{R}^{d\times k}$ are orthonormal matrices that satisfy

$$\mathbf{V}^{(t+1)} \cdot \mathbf{R}_1^{(t+1)} = \widetilde{\mathbf{V}}^{(t+1)}, \quad \mathbf{U}^{(t+1)} \cdot \mathbf{R}_2^{(t+1)} = \widetilde{\mathbf{U}}^{(t+1)},$$

where $\mathbf{R}_1^{(t+1)}, \mathbf{R}_2^{(t+1)} \in \mathbb{R}^{k\times k}$. This decomposition could be accomplished with $\mathcal{O}(d\cdot k^2)$ operations using the Householder algorithm (Householder, 1958). Here we remind that $k$ is the rank of the principal subspace of interest, which is much smaller than the dimension $d$.

---

**Algorithm 1** "Tighten" Stage: Approximately solving the nonconvex optimization problem in (2.4) using the proposed sparse orthogonal iteration pursuit (SOAP) algorithm. Here the initial estimator $\mathbf{U}^{\mathrm{init}}$ is obtained from the "relax" stage, and $T$ is the total number of iterations of the "relax" stage. To unify the later analysis, at the "tighten" stage, we set the iteration index $t$ to start from $T+1$.

---

1: **Function**: $\widehat{\mathbf{U}} \leftarrow \mathsf{SOAP}\big(\widehat{\mathbf{\Sigma}}, \mathbf{U}^{\mathrm{init}}\big)$
2: **Input:** Covariance Matrix Estimator $\widehat{\mathbf{\Sigma}}$, Initialization $\mathbf{U}^{\mathrm{init}}$
3: **Parameter:** Sparsity Parameter $\widehat{s}$, Maximum Number of Iterations $\widetilde{T}$
4: **Initialization:** $\widetilde{\mathbf{U}}^{(T+1)} \leftarrow \mathsf{Truncate}\big(\mathbf{U}^{\mathrm{init}}, \widehat{s}\big)$, $\mathbf{U}^{(T+1)}, \mathbf{R}_2^{(T+1)} \leftarrow \mathsf{Thin_QR}\big(\widetilde{\mathbf{U}}^{(T+1)}\big)$
5: **For** $t = T+1, \ldots, T+\widetilde{T}-1$
6: $\quad \widetilde{\mathbf{V}}^{(t+1)} \leftarrow \widehat{\mathbf{\Sigma}} \cdot \mathbf{U}^{(t)}$
7: $\quad \mathbf{V}^{(t+1)}, \mathbf{R}_1^{(t+1)} \leftarrow \mathsf{Thin_QR}\big(\widetilde{\mathbf{V}}^{(t+1)}\big)$
8: $\quad \widetilde{\mathbf{U}}^{(t+1)} \leftarrow \mathsf{Truncate}\big(\mathbf{V}^{(t+1)}, \widehat{s}\big)$
9: $\quad \mathbf{U}^{(t+1)}, \mathbf{R}_2^{(t+1)} \leftarrow \mathsf{Thin_QR}\big(\widetilde{\mathbf{U}}^{(t+1)}\big)$
10: **End For**
11: **Output:** $\widehat{\mathbf{U}} \leftarrow \mathbf{U}^{(T+\widetilde{T})}$

---

**Algorithm 2** "Tighten" Stage: The $\mathsf{Truncate}(\cdot,\cdot)$ function used in Line 8 of Algorithm 1. In Lines 2 and 4, $\mathbf{V}_{i,*}^{(t+1)}$ and $\widetilde{\mathbf{U}}_{i,*}^{(t+1)}$ denote the $i$-th row vectors of $\mathbf{V}^{(t+1)}$ and $\widetilde{\mathbf{U}}^{(t+1)}$.

---

1: **Function**: $\widetilde{\mathbf{U}}^{(t+1)} \leftarrow \mathsf{Truncate}\big(\mathbf{V}^{(t+1)}, \widehat{s}\big)$
2: **Row Sorting:** $\mathcal{I}_{\widehat{s}} \leftarrow$ The set of row index $i$'s corresponding to the top $\widehat{s}$ largest $\big\|\mathbf{V}_{i,*}^{(t+1)}\big\|_2$'s
3: **For** $i \in \{1, \ldots, d\}$
4: $\quad \widetilde{\mathbf{U}}_{i,*}^{(t+1)} \leftarrow \mathbb{1}\big[i \in \mathcal{I}_{\widehat{s}}\big] \cdot \mathbf{V}_{i,*}^{(t+1)}$
5: **End For**
6: **Output:** $\widetilde{\mathbf{U}}^{(t+1)}$

---

Algorithm 1 consists of two key steps:

- In Lines 6 and 7, we conduct a matrix multiplication, and then a renormalization with the QR decomposition. This step is also known as orthogonal iteration in numerical analysis literature. If the first leading eigenvector is of interest, i.e., $k = 1$, Lines 6 and 7 reduce to the classical power iteration.

  This step can also be viewed as a generalization of projected gradient descent applied on (2.4). Consider the minimization problem in (2.4) without the row-sparsity constraint, i.e.,

$$\underset{\mathbf{U}\in\mathbb{R}^{d\times k}}{\text{minimize}} -\big\langle \widehat{\boldsymbol{\Sigma}}, \mathbf{U}\mathbf{U}^T \big\rangle, \quad \text{subject to } \mathbf{U} \text{ orthonormal.}$$

  The gradient of the objective function is $-2\widehat{\boldsymbol{\Sigma}}\cdot\mathbf{U}$. Given $\mathbf{U}^{(t)}$, a gradient descent step produces $\mathbf{U}^{(t)}+\eta\cdot 2\widehat{\boldsymbol{\Sigma}}\cdot\mathbf{U}^{(t)}$, where $\eta > 0$ is the step size. To enforce the orthonormality constraint on $\mathbf{U}$, we renormalize $\mathbf{U}^{(t)}+\eta\cdot 2\widehat{\boldsymbol{\Sigma}}\cdot\mathbf{U}^{(t)}$ using the QR decomposition. Intuitively, this is analogous to the projection step in projected gradient descent for the settings where the constraint set is convex. We then have the following update scheme

$$\widetilde{\mathbf{V}}^{(t+1)} \leftarrow \mathcal{P}_{\text{orth}}\big(\mathbf{U}^{(t)}+\eta\cdot 2\widehat{\boldsymbol{\Sigma}}\cdot\mathbf{U}^{(t)}\big), \tag{3.1}$$

  where $\mathcal{P}_{\text{orth}}(\cdot)$ denotes the renormalization step. Moreover, for (3.1) Arora et al. (2013) prove that the optimal step size is $\eta = \infty$. In this case, we have

$$\mathcal{P}_{\text{orth}}\big(\mathbf{U}^{(t)}+\eta\cdot 2\widehat{\boldsymbol{\Sigma}}\cdot\mathbf{U}^{(t)}\big) = \mathcal{P}_{\text{orth}}\big(\eta\cdot 2\widehat{\boldsymbol{\Sigma}}\cdot\mathbf{U}^{(t)}\big) = \mathcal{P}_{\text{orth}}\big(\widehat{\boldsymbol{\Sigma}}\cdot\mathbf{U}^{(t)}\big),$$

  which implies that (3.1) is equivalent to Line 6 and Line 7.

- In Lines 8 and 9, we take a truncation step to enforce the row-sparsity constraint in (2.4). In detail, we greedily select the rows with the top $\widehat{s}$ largest row $\ell_2$ norms at each iteration. To enforce the orthonormality constraint in (2.4), we take another renormalization step (Line 9) afterwards. Note the QR decomposition in Line 9 produces an orthonormal and row-sparse $\mathbf{U}^{(t+1)}$. This is because the QR decomposition doesn't change the row-sparsity of a matrix.

By iteratively performing these two steps, we are approximately solving the nonconvex minimization problem in (2.4). Recall that solving (2.4) is NP-hard, i.e., it is intractable to compute the global minimum of (2.4). However, in §4 we will prove that, for $\widetilde{T}$ sufficiently large, the principal subspace estimator attained by SOAP well approximates the global minimum, in the sense that it enjoys the same optimal statistical rate of convergence as the global minimum.

It is worth noting that, when $k = 1$, i.e, the leading eigenvector is of interest, SOAP is equivalent to the truncated power method proposed in Yuan and Zhang (2013). However, when $k > 1$, SOAP directly estimates the principal subspace, while the truncated power method uses deflation method to estimate the rest leading eigenvectors, which raises identifiability, consistency and orthogonality issues in the presence of identical eigenvalues.

A single iteration of the SOAP algorithm (Line 5 to Line 10 in Algorithm 1) is highly efficient in terms of computational complexity. In particular, the truncation step (Line 8) takes $\mathcal{O}(k\cdot d\log d)$ operations for ranking the rows and $\mathcal{O}(k\cdot d)$ for setting rows to zero. If $\widehat{\boldsymbol{\Sigma}}$ is the sample covariance matrix, i.e., $\widehat{\boldsymbol{\Sigma}} = 1/n\cdot\sum_{i=1}^n \mathbf{x}_i\mathbf{x}_i^T$, Line 6 needs $\mathcal{O}(k\cdot d\cdot n)$ operations, since we can first compute the multiplication $\mathbf{x}_i^T \cdot \mathbf{U}^{(t)}$, and then the left multiplication with $\mathbf{x}_i$ and the summation. If $\widehat{\mathbf{\Sigma}}$ is some other covariance matrix estimator without such a structure, the matrix multiplication step in Line 6 requires $\mathcal{O}(k \cdot d^2)$ operations. Meanwhile, the QR decompositions in Lines 7 and 9 require $\mathcal{O}(d \cdot k^2)$ operations using the Householder algorithm. Note that we are interested in high dimensional regimes where $d \gg n$. Also, we have $k \leq s^* \ll n$, where the first inequality holds because if $k > s^*$, then $\mathbf{U}^*$ is not rank $k$, since it only has $s^*$ nonzero rows. Hence, each iteration requires $\mathcal{O}(k \cdot d^2)$ operations (or $\mathcal{O}(k \cdot d \cdot n)$ for $\widehat{\mathbf{\Sigma}}$ being the sample covariance matrix).

### 3.2 "Relax" Stage: Convex Relaxation

To obtain a desired initial estimator that falls within the basin of attraction for SOAP, we consider the following convex minimization problem

$$\text{minimize} \left\{ -\langle \widehat{\mathbf{\Sigma}}, \mathbf{\Pi} \rangle + \rho \|\mathbf{\Pi}\|_{1,1} \;\middle|\; \text{Tr}(\mathbf{\Pi}) = k, \; \mathbf{0} \preceq \mathbf{\Pi} \preceq \mathbf{I}_d \right\}, \tag{3.2}$$

which is a relaxation of the nonconvex minimization problem in (2.4) (d'Aspremont et al., 2007; Vu et al., 2013). The intuition behind this convex relaxation can be understood as follows:

- First we reparameterize $\mathbf{U}\mathbf{U}^T$ as $\mathbf{\Pi}$. Since $\mathbf{U}$ is orthonormal, $\mathbf{\Pi}$ is the projection matrix of a $k$-dimensional subspace of $\mathbb{R}^d$. As an (orthogonal) projection matrix, $\mathbf{\Pi}$ is symmetric and has $k$ eigenvalues of one and $d - k$ eigenvalues of zero. Such a constraint on $\mathbf{\Pi}$'s eigenvalues is relaxed into $\text{Tr}(\mathbf{\Pi}) = k$ and $0 \preceq \mathbf{\Pi} \preceq \mathbf{I}_d$. Note that the trace of a matrix is equal to the sum of its eigenvalues. This relaxed constraint states that the eigenvalues of $\mathbf{\Pi}$ should be between zero and one, while the sum of them should be equal to $k$.

- For the constraint of row-sparsity on $\mathbf{U}$ in (2.4), recall that we have $\|\mathbf{U}\|_{2,0} = \big|\text{supp}[\text{diag}(\mathbf{\Pi})]\big|$ by (2.2). By (2.3), for any $i \in \{1, \ldots, d\}$, if the diagonal entry $\mathbf{\Pi}_{i,i} = 0$, then all the entries of the same row and column are zero, i.e., $\mathbf{\Pi}_{i,j} = \mathbf{\Pi}_{j,i} = 0$ for all $j \in \{1, \ldots, d\}$. Thus, the total number of nonzero entires in $\mathbf{\Pi}$, denoted by $\|\mathbf{\Pi}\|_{0,0}$, satisfies

$$\|\mathbf{\Pi}\|_{0,0} \leq \big|\text{supp}[\text{diag}(\mathbf{\Pi})]\big|^2 = \|\mathbf{U}\|_{2,0}^2 \leq (s^*)^2.$$

  This constraint could be further relaxed into a constraint on $\|\mathbf{\Pi}\|_{1,1}$, i.e., the sum of absolute values of all $\mathbf{\Pi}$'s entries. Such a relaxation is analogous to replacing the $\ell_0$ constraint with $\ell_1$ constraint in sparse linear regression. Then by Lagrangian duality, we obtain (3.2).

For notational simplicity, we define the convex set

$$\mathcal{A} = \left\{ \mathbf{\Pi} \colon \; \mathbf{\Pi} \in \mathbb{R}^{d \times d}, \;\; \text{Tr}(\mathbf{\Pi}) = k, \; \mathbf{0} \preceq \mathbf{\Pi} \preceq \mathbf{I}_d \right\}. \tag{3.3}$$

Since (3.2) has both nonsmooth regularization term and nontrivial constraint set $\mathcal{A}$, it is difficult to directly apply gradient descent and its variants. Though projected subgradient descent is feasible for solving (3.2), it converges at the rate of $1/\sqrt{t}$ in terms of objective function value, which translates to a slow $1/t^{1/4}$ rate in terms of subspace distance by our analysis. To make it easier to develop the corresponding convex optimization algorithm, we consider an equivalent form of (3.2)

$$\text{minimize} \left\{ -\langle \widehat{\mathbf{\Sigma}}, \mathbf{\Pi} \rangle + \rho \|\mathbf{\Phi}\|_{1,1} \;\middle|\; \mathbf{\Pi} = \mathbf{\Phi}, \; \mathbf{\Pi} \in \mathcal{A}, \; \mathbf{\Phi} \in \mathcal{B} \right\}, \quad \text{where} \;\; \mathcal{B} = \mathbb{R}^{d \times d}. \tag{3.4}$$

The Lagrangian corresponding to (3.4) is

$$L(\mathbf{\Pi}, \mathbf{\Phi}, \mathbf{\Theta}) = -\big\langle \widehat{\mathbf{\Sigma}}, \mathbf{\Pi} \big\rangle + \rho \|\mathbf{\Phi}\|_{1,1} - \langle \mathbf{\Theta}, \mathbf{\Pi} - \mathbf{\Phi} \rangle, \quad \mathbf{\Pi} \in \mathcal{A},\ \mathbf{\Phi} \in \mathcal{B},\ \mathbf{\Theta} \in \mathbb{R}^{d\times d}, \tag{3.5}$$

where $\mathbf{\Theta} \in \mathbb{R}^{d\times d}$ is the Lagrange multiplier associated with the equality constraint $\mathbf{\Pi} = \mathbf{\Phi}$ in (3.4). We leverage the alternating direction method of multipliers (ADMM), which iteratively minimizes the augmented Lagrangian

$$L(\mathbf{\Pi}, \mathbf{\Phi}, \mathbf{\Theta}) + \beta/2 \cdot \|\mathbf{\Pi} - \mathbf{\Phi}\|_{\mathrm{F}}^2 \tag{3.6}$$

with respect to $\mathbf{\Pi}$ and $\mathbf{\Phi}$ separately, and then updates the dual variable $\mathbf{\Theta}$. Here $\beta > 0$ is a penalty parameter that enforces the equality constraint $\mathbf{\Pi} = \mathbf{\Phi}$. The detailed update scheme is described in Algorithm 3.

In particular, to obtain $\mathbf{\Pi}^{(t+1)}$ and $\mathbf{\Phi}^{(t+1)}$, we need to solve two subproblems (Lines 6 and 7 in Algorithm 3). The first can be solved by projecting $\mathbf{\Phi}^{(t)} - \mathbf{\Theta}^{(t)} + \widehat{\mathbf{\Sigma}}/\rho$ onto $\mathcal{A}$ using Algorithm 4, in which the quadratic program in Line 4 can easily be solved (see Lemma 4.1 of Vu et al. (2013) for details). The second can be solved using entry-wise soft-thresholding (Algorithm 5). We defer the derivation of Algorithms 4 and 5 to §A.

After $T$ iterations, we early stop the "relax" stage and calculate $\overline{\mathbf{\Pi}}^{(T)}$ (Line 10 in Algorithm 3). Then we set the columns of the initial estimator $\mathbf{U}^{\text{init}} \in \mathbb{R}^{d\times k}$ for the "tighten" stage to be the top $k$ leading eigenvectors of $\overline{\mathbf{\Pi}}^{(T)}$. In §4 we will prove that, for a sufficiently large $T$, the column space $\mathcal{U}^{\text{init}}$ of $\mathbf{U}^{\text{init}}$ falls within the basin of attraction defined in (1.5). Thus, initialized with $\mathbf{U}^{\text{init}}$, SOAP (Algorithm 1) provably converges to the optimal principal subspace estimator at a geometric rate.

---

**Algorithm 3** "Relax" Stage: Solving convex relaxation (3.2) using the alternating direction method of multipliers (ADMM) algorithm.

---

1: **Function**: $\mathbf{U}^{\text{init}} \leftarrow \mathsf{ADMM}\big(\widehat{\mathbf{\Sigma}}\big)$
2: **Input:** Covariance Matrix Estimator $\widehat{\mathbf{\Sigma}}$
3: **Parameter:** Regularization Parameter $\rho > 0$ in (3.2), Penalty Parameter $\beta > 0$ of the Augmented Lagrangian in (3.6), Maximum Number of Iterations $T$
4: $\mathbf{\Pi}^{(0)} \leftarrow \mathbf{0}$, $\mathbf{\Phi}^{(0)} \leftarrow \mathbf{0}$, $\mathbf{\Theta}^{(0)} \leftarrow \mathbf{0}$
5: **For** $t = 0, \ldots, T-1$
6: $\quad \mathbf{\Pi}^{(t+1)} \leftarrow \operatorname{argmin}\Big\{ L\big(\mathbf{\Pi}, \mathbf{\Phi}^{(t)}, \mathbf{\Theta}^{(t)}\big) + \beta/2 \cdot \big\|\mathbf{\Pi} - \mathbf{\Phi}^{(t)}\big\|_{\mathrm{F}}^2 \;\big|\; \mathbf{\Pi} \in \mathcal{A} \Big\}$ (Algorithm 4)
7: $\quad \mathbf{\Phi}^{(t+1)} \leftarrow \operatorname{argmin}\Big\{ L\big(\mathbf{\Pi}^{(t+1)}, \mathbf{\Phi}, \mathbf{\Theta}^{(t)}\big) + \beta/2 \cdot \big\|\mathbf{\Pi}^{(t+1)} - \mathbf{\Phi}\big\|_{\mathrm{F}}^2 \;\big|\; \mathbf{\Phi} \in \mathcal{B} \Big\}$ (Algorithm 5)
8: $\quad \mathbf{\Theta}^{(t+1)} \leftarrow \mathbf{\Theta}^{(t)} - \beta\big(\mathbf{\Pi}^{(t+1)} - \mathbf{\Phi}^{(t+1)}\big)$
9: **End For**
10: $\overline{\mathbf{\Pi}}^{(T)} \leftarrow 1/T \cdot \sum_{t=0}^{T} \mathbf{\Pi}^{(t)}$
11: Set the columns of $\mathbf{U}^{\text{init}} \in \mathbb{R}^{d\times k}$ to be the top $k$ leading eigenvectors of $\overline{\mathbf{\Pi}}^{(T)}$
12: **Output:** $\mathbf{U}^{\text{init}}$

---

The convex relaxation approach itself has two shortcomings:

- Statistically, the principal subspace estimator corresponding to the exact minimizer of (3.2) has a suboptimal statistical rate of convergence, which is reflected by the first term in (1.2).

**Algorithm 4** "Relax" Stage: Solving the subproblem in Line 6 of Algorithm 3.

**Function**: $\mathbf{\Pi}^{(t+1)} \leftarrow \mathsf{Projection}\big(\mathbf{\Phi}^{(t)}, \mathbf{\Theta}^{(t)}, \widehat{\mathbf{\Sigma}}, \beta\big)$

**Eigenvalue Decomposition:** $\mathbf{Q}\mathbf{\Lambda}^{(t)}\mathbf{Q}^T \leftarrow \mathbf{\Phi}^{(t)} + \mathbf{\Theta}^{(t)}/\beta + \widehat{\mathbf{\Sigma}}/\beta$

**Quadratic Programming:**

$$\mathbf{v}' \leftarrow \mathrm{argmin}\Big\{\big\|\mathbf{v} - \mathrm{diag}\big(\mathbf{\Lambda}^{(t)}\big)\big\|_2^2 \;\big|\; \mathbf{v} \in \mathbb{R}^d,\; \textstyle\sum_{j=1}^d v_j = k,\; v_j \in [0,1] \text{ for all } j\Big\}$$

**Output:** $\mathbf{\Pi}^{(t+1)} \leftarrow \mathbf{Q} \cdot \mathrm{diag}\{v_1', \ldots, v_d'\} \cdot \mathbf{Q}^T$

**Algorithm 5** "Relax" Stage: Solving the subproblem in Line 7 of Algorithm 3.

**Function**: $\mathbf{\Phi}^{(t+1)} \leftarrow \mathsf{Soft_Thresholding}\big(\mathbf{\Pi}^{(t+1)}, \mathbf{\Theta}^{(t)}, \rho, \beta\big)$

**For** $i, j \in \{1, \ldots, d\}$

$$\mathbf{\Phi}_{i,j}^{(t+1)} \leftarrow \begin{cases} 0 & \text{if } \big|\mathbf{\Pi}_{i,j}^{(t+1)} - \mathbf{\Theta}_{i,j}^{(t)}/\beta\big| \le \rho/\beta \\ \mathrm{sign}\Big(\mathbf{\Pi}_{i,j}^{(t+1)} - \mathbf{\Theta}_{i,j}^{(t)}/\beta\Big)\Big(\big|\mathbf{\Pi}_{i,j}^{(t+1)} - \mathbf{\Theta}_{i,j}^{(t)}/\beta\big| - \rho/\beta\Big) & \text{if } \big|\mathbf{\Pi}_{i,j}^{(t+1)} - \mathbf{\Theta}_{i,j}^{(t)}/\beta\big| > \rho/\beta \end{cases}$$

**End For**

**Output:** $\mathbf{\Phi}^{(t+1)}$

- The computational complexity at each iteration of Algorithm 3 is relatively high. The bottleneck is the eigenvalue decomposition used in Algorithm 4, which takes $\mathcal{O}(d^3)$ operations in practice. In high dimensional settings, a single iteration of ADMM is more time consuming than an iteration of SOAP. Also, in terms of optimization error, the $1/\sqrt{t}$ rate of convergence in (1.2) is significantly slower compared with the geometric rate of SOAP.

Our two-stage approach avoids these shortcomings. In particular, we don't aim to obtain the exact minimizer to (3.2). Instead, we early stop the "relax" stage (Algorithm 3) after $T$ iterations (In §4 and §6, we will discuss the proper choice of $T$ both in theory and practice.). In other words, we only need an intermediate solution on the solution path of ADMM to be the initial estimator for the "tighten" stage. Thus, the total number iterations of the "relax" stage is relatively small, which doesn't incur too much computational cost.

## 4 Theoretical Results

We provide theoretical justification for the proposed two-stage method. To describe our results, we define the model class $\mathcal{M}_d(\mathbf{\Sigma}, k, s^*)$ as follows,

$$\mathcal{M}_d(\mathbf{\Sigma}, k, s^*) : \begin{cases} \boldsymbol{X} = \mathbf{\Sigma}^{1/2}\boldsymbol{Z}, \text{ where } \boldsymbol{Z} \in \mathbb{R}^d \text{ is sub-Gaussian with mean zero,} \\ \qquad \text{variance proxy less than one, and covariance matrix } \mathbf{I}_d; \\ \text{The } k\text{-dimensional principal subspace } \mathcal{U}^* \text{ of } \mathbf{\Sigma} \text{ is } s^*\text{-sparse as defined in §2;} \\ \lambda_k(\mathbf{\Sigma}) - \lambda_{k+1}(\mathbf{\Sigma}) > 0. \end{cases}$$

Here $\mathbf{\Sigma}^{1/2}$ satisfies $\mathbf{\Sigma}^{1/2} \cdot \mathbf{\Sigma}^{1/2} = \mathbf{\Sigma}$. Remind that the sparsity of $\mathcal{U}^*$ is defined in (2.2) and $\lambda_j(\mathbf{\Sigma})$ is the $j$-th leading eigenvalue of $\mathbf{\Sigma}$. For notational simplicity, hereafter we abbreviate $\lambda_j(\mathbf{\Sigma})$ to be $\lambda_j$. Unless otherwise stated, our discussion is within this model class. We will extend it to more general settings in §4.5.

It is worth noting that this model class doesn't restrict the population covariance matrix $\mathbf{\Sigma}$ to be spiked covariance matrix, which takes the form

$$\mathbf{\Sigma} = (\lambda_1 - \lambda_{k+1})\mathbf{u}_1^*(\mathbf{u}_1^*)^T + (\lambda_2 - \lambda_{k+1})\mathbf{u}_2^*(\mathbf{u}_2^*)^T + \cdots + (\lambda_k - \lambda_{k+1})\mathbf{u}_k^*(\mathbf{u}_k^*)^T + \lambda_{k+1}\mathbf{I}_d.$$

The $k+1, \ldots, d$-th eigenvalues of a spiked covariance matrix $\mathbf{\Sigma}$ must be identical, which is rather restrictive. Furthermore, our theory doesn't require $\boldsymbol{X}$ to be Gaussian, as required in several previous works, e.g., Ma (2013); Cai et al. (2013).

**Notation:** Before we present the main theoretical results, we introduce some notation. Recall that $D(\cdot,\cdot)$ is the subspace distance defined in (2.1). Meanwhile, $\gamma(\mathbf{\Sigma})$ is defined in (1.4), which is abbreviated as $\gamma$ hereafter, i.e., $\gamma = (3\lambda_{k+1} + \lambda_k)/(\lambda_{k+1} + 3\lambda_k) < 1$. We define

$$n_{\min} = C \cdot (s^*)^2 \log d \cdot \left(\frac{\lambda_1}{\lambda_k - \lambda_{k+1}}\right)^2 \Bigg/ \min\left\{\sqrt{\frac{k\cdot\gamma(1-\gamma^{1/2})}{2}}, \frac{\sqrt{2\gamma}}{4}\right\}^2 \tag{4.1}$$

to be the required minimum number of data points. For notational simplicity, we define several key quantities. First, we define

$$\zeta_1 = \frac{C\lambda_1}{\lambda_k - \lambda_{k+1}} \cdot s^* \sqrt{\frac{\log d}{n}}, \qquad \zeta_2 = \frac{C'\sqrt{\lambda_1}}{\sqrt{\lambda_k - \lambda_{k+1}}} \cdot \left(\frac{k \cdot d^2 \log d}{n}\right)^{1/4}, \tag{4.2}$$

which will be used in the analysis of the "relax" stage, and

$$\xi_1 = C\sqrt{k} \cdot \left(\frac{\lambda_k}{\lambda_k - \lambda_{k+1}}\right)^2 \cdot \frac{\sqrt{\lambda_1\lambda_{k+1}}}{\lambda_k - \lambda_{k+1}} \cdot \sqrt{\frac{s^* \cdot (k + \log d)}{n}}, \tag{4.3}$$

which will be used in the analysis of the "tighten" stage. Recall that, as defined in (1.5), the radius of the basin of attraction for SOAP is

$$R = \min\left\{\sqrt{\frac{k\gamma(1-\gamma^{1/2})}{2}}, \frac{\sqrt{2\gamma}}{4}\right\}.$$

The required minimum iteration number of the "relax" stage (Algorithm 3) and the "tighten" stage (Algorithm 1) are denoted as

$$T_{\min} = \left\lceil \frac{\zeta_2^2}{(R - \zeta_1)^2} \right\rceil, \qquad \widetilde{T}_{\min} = \left\lceil \frac{4\log(R/\xi_1)}{\log(1/\gamma)} \right\rceil \tag{4.4}$$

respectively. We will provide more intuition behind these quantities later in this section.

## 4.1 Main Result

Let $\mathcal{U}^{(t)}$ be the column space of $\mathbf{U}^{(t)}$ in Algorithm 1. Using the notation in (4.1)-(4.4), our main result is as follows.

**Theorem 4.1.** Let $\mathbf{x}_1, \ldots, \mathbf{x}_n$ be $n$ independent realizations of $\boldsymbol{X} \in \mathcal{M}_d(\boldsymbol{\Sigma}, k, s^*)$ with $n > n_{\min}$, and $\widehat{\boldsymbol{\Sigma}}$ be the sample covariance matrix. We set the regularization parameter $\rho$ in (3.2) to be $\rho = C\lambda_1\sqrt{\log d/n}$ for a sufficiently large $C > 0$ and the penalty parameter $\beta$ of ADMM (Line 3 of Algorithm 3) to be $\beta = d \cdot \rho/\sqrt{k}$. Suppose that the sparsity parameter $\widehat{s}$ in Algorithm 1 (Line 3) is chosen such that

$$\widehat{s} = C' \max\left\{\left\lceil \frac{4k}{(\gamma^{-1/2}-1)^2} \right\rceil, 1\right\} \cdot s^*$$

for some integer constant $C' \geq 1$. Taking $T \geq T_{\min}$ iterations of Algorithm 3 and then $\widetilde{T} \geq \widetilde{T}_{\min}$ iterations of Algorithm 1, we obtain the final estimator $\widehat{\mathcal{U}} = \mathcal{U}^{(T+\widetilde{T})}$, which satisfies

$$D\big(\mathcal{U}^*, \widehat{\mathcal{U}}\big) \leq C'' \cdot \xi_1 = C'''\sqrt{k} \cdot \left(\frac{\lambda_k}{\lambda_k - \lambda_{k+1}}\right)^2 \cdot \frac{\sqrt{\lambda_1\lambda_{k+1}}}{\lambda_k - \lambda_{k+1}} \cdot \sqrt{\frac{s^* \cdot (k + \log d)}{n}} \tag{4.5}$$

with high probability.

*Proof.* See §5 for a detailed proof. □

In the following we show that $\widehat{\mathcal{U}}$ attains the minimax lower bound and is thus rate optimal.

**Minimax-Optimality:** To show the optimality of the upper bound in Theorem 4.1, we focus on a smaller model class $\widetilde{\mathcal{M}}_d(\boldsymbol{\Sigma}, k, s^*, \kappa)$, which is the same as $\mathcal{M}_d(\boldsymbol{\Sigma}, k, s^*)$ except that the eigengap of $\boldsymbol{\Sigma}$ satisfies

$$\lambda_k - \lambda_{k+1} > \kappa\lambda_k$$

for some constant $\kappa > 0$. This condition is mild compared to previous works, e.g., Ma (2013) assumes the condition that $\lambda_k - \lambda_{k+1} \geq \kappa\lambda_1$, which is more restrictive than ours because $\lambda_1 \geq \lambda_k$. Within this model class, we assume that the dimension $k$ of the principal subspace is fixed. This assumption has been justified in real-world applications such as population genetics (Engelhardt and Stephens, 2010), in which the dimension $k$ of the principal subspace represents the number of population groups, which doesn't increase as the sparsity level $s^*$, dimension $d$ and sample size $n$ are growing. In other words, $s^*$, $d$ and $n$ are of our primary concern.

Theorem 3.1 of Vu and Lei (2013) implies the following minimax lower bound

$$\inf_{\widetilde{\mathcal{U}}} \sup_{\boldsymbol{X} \in \widetilde{\mathcal{M}}_d(\boldsymbol{\Sigma}, k, s^*, \kappa)} \mathbb{E}\left[D\big(\widetilde{\mathcal{U}}, \mathcal{U}^*\big)^2\right] \geq C \cdot \frac{\lambda_1\lambda_{k+1}}{(\lambda_k - \lambda_{k+1})^2} \cdot \frac{(s^* - k) \cdot \big\{k + \log[(d-k)/(s^*-k)]\big\}}{n}, \tag{4.6}$$

where $\widetilde{\mathcal{U}}$ denotes any principal subspace estimator. Assuming that $s^*$ is sufficiently large such that $s^* - k \geq s^*/2$, and $d$ is sufficiently large such that $\log(d-k) \geq 2\log(s^*-k), \log(d-k) \geq 1/2 \cdot \log d$ (to avoid trivial cases), the right-hand side of (4.6) can be lower bounded by

$$C' \cdot \frac{\lambda_1\lambda_{k+1}}{(\lambda_k - \lambda_{k+1})^2} \cdot \frac{s^* \cdot (k + 1/4 \cdot \log d)}{n}. \tag{4.7}$$

Note that by Lemma 2.1 we have $D(\mathcal{U}^*,\widehat{\mathcal{U}}) \le \sqrt{2k}$. For $s^*$, $d$ and $n$ sufficiently large, it is easy to derive the following upper bound in expectation from the upper bound in high probability in (4.5) of Theorem 4.1,

$$\mathbb{E}\left[D(\mathcal{U}^*,\widehat{\mathcal{U}})^2\right] \le \left(C'' \cdot k \cdot \kappa^{-4}\right) \cdot \frac{\lambda_1\lambda_{k+1}}{(\lambda_k - \lambda_{k+1})^2} \cdot \frac{s^* \cdot (k + \log d)}{n}, \tag{4.8}$$

which attains the minimax lower bound in (4.7) within $\widetilde{\mathcal{M}}_d(\mathbf{\Sigma}, k, s^*, \kappa)$ up to constants. Hence, we conclude that $\widehat{\mathcal{U}}$ is minimax-optimal within this model class. Within more general model classes where $k$ is also increasing, the upper bound in (4.8) is larger than the minimax lower bound in (4.7) by a factor of $k$. In §5 we will show that this is an artifact of our analysis, and this factor of $k$ will disappear if one manages to get a tighter bound in Lemma 5.5. Nevertheless, for most applications, the rate of convergence with respect to $s^*$, $d$ and $n$ is of primary concern, and our result gives the optimal $\sqrt{s^* \log d/n}$ rate.

In the following, we present the detailed computational and statistical characterization of our procedure. In particular, we establish upper bounds for the distances between the iterative sequence of subspace estimators and the true principal subspace $\mathcal{U}^*$. We will first analyze the "tighten" stage and then the "relax" stage.

## 4.2 Analysis of the "Tighten" Stage

Within the "tighten" stage, we employ SOAP (Algorithm 1) to obtain the final estimator $\widehat{\mathcal{U}}$ for $\mathcal{U}^*$. Recall that $\mathcal{U}^{(t)}$ is the column space of $\mathbf{U}^{(t)}$ in Algorithm 1. The initialization obtained from the "relax" stage is denoted as $\mathbf{U}^{\text{init}}$ and its column space is $\mathcal{U}^{\text{init}}$.

Using the notation defined in (4.1)-(4.4), the following theorem characterizes the evolution of $D(\mathcal{U}^*,\mathcal{U}^{(t)})$ within the "tighten" stage.

**Theorem 4.2.** Let $\mathbf{x}_1, \ldots, \mathbf{x}_n$ be $n$ independent realizations of $\boldsymbol{X} \in \mathcal{M}_d(\mathbf{\Sigma}, k, s^*)$ with $n > n_{\min}$, and $\widehat{\mathbf{\Sigma}}$ be the sample covariance matrix. Suppose the sparsity parameter $\widehat{s}$ in Algorithm 1 (Line 3) is chosen such that

$$\widehat{s} = C \max\left\{\left\lceil \frac{4k}{(\gamma^{-1/2} - 1)^2} \right\rceil, 1\right\} \cdot s^* \tag{4.9}$$

for some integer constant $C \ge 1$. For Algorithm 1, if the column space $\mathcal{U}^{\text{init}}$ of the initial estimator $\mathbf{U}^{\text{init}}$ satisfies

$$D(\mathcal{U}^*,\mathcal{U}^{\text{init}}) \le R = \min\left\{\sqrt{\frac{k\gamma(1-\gamma^{1/2})}{2}}, \frac{\sqrt{2\gamma}}{4}\right\}, \tag{4.10}$$

then the iterative sequence $\{\mathcal{U}^{(t)}\}_{t=T+1}^{\infty}$ satisfies

$$\begin{aligned} D(\mathcal{U}^*,\mathcal{U}^{(t)}) &\le \xi_1 + \gamma^{(t-T-1)/4} \cdot \gamma^{-1/2} R \qquad (4.11)\\ &= \underbrace{C'\sqrt{k}\cdot \left(\frac{\lambda_k}{\lambda_k - \lambda_{k+1}}\right)^2 \cdot \frac{\sqrt{\lambda_1\lambda_{k+1}}}{\lambda_k - \lambda_{k+1}} \cdot \sqrt{\frac{s^* \cdot (k + \log d)}{n}}}_{\text{Statistical Error}} + \underbrace{\gamma^{(t-T-1)/4} \cdot \min\left\{\sqrt{\frac{k(1-\gamma^{1/2})}{2}}, \frac{\sqrt{2}}{4}\right\}}_{\text{Optimization Error}} \end{aligned}$$

with high probability. Here $C' > 0$ is a constant.

*Proof.* See §5.1 for a detailed proof. □

The condition in (4.9) states that the sparsity parameter $\widehat{s}$ in Algorithm 1 should be of the same order as the true sparsity level $s^*$, while the condition in (4.10) states that the initialization $\mathcal{U}^{(T)}$ should be sufficiently close to the true principal subspace $\mathcal{U}^*$. This condition characterizes the region where SOAP has the theoretical guarantee in (4.11). In particular, as long as the initial estimator $\mathcal{U}^{\text{init}}$ falls within such a basin of attraction whose radius is $R$, the distance between $\mathcal{U}^{(t)}$ and $\mathcal{U}^*$ is upper bounded by the sum of two terms with high probability. The first term is the upper bound of statistical error, which doesn't depend on the number of iterations $t$. The second term is the upper bound of optimization error, which decreases to zero at a geometric rate of convergence, because we have $\gamma = (3\lambda_{k+1} + \lambda_k)/(\lambda_{k+1} + 3\lambda_k) < 1$. Note this geometric optimization rate of convergence also depends on the eigengap of $\mathbf{\Sigma}$. In particular, when $\lambda_k$ is closer to $\lambda_{k+1}$, $\gamma$ is closer to one. In this case, the optimization error term decreases at a slower rate. Note that the entire optimization error term doesn't increase with the dimension $d$, which makes the SOAP algorithm suitable for ultra-high dimensional settings where $d$ is huge.

In (4.11), when $t$ is sufficiently large such that

$$\gamma^{(t-T)/4} \cdot R \leq \xi_1, \tag{4.12}$$

its right-hand side becomes $2\xi_1$. In other words, if the optimization error term in (4.11) decays to the same order as the statistical error term $\xi_1$, the subspace distance $D\big(\mathcal{U}^*, \mathcal{U}^{(t)}\big)$ is also of the same order. At this moment we halt the SOAP algorithm and obtain the subspace estimator $\widehat{\mathcal{U}} = \mathcal{U}^{(T+\widetilde{T})}$. Recall that $\widetilde{T}$ is the total number of iterations in Algorithm 1 (Line 3). To ensure (4.12) holds, $\widetilde{T}$ should satisfy

$$\widetilde{T} \geq \widetilde{T}_{\min} = \left\lceil \frac{4\log(R/\xi_1)}{\log(1/\gamma)} \right\rceil,$$

where $\gamma$, $R$ and $\xi_1$ are defined in (1.4), (1.5) and (4.3) respectively. This justifies the requirement that $\widetilde{T} \geq \widetilde{T}_{\min}$ in Theorem 4.1.

We now analyze the "relax" stage, which is used to obtain the initialization $\mathcal{U}^{\text{init}}$ that falls into the basin of attraction for SOAP.

### 4.3 Analysis of the "Relax" Stage

Within the "relax" stage, we employ ADMM to compute the minimization problem in (3.2). Recall that $\mathcal{U}^*$ denotes the true $k$-dimensional principal subspace, and $\mathbf{\Pi}^*$ is the corresponding projection matrix. In Algorithm 3, $\mathbf{\Pi}^{(t)}$ denotes the estimator for $\mathbf{\Pi}^*$ at the $t$-th iteration. We define

$$\overline{\mathbf{\Pi}}^{(t)} = 1/t \cdot \sum_{i=1}^{t} \mathbf{\Pi}^{(i)}, \tag{4.13}$$

and $\mathcal{U}^{(t)}$ as the $k$-dimensional subspace spanned by the top $k$ leading eigenvectors of $\overline{\mathbf{\Pi}}^{(t)}$.

Using the notation defined in (4.1)-(4.4), the following theorem characterizes the evolution of $D\big(\mathcal{U}^*, \mathcal{U}^{(t)}\big)$ within the "relax" stage.

**Theorem 4.3.** Let $\mathbf{x}_1, \ldots, \mathbf{x}_n$ be independent realizations of $\boldsymbol{X} \in \mathcal{M}_d(\boldsymbol{\Sigma}, k, s^*)$, and $\widehat{\boldsymbol{\Sigma}}$ be the sample covariance matrix. We set the regularization parameter $\rho$ in (3.2) to be $\rho = C\lambda_1\sqrt{\log d/n}$ for a sufficiently large constant $C>0$, and the penalty parameter $\beta$ (Line 3 of Algorithm 3) to be $\beta = d \cdot \rho/\sqrt{k}$. Then the iterative sequence of $k$-dimensional subspaces $\{\mathcal{U}^{(t)}\}_{t=1}^{T}$ satisfies

$$D\big(\mathcal{U}^*, \mathcal{U}^{(t)}\big) \leq \zeta_1 + \zeta_2 \cdot \frac{1}{\sqrt{t}} \tag{4.14}$$

$$= \underbrace{\frac{C'\lambda_1}{\lambda_k - \lambda_{k+1}} \cdot s^*\sqrt{\frac{\log d}{n}}}_{\text{Statistical Error}} + \underbrace{\frac{C''\sqrt{\lambda_1}}{\sqrt{\lambda_k - \lambda_{k+1}}} \cdot \left(\frac{k \cdot d^2 \log d}{n}\right)^{1/4} \cdot \frac{1}{\sqrt{t}}}_{\text{Optimization Error}}$$

with high probability. Here $C' = 4C$, where $C$ is the constant in the expression of $\rho$.

*Proof.* See §5.2 for a detailed proof. □

In (4.14), the optimization error term decreases to zero at the rate of $1/\sqrt{t}$. Note the optimization error term increases with $k$, $s^*$ and $d$. In particular, compared with the optimization error term in (4.11) of Theorem 4.2, which doesn't increase with $d$, this term increases with $d$ at the rate of $\sqrt{d} \cdot (\log d)^{1/4}$. Meanwhile, according to our analysis in §3.2, each iteration of Algorithm 3 requires $\mathcal{O}(d^3)$ operations. Furthermore, since the statistical error term has a suboptimal statistical rate of convergence, even if we take $t \to \infty$ in (4.14), the upper bound of $D\big(\mathcal{U}^*, \mathcal{U}^{(t)}\big)$ is still sub-optimal. Therefore, we early stop ADMM after $T$ iterations as soon as $\mathcal{U}^{(T)}$ enters the basin of attraction of the "tighten" stage, and switch to SOAP to obtain the optimal estimator. To ensure that $\mathcal{U}^{(T)}$ falls within the basin of attraction for SOAP, $T$ should be sufficiently large such that $D\big(\mathcal{U}^*, \mathcal{U}^{(T)}\big) \leq R$. By (4.14), $T$ should satisfy

$$\zeta_1 + \zeta_2 \cdot \frac{1}{\sqrt{T}} \leq R, \tag{4.15}$$

which is equivalent to

$$T \geq T_{\min} = \left\lceil \frac{\zeta_2^2}{(R - \zeta_1)^2} \right\rceil,$$

where $\zeta_1$, $\zeta_2$ and $R$ are defined in (4.2) and (1.5) respectively. Meanwhile, to ensure that $\zeta_1 < R$ in (4.15), the sample size $n$ should satisfy $n > n_{\min}$ where $n_{\min}$ is defined in (4.1). This is obtained by solving $n$ in $\zeta_1 < R$. Thus, the requirements $T \geq T_{\min}$ and $n > n_{\min}$ in Theorem 4.1 are justified.

The development of Theorem 4.3 is nontrivial in several aspects:

- Note that Theorem 4.3 reduces to the pure statistical result in Vu et al. (2013) when $t \to \infty$. However, our result can't be obtained from a straightforward combination of statistical and optimization results. In detail, a direct way to derive such an upper bound is using triangular inequality

$$D\big(\mathcal{U}^*, \mathcal{U}^{(t)}\big) \leq D\big(\mathcal{U}^*, \widehat{\mathcal{U}}'\big) + D\big(\widehat{\mathcal{U}}', \mathcal{U}^{(t)}\big). \tag{4.16}$$

Here $\widehat{\mathcal{U}}'$ denotes the subspace spanned by the top $k$ leading eigenvectors of $\widehat{\mathbf{\Pi}}'$, which is the exact minimizer of (3.2). For the first term on the right-hand side of (4.16), Vu et al. (2013)'s result applies. However, it is hard to obtain an upper bound for the second term because the exact minimizer of (3.2) may not be unique (Note that the objective function in (3.2) is not strongly convex).

- Meanwhile, although the introduced equality constraint in (3.4) eases the development of the corresponding convex optimization algorithm, it makes the theoretical analysis more challenging. First, for a finite $t$, the equality constraint $\mathbf{\Pi}^{(t)} = \mathbf{\Phi}^{(t)}$ only holds when the penalty parameter $\beta$ in (3.6) is set to infinity, which is not practical. Therefore, $\mathbf{\Pi}^{(t)}$ and $\mathbf{\Phi}^{(t)}$ don't fall in the feasible set of (3.4). Second, because ADMM is essentially a primal-dual algorithm, the iterative sequence of objective function values is not monotone with respect to $t$.

To address these issues, instead of separately analyzing the statistical and optimization error terms on the right-hand side of (4.16), in §5 we will conduct a joint statistical and optimization analysis, which is bridged via variational inequality. This technique provides a new way of characterizing the statistical accuracy of the intermediate solutions along the solution path of ADMM.

In the following, we analyze the total computational complexity of our framework and illustrate an interesting phenomenon: Larger sample size can reduce the overall computational complexity of our framework.

## 4.4 Blessing of Large Sample Size

Recall that the "relax" stage requires at least $T_{\min}$ iterations, and each iteration takes $\mathcal{O}(d^3)$ operations. Meanwhile, the "tighten" stage requires at least $\widetilde{T}_{\min}$ iterations, and each iteration requires $\mathcal{O}(k \cdot d^2)$ operations (or $\mathcal{O}(k \cdot d \cdot n)$ operations when $\widehat{\mathbf{\Sigma}}$ is the sample covariance matrix; In the following, we will focus on the first case for simplicity). Hence, the total number of operations needed to attain the minimax-optimal estimator $\widehat{\mathcal{U}}$ is

$$\mathcal{O}\Big(\underbrace{T_{\min} \cdot d^3}_{\text{(i)}} + \underbrace{\widetilde{T}_{\min} \cdot k \cdot d^2}_{\text{(ii)}}\Big). \tag{4.17}$$

Recall that, as defined in (4.4),

$$T_{\min} = \left\lceil \frac{\zeta_2^2}{(R - \zeta_1)^2} \right\rceil, \qquad \widetilde{T}_{\min} = \left\lceil \frac{4 \log(R/\xi_1)}{\log(1/\gamma)} \right\rceil$$

where by (4.2)-(4.3) we have $\zeta_1 \propto \sqrt{\log d/n}$, $\zeta_2 \propto (d^2 \log d/n)^{1/4}$ and $\xi_1 \propto \sqrt{\log d/n}$. Thus, for terms (i) and (ii) in (4.17) we have

$$\text{(i)}: \ T_{\min} \cdot d^3 \propto d^4 \cdot \sqrt{\frac{\log d}{n}}, \qquad \text{(ii)}: \ \widetilde{T}_{\min} \cdot k \cdot d^2 \propto d^2 \cdot \sqrt{\log n},$$

where we ignore the terms involving $s^*$ and $k$ because they are much smaller than $d$ and $n$. In the regimes where $d \gg n$, the first term in (4.17), i.e., the cost of the "relax" stage is the dominating term. When $n$ increases, the total number of operations required to obtain the optimal principal subspace estimator $\widehat{\mathcal{U}}$ decreases.

It is well-known that larger sample size improves the accuracy of estimation. Here we show that, larger sample size can also accelerate computation. This phenomenon, named the blessing of large sample size, has been observed in other settings like semi-supervised learning (Urner et al., 2011), but is observed for the first time for sparse PCA by us.

This phenomenon can be understood as follows. The radius of the basin of attraction $R$ defined in (1.5) doesn't depend on the dimension $d$ or sample size $n$. In the "relax" stage, as $n$ increases, the statistical error term in (4.14) decreases. Consequently, the iterative sequence of $\mathcal{U}^{(t)}$ generated by the "relax" stage enters the basin of attraction with a smaller number of iterations $T_{\min}$. Meanwhile, note that the computational complexity of each iteration of ADMM is higher than that of SOAP by an order of $d$. Moreover, the optimization rate of convergence of ADMM is slower than that of SOAP. Hence, the total complexity of the "relax" stage is the dominating term. Thus, a smaller $T_{\min}$ reduces the overall computational complexity. Intuitively, as $n$ increases, the objective function of the convex relaxation in (3.2) becomes "steeper", in the sense that it is easier for the iterative sequence of $\mathcal{U}^{(t)}$ of the "relax" stage to reach the basin of attraction for SOAP.

In the sequel, we will extend the sub-Gaussian assumption in $\mathcal{M}_d(\mathbf{\Sigma}, k, s^*)$ to more general settings. We will also characterize the performance of our framework for the settings where the data samples are not independently distributed.

### 4.5 Non-Gaussian and Dependent Data

**Non-Gaussian Data:** We consider a general distribution family named the transelliptical family (Han and Liu, 2012), which extends the elliptical distribution family with monotone marginal transformations. Before we define the transelliptical family, we briefly introduce the elliptical family.

In the following, we denote by $\boldsymbol{Z}_1 \overset{d}{=} \boldsymbol{Z}_2$ if $\boldsymbol{Z}_1$ and $\boldsymbol{Z}_2$ have the same distribution. We denote the $d$-dimensional $\ell_2$ unit sphere $\{\mathbf{v} : \|\mathbf{v}\|_2 = 1, \mathbf{v} \in \mathbb{R}^d\}$ by $\mathbb{S}^{d-1}$. For a matrix $\mathbf{M} \in \mathbb{R}^{d \times d}$, we overload $\mathrm{diag}(\mathbf{M})$ to be a diagonal matrix with diagonal entries $[\mathrm{diag}(\mathbf{M})]_{jj} = \mathbf{M}_{jj}$ $(j = 1, \ldots, d)$.

**Definition 4.4** (Elliptical Distribution)**.** For $\boldsymbol{\mu} \in \mathbb{R}^d$ and $\mathbf{A} \in \mathbb{R}^{d \times r}$ satisfying $\mathbf{A}\mathbf{A}^T = \mathbf{\Sigma}$ and $\mathrm{rank}(\mathbf{\Sigma}) = r \le d$, a random vector $\boldsymbol{W} \in \mathbb{R}^d$ follows an elliptical distribution, denoted by $\mathrm{EC}_d(\boldsymbol{\mu}, \mathbf{\Sigma}, \Xi)$, if and only if

$$\boldsymbol{W} \overset{d}{=} \boldsymbol{\mu} + \Xi \mathbf{A} \boldsymbol{U}.$$

Here $\Xi$ is a nonnegative random variable, $\boldsymbol{U} \in \mathbb{R}^r$ is a random vector uniformly distributed on $\mathbb{S}^{r-1}$ that is independent with $\Xi$.

**Remark 4.5.** Note that simultaneously scaling $\Xi$ and $\boldsymbol{U}$, e.g., $\Xi \to \Xi/C$ and $\boldsymbol{U} \to \boldsymbol{U}/C$, where $C$ is a constant, yields the same elliptical distribution. To ensure that the statistical model is identifiable, we assume $\mu_j = \mathbb{E}(W_j)$ and $\mathbf{\Sigma}_{jj} = \mathrm{Var}(W_j)$.

The transelliptical family is defined as below.

**Definition 4.6** (Transelliptical Distribution)**.** For a random vector $\boldsymbol{X} = (X_1, \ldots, X_d)^T$, if there exist a set of strictly increasing univariate functions $f_1(\cdot), \ldots, f_d(\cdot)$ such that $\big[f_1(X_1), \ldots, f(X_d)\big]^T$ follows the elliptical distribution $\mathrm{EC}_d(\mathbf{0}, \boldsymbol{\Sigma}, \Xi)$, where $\mathrm{diag}(\boldsymbol{\Sigma}) = \mathbf{I}_d$ and $\mathbb{P}(\Xi = 0) = 0$, then we say $\boldsymbol{X}$ follows a transelliptical distribution, denoted by $\mathrm{TE}_d(\boldsymbol{\Sigma}, \Xi; f_1, \ldots, f_d)$. Here $\boldsymbol{\Sigma}$ is referred to as the latent generalized correlation matrix.

Since the transelliptical family allows marginal transformations, it gains more flexibility than the elliptical family, e.g., distributions with asymmetric and multi-mode density. Within the transelliptical family, our goal is to estimate the $k$-dimensional principal subspace of the latent generalized correlation matrix $\boldsymbol{\Sigma}$. To this end, we employ a rank-based estimator $\widehat{\mathbf{R}}$ as suggested in (Liu et al., 2012; Han and Liu, 2012, 2013b). In detail, let $\mathbf{x}_1, \ldots, \mathbf{x}_n \in \mathbb{R}^d$ with $\mathbf{x}_i = (x_{i,1}, \ldots, x_{i,d})^T$ be $n$ independent observations of $\boldsymbol{X}$. The Kendall's tau correlation coefficient is defined as

$$\widehat{\tau}_{j,k}(\mathbf{x}_1, \ldots, \mathbf{x}_n) = \begin{cases} \displaystyle\sum_{1 \le i < i' < n} \frac{2\,\mathrm{sign}\left(x_{i,j} - x_{i',j}\right)\mathrm{sign}\left(x_{i,k} - x_{i',k}\right)}{n(n-1)}, & \text{for } j \neq k, \\ 1, & \text{for } j = k. \end{cases}$$

Then estimator $\widehat{\mathbf{R}}$ of the latent generalized correlation matrix $\boldsymbol{\Sigma}$ is defined as

$$\widehat{\mathbf{R}} = \big[\widehat{\mathbf{R}}_{j,k}\big] = \left[\sin\left(\frac{\pi}{2} \cdot \widehat{\tau}_{j,k}\left(\mathbf{x}_1, \ldots, \mathbf{x}_n\right)\right)\right]. \tag{4.18}$$

We then use $\widehat{\mathbf{R}}$ to replace the covariance estimator $\widehat{\boldsymbol{\Sigma}}$ in Algorithms 1 and 3. In theory, our framework has similar guarantees as in the sub-Gaussian setting. The only difference is to replace the key quantity $\zeta_1$ defined in (4.2), which is used in the analysis of the "relax" stage, with

$$\zeta_1^{\mathrm{TE}} = \frac{C}{\lambda_k - \lambda_{k+1}} \cdot s^* \sqrt{\frac{\log d}{n}}, \tag{4.19}$$

and replace $\xi_1$ in (4.3), which is used in the analysis of the "tighten" stage, with

$$\xi_1^{\mathrm{TE}} = C'k \cdot \left(\frac{\lambda_k}{\lambda_k - \lambda_{k+1}}\right)^2 \cdot \frac{\lambda_1}{\lambda_k - \lambda_{k+1}} \cdot \sqrt{\frac{s^* \log d}{n}}. \tag{4.20}$$

Moreover, in Theorems 4.1 and 4.3 we need to replace the regularization parameter $\rho = C\lambda_1\sqrt{\log d/n}$ with

$$\rho^{\mathrm{TE}} = C'\sqrt{\log d/n}, \tag{4.21}$$

where $C'$ is a sufficiently large constant. Then Theorems 4.1-4.3 still hold. Recall that the minimax lower bound in (4.6) is constructed using a worst case in which the data are Gaussian distributed (Vu and Lei, 2013). Therefore, (4.6) is a lower bound for the transelliptical family since it includes Gaussian. Thus, the statistical rate of convergence given in Theorem 4.1 (with $\xi_1$ replaced by $\xi_1^{\mathrm{TE}}$) is optimal with respect to $s^*$, $d$ and $n$ in the transelliptical family. See §5.3 for a detailed proof.

**Dependent Data:** We consider data with temporal dependency. In particular, we assume that the data are generated by an underlying vector autoregressive (VAR) model, which is a widely adopted model in time series analysis. In detail, we consider a stochastic process $\{\boldsymbol{X}_i\}_{i=-\infty}^{\infty}$ that satisfies

$$\boldsymbol{X}_{i+1} = \mathbf{T}\boldsymbol{X}_i + \boldsymbol{Z}_{i+1} \quad (i \in \mathbb{Z}), \tag{4.22}$$

where $\mathbf{T} \in \mathbb{R}^{d\times d}$ is the transition matrix and $\boldsymbol{Z}_{i+1} \sim N_d(\mathbf{0}, \boldsymbol{\Psi})$ is Gaussian noise. Here $\boldsymbol{X}_i, \boldsymbol{Z}_{i+1}$ are independent, while $\boldsymbol{Z}_i$'s ($i \in \mathbb{Z}$) are mutually independent. A necessary and sufficient condition for such a vector autoregressive process to be weakly stationary is $\|\mathbf{T}\|_2 < 1$. Given the realizations of $\boldsymbol{X}_1, \ldots, \boldsymbol{X}_n$, namely, $\mathbf{x}_1, \ldots, \mathbf{x}_n$, our goal is to estimate the $k$-dimensional principal subspace of the marginal covariance matrix $\boldsymbol{\Sigma} = \mathrm{Cov}(\boldsymbol{X}_i) = \mathrm{Cov}(\boldsymbol{X}_{i+1}) = \cdots$. Because this process is weakly stationary, $\boldsymbol{\Sigma}$ satisfies

$$\boldsymbol{\Sigma} = \mathbf{T}\boldsymbol{\Sigma}\mathbf{T}^T + \boldsymbol{\Psi}.$$

Though $\mathbf{x}_1, \ldots, \mathbf{x}_n$ are not independent, we can still plug in the sample covariance matrix to be the covariance estimator $\widehat{\boldsymbol{\Sigma}}$ in Algorithms 1 and 3, and get similar theoretical guarantees as in the independence setting. The difference is to replace $\zeta_1$ and $\xi_1$ defined in (4.2)-(4.3) with

$$\zeta_1^{\mathrm{VAR}} = \frac{C}{1-\|\mathbf{T}\|_2} \cdot \frac{\max_j(\boldsymbol{\Sigma}_{j,j})}{\min_j(\boldsymbol{\Sigma}_{j,j})} \cdot \frac{\lambda_1}{\lambda_k - \lambda_{k+1}} \cdot s^* \sqrt{\frac{\log d}{n}}, \tag{4.23}$$

$$\xi_1^{\mathrm{VAR}} = \frac{C'k}{\sqrt{1-\|\mathbf{T}\|_2}} \cdot \left(\frac{\lambda_k}{\lambda_k - \lambda_{k+1}}\right)^2 \cdot \frac{\sqrt{\lambda_1}}{\lambda_k - \lambda_{k+1}} \cdot \sqrt{\frac{s^* \log d}{n}}, \tag{4.24}$$

where $\max_j(\boldsymbol{\Sigma}_{j,j})$ and $\min_j(\boldsymbol{\Sigma}_{j,j})$ denote the maximum and minimum diagonal entries of $\boldsymbol{\Sigma}$. Also, we replace the regularization parameter $\rho = C\lambda_1\sqrt{\log d/n}$ in Theorems 4.1 and 4.3 with

$$\rho^{\mathrm{VAR}} = \frac{C'\lambda_1}{1-\|\mathbf{T}\|_2} \cdot \frac{\max_j(\boldsymbol{\Sigma}_{j,j})}{\min_j(\boldsymbol{\Sigma}_{j,j})} \cdot \sqrt{\frac{\log d}{n}}, \tag{4.25}$$

where $C'$ is a sufficiently large constant. Then Theorems 4.1-4.3 still hold.

In detail, in the modified version of Theorem 4.1 (with $\xi_1$ replaced by $\xi_1^{\mathrm{VAR}}$), $\xi_1^{\mathrm{VAR}}$ denotes the statistical rate of convergence of the final estimator $\widehat{\mathcal{U}}$. This statistical rate explicitly characterizes how data dependency affects the final estimation accuracy. Particularly, $\|\mathbf{T}\|_2 \in [0, 1)$ quantifies the degree of data dependency. For example, if $\|\mathbf{T}\|_2 = 0$, i.e., $\mathbf{T} = \mathbf{0}$, then $\mathbf{x}_1, \ldots, \mathbf{x}_n$ are independent. If $\|\mathbf{T}\|_2$ is close to one, e.g., $\mathbf{T}$ is close to the identity matrix, then $\mathbf{x}_1, \ldots, \mathbf{x}_n$ have higher degree of dependency. For the latter situation, $\xi_1^{\mathrm{VAR}}$ is larger, i.e., the statistical rate of convergence of $\widehat{\mathcal{U}}$ is slower. Hence, the estimation is less accurate. However, assuming that $\|\mathbf{T}\|_2$ doesn't increase with $s^*$, $d$ and $n$, while $k$ is fixed, $\widehat{\mathcal{U}}$ still achieves the minimax lower bound in (4.7) with respect to $s^*$, $d$ and $n$. This justifies the popular practice of directly applying sparse PCA on temporally dependent data, e.g., functional magnetic resonance imaging (fMRI) data or stock price data.

Furthermore, note that in (4.23)-(4.24), a lower degree of data dependency, i.e., a smaller $\|\mathbf{T}\|_2$, leads to smaller statistical error terms $\zeta_1^{\mathrm{VAR}}$ and $\xi_1^{\mathrm{VAR}}$. Equivalently, it can be viewed as increasing the sample size in the i.i.d. data setting. As a direct consequence of our analysis for the "blessing of large sample size" phenomenon in §4.4, a lower degree of data dependency can also accelerate the computation of our framework.

## 5 Proof of Main Results

We sketch the proof of main results in this section. We first establish Theorem 4.2 for the "tighten" stage, and then Theorem 4.3 for the "relax" stage. Combining these two theorems, the main result in Theorem 4.1 directly follows. We then establish the theoretical guarantees for non-Gaussian and dependent data.

### 5.1 Proof of Theorem 4.2

Before we lay out the proof, we introduce some notation. Recall that $\mathbf{U}^*$ is the orthonormal matrix corresponding to the $k$-dimensional principal subspace of $\boldsymbol{\Sigma}$, and $\mathcal{S}^*$ is the row-support of $\mathbf{U}^*$, i.e., the index set of its nonzero rows. Let $\mathcal{I} \subseteq \{1, \ldots, d\}$ be an index set. Recall that $\widehat{\boldsymbol{\Sigma}}_{\mathcal{I}} \in \mathbb{R}^{d \times d}$ denotes the restriction of $\widehat{\boldsymbol{\Sigma}}$ onto the columns and rows indexed by $\mathcal{I}$. Hereafter, we define $\widehat{\mathbf{U}}(\mathcal{I}) \in \mathbb{R}^{d \times k}$ to be the orthonormal matrix consisting of the top $k$ leading eigenvectors of $\widehat{\boldsymbol{\Sigma}}_{\mathcal{I}}$. Correspondingly, the column space of $\widehat{\mathbf{U}}(\mathcal{I})$ is denoted by $\widehat{\mathcal{U}}(\mathcal{I})$. For notational simplicity, we define

$$\big\|\widehat{\boldsymbol{\Sigma}} - \boldsymbol{\Sigma}\big\|_{2,|\mathcal{I}|} = \sup_{\|\mathbf{v}\|_0 \le |\mathcal{I}|,\ \|\mathbf{v}\|_2 = 1} \big|\mathbf{v}^T\big(\widehat{\boldsymbol{\Sigma}} - \boldsymbol{\Sigma}\big)\mathbf{v}\big|. \tag{5.1}$$

Next, we lay out an assumption which significantly simplifies our proof. Given this assumption, the proof of Theorem 4.2 is in a deterministic fashion. In Lemmas 5.2 and 5.3, we prove that this assumption holds with high probability for different applications. This assumption also provides a convenient interface to extend our framework to non-Gaussian and dependent data.

**Assumption 5.1.** Under the above notation, we assume:

- If $\mathcal{S}^* \subseteq \mathcal{I}$, then

$$D\big[\mathcal{U}^*, \widehat{\mathcal{U}}(\mathcal{I})\big] \le C \cdot \frac{\sqrt{\lambda_1 \lambda_{k+1}}}{\lambda_k - \lambda_{k+1}} \cdot \sqrt{\frac{|\mathcal{I}| \cdot (k + \log d)}{n}}. \tag{5.2}$$

- For any $\mathcal{I} \subseteq \{1, \ldots, d\}$ with $|\mathcal{I}| \le d/2$,

$$\big\|\widehat{\boldsymbol{\Sigma}} - \boldsymbol{\Sigma}\big\|_{2,|\mathcal{I}|} \le C' \lambda_1 \sqrt{\frac{|\mathcal{I}| \cdot \log d}{n}}. \tag{5.3}$$

The following two lemmas prove that Assumption 5.1 holds with high probability.

**Lemma 5.2.** Let $\mathbf{x}_1, \ldots, \mathbf{x}_n$ be independent realizations of $\boldsymbol{X} \in \mathcal{M}_d(\boldsymbol{\Sigma}, k, s^*)$, and $\widehat{\boldsymbol{\Sigma}}$ be the sample covariance matrix. For $n$ sufficiently large, (5.2) holds with probability at least $1 - 4/(n-1) - 1/d - 6\log n/n$.

*Proof.* This proof is extended from the derivation for the main upper bound in Vu and Lei (2013). See §B.1 for details. □

**Lemma 5.3.** Let $\mathbf{x}_1, \ldots, \mathbf{x}_n$ be independent realizations of $\boldsymbol{X} \in \mathcal{M}_d(\boldsymbol{\Sigma}, k, s^*)$, and $\widehat{\boldsymbol{\Sigma}}$ be the sample covariance matrix. For any $\mathcal{I} \subseteq \{1, \ldots, d\}$ with $|\mathcal{I}| \leq d/2$, we have

$$\big\|\widehat{\boldsymbol{\Sigma}} - \boldsymbol{\Sigma}\big\|_{2,|\mathcal{I}|} \leq C\lambda_1 \sqrt{\frac{|\mathcal{I}| \cdot \log d}{n}}$$

with probability at least $1 - 1/C$, for a sufficiently large constant $C > 0$ and sample size $n$.

*Proof.* The proof follows from Lemma 3.2.4 of Vu and Lei (2012). □

Combining Lemmas 5.2 and 5.3, we prove that Assumption 5.1 holds with probability at least $1 - 4/(n-1) - 1/d - 6\log n/n - 1/C'$ where $C'$ is a sufficiently large constant in (5.3). Under Assumption 5.1, the remaining proofs are all deterministic. Our proof strategy is as follows.

**Proof Strategy:** We first present two key lemmas. Lemma 5.4 characterizes the step of matrix multiplication and renormalization in Lines 6 and 7 of Algorithm 1. As is discussed in §3.1, such a step can be viewed as a generalized projected gradient step. We will prove that, this step decreases the distance between the estimated subspace and true subspace $\mathcal{U}^*$ by a multiplicative factor. Lemma 5.5 characterizes the sparsification step in Lines 8 and 9 of Algorithm 1. This lemma states that, the sparsification step introduces extra error to the distance between the estimated subspace and true subspace $\mathcal{U}^*$. However, this error can be upper bounded so that each iteration of SOAP decays the overall optimization error by a multiplicative factor. Equipped with these two lemmas and an analysis for the initialization step in Line 4, the results in Theorem 4.2 can be proved by mathematical induction.

Recall in Algorithm 1, the first key step is the orthogonal iteration operation (Line 6 and Line 7), i.e,

$$\widetilde{\mathbf{V}}^{(t+1)} \leftarrow \widehat{\boldsymbol{\Sigma}} \cdot \mathbf{U}^{(t)} \tag{5.4}$$

$$\mathbf{V}^{(t+1)}, \mathbf{R}_1^{(t+1)} \leftarrow \mathsf{Thin_QR}\big(\widetilde{\mathbf{V}}^{(t+1)}\big), \tag{5.5}$$

where $\mathbf{U}^{(t)}$ is obtained from the previous iteration of SOAP. Recall that for $t > T+1$, $\mathbf{U}^{(t)}$ is computed by first taking a truncation step (Line 8), and then a renormalization step (Line 9), which doesn't change its row-sparsity pattern. Thus, $\mathbf{U}^{(t)}$ is orthonormal and row-sparse. Meanwhile, for $t = T+1$, since we obtain $\mathbf{U}^{(t)}$ by truncating and then renormalizing the initial estimator $\mathbf{U}^{\text{init}}$ (Line 4), $\mathbf{U}^{(t)}$ is also orthonormal and row-sparse. The next lemma characterizes the relationship between $\mathbf{V}^{(t+1)}$ and $\mathbf{U}^{(t)}$ in (5.4)-(5.5). In the following, we denote $\mathcal{U}^{(t)}$ to be the column space of $\mathbf{U}^{(t)}$ and $\mathcal{V}^{(t+1)}$ to be that of $\mathbf{V}^{(t+1)}$. Also remind that $\gamma = (3\lambda_{k+1} + \lambda_k)/(\lambda_{k+1} + 3\lambda_k) < 1$.

**Lemma 5.4.** Let $\mathcal{I}$ be a superset of the row-support of $\mathbf{U}^{(t)}$. Under Assumption 5.1, we choose $n$ sufficiently large so that

$$\big\|\widehat{\boldsymbol{\Sigma}} - \boldsymbol{\Sigma}\big\|_{2,|\mathcal{I}|} \leq (\lambda_k - \lambda_{k+1})/4. \tag{5.6}$$

If $D\big[\mathcal{U}^{(t)}, \widehat{\mathcal{U}}(\mathcal{I})\big] < \sqrt{2}$, then we have

$$D\big[\mathcal{V}^{(t+1)}, \widehat{\mathcal{U}}(\mathcal{I})\big] \leq \frac{D\big[\mathcal{U}^{(t)}, \widehat{\mathcal{U}}(\mathcal{I})\big]}{\sqrt{1 - D\big[\mathcal{U}^{(t)}, \widehat{\mathcal{U}}(\mathcal{I})\big]^2/(2k)}} \cdot \gamma. \tag{5.7}$$

*Proof.* See §B.2 for a detailed proof. □

Lemma 5.4 states that, the output $\mathcal{V}^{(t+1)}$ of the orthogonal iteration step, which is described in (5.4)-(5.5), is closer to $\widehat{\mathcal{U}}(\mathcal{I})$ than the input $\mathcal{U}^{(t)}$. In detail, when $D\big[\mathcal{U}^{(t)},\widehat{\mathcal{U}}(\mathcal{I})\big]$ is sufficiently small, (5.7) indicates that $D\big[\mathcal{V}^{(t+1)},\widehat{\mathcal{U}}(\mathcal{I})\big]$ is roughly smaller than $D\big[\mathcal{U}^{(t)},\widehat{\mathcal{U}}(\mathcal{I})\big]$ by a factor of $\gamma$. Suppose that $\mathcal{I}$ also contains $\mathcal{S}^*$. Then by (5.2), $\widehat{\mathcal{U}}(\mathcal{I})$ and the true principal subspace $\mathcal{U}^*$ are sufficiently close. Hence, $D\big[\mathcal{V}^{(t+1)},\mathcal{U}^*\big]$ is also roughly smaller than $D\big[\mathcal{U}^{(t)},\mathcal{U}^*\big]$ by a multiplicative factor. In the sequel, we lay out the second key lemma that characterizes the sparsification step.

Recall that in Algorithm 1, the second key step is the sparsification step (Line 8 and Line 9), i.e,

$$\widetilde{\mathbf{U}}^{(t+1)} \leftarrow \mathsf{Truncate}\big(\mathbf{V}^{(t+1)},\widehat{s}\big), \tag{5.8}$$

$$\mathbf{U}^{(t+1)},\mathbf{R}_2^{(t+1)} \leftarrow \mathsf{Thin_QR}\big(\widetilde{\mathbf{U}}^{(t+1)}\big), \tag{5.9}$$

where $\widehat{s}$ is the sparsity parameter of Algorithm 1. The next lemma characterizes the relationship between $D\big[\mathcal{U}^*,\mathcal{U}^{(t+1)}\big]$ and $D\big[\mathcal{U}^*,\mathcal{V}^{(t+1)}\big]$.

**Lemma 5.5.** Suppose that $\sqrt{s^*/\widehat{s}} \leq 1$ and $D\big[\mathcal{U}^*,\mathcal{V}^{(t+1)}\big] \leq 1$. We have

$$D\big[\mathcal{U}^*,\mathcal{U}^{(t+1)}\big] \leq \left(1+2\sqrt{\frac{k\cdot s^*}{\widehat{s}}}\right)\cdot D\big[\mathcal{U}^*,\mathcal{V}^{(t+1)}\big]. \tag{5.10}$$

*Proof.* See §B.3 for a detailed proof. □

Lemma 5.5 characterizes the error introduced by the sparsification step, which is described in (5.8)-(5.9). Since we greedily select the top $\widehat{s}$ most significant rows, the error is upper bounded by $2\sqrt{k\cdot s^*/\widehat{s}}\cdot D\big[\mathcal{U}^*,\mathcal{V}^{(t+1)}\big]$. Particularly, if we choose a larger sparsity parameter $\widehat{s}$ in Algorithm 1, this error is smaller. Intuitively, this is because more information is preserved at the truncation step. In fact, the $\sqrt{k}$ factor in this error is due to an artifact in our analysis. In the proof we will show that, if one can get a tighter upper bound for $\big\|\mathbf{U}^*_{\mathcal{I}_1,\cdot}\big\|_2$ in (B.35), then this $\sqrt{k}$ factor can be avoided. Consequently, the $\sqrt{k}$ factor in $\xi_1$ defined in (4.3) can also be avoided. Then the statistical rate of convergence in Theorem 4.1 is optimal within more general regimes, where $k$ is also allowed to increase with $s^*$, $d$ and $n$. However, in this paper, our primary focus is on the statistical rate of convergence with respect to $s^*$, $d$ and $n$, and hence we treat $k$ as fixed. See our discussion for minimax-optimality in §4.1 for a justification using the application of population genetics.

Equipped with Lemmas 5.4 and 5.5, we now establish our main lemma, which characterizes the evolution of estimation error within a single iteration of SOAP. For notational simplicity, we define

$$\Delta(s) = C\cdot\frac{\sqrt{\lambda_1\lambda_{k+1}}}{\lambda_k-\lambda_{k+1}}\cdot\sqrt{\frac{s\cdot(k+\log d)}{n}}, \tag{5.11}$$

where $s$ is an integer and $C$ is the constant in (5.2) of Assumption 5.1.

**Lemma 5.6.** We assume that

$$\widehat{s} = C\max\left\{\left\lceil\frac{4k}{(\gamma^{-1/2}-1)^2}\right\rceil,1\right\}\cdot s^*, \quad \text{and} \quad D\big[\mathcal{U}^*,\mathcal{U}^{(t)}\big] \leq \min\left\{\sqrt{2k\big[1-\gamma^{1/2}\big]},\sqrt{2}/2\right\},$$

where $C \geq 1$ is an integer constant. Under Assumption 5.1, suppose $n$ is sufficiently large such that $\Delta(2\widehat{s}) \leq 1/24$. We have

$$D\big[\mathcal{U}^*,\mathcal{U}^{(t+1)}\big] \leq \gamma^{1/4} \cdot D\big[\mathcal{U}^*,\mathcal{U}^{(t)}\big] + 3\gamma^{1/2} \cdot \Delta(2\widehat{s}).$$

*Proof.* See §B.4 for a detailed proof. □

Lemma 5.6 states that, if $D\big[\mathcal{U}^*,\mathcal{U}^{(t)}\big]$ is sufficiently small, then after the $(t+1)$-th iteration of SOAP, the overall estimation error $D\big[\mathcal{U}^*,\mathcal{U}^{(t+1)}\big]$ decays by a factor of $\gamma^{1/4}$ compared to $D\big[\mathcal{U}^*,\mathcal{U}^{(t)}\big]$, with a statistical error term $3\gamma^{1/2} \cdot \Delta(2\widehat{s})$ introduced. Given Lemma 5.6, we can then establish Theorem 4.2 using mathematical induction. See §B.5 for a detailed proof. Note that Lemma 5.6 requires $\mathcal{U}^{(t)}$ and $\mathcal{U}^*$ to be sufficiently close. Consequently, Theorem 4.2 requires the initial estimator $\mathcal{U}^{\text{init}}$ of SOAP to be sufficiently good such that it falls within the basin of attraction defined in (1.5). Recall that the results in Lemmas 5.4 and 5.6 are deterministic given Assumption 5.1, which holds with high probability as shown in Lemmas 5.2 and 5.3. Thus, the result in Theorem 4.2 holds with high probability.

## 5.2 Proof of Theorem 4.3

As in §5.1, we first lay out an assumption, conditioning on which the entire proof is in a deterministic fashion. We will then present a lemma that shows this assumption holds with high probability. Recall that $\rho$ is the regularization parameter in Line 3 of Algorithm 3.

**Assumption 5.7.** For $\rho = C\lambda_1\sqrt{\log d/n}$ with a sufficiently large constant $C > 0$, we assume

$$\big\|\widehat{\mathbf{\Sigma}} - \mathbf{\Sigma}\big\|_{\infty,\infty} \leq \rho. \tag{5.12}$$

The following lemma proves Assumption 5.7 holds with high probability.

**Lemma 5.8.** Let $\mathbf{x}_1, \ldots, \mathbf{x}_n$ be independent realizations of $\boldsymbol{X} \in \mathcal{M}_d(\mathbf{\Sigma}, k, s^*)$, and $\widehat{\mathbf{\Sigma}}$ be the sample covariance matrix. For $n$ sufficiently large, (5.12) holds with probability at least $1 - 4/d^2$.

*Proof.* Applying Bernstein's inequality and then union bound, we reach the conclusion. □

We now introduce some extra notation. Recall that in Algorithm 3, $\mathbf{\Pi}$ and $\mathbf{\Phi}$ are the primal variables in the convex relaxation in (3.4). Meanwhile, $\mathbf{\Theta}$ denotes the Lagrangian multiplier corresponding to the equality constraint of (3.4). For notational simplicity, we denote

$$\mathbf{Z} = \begin{bmatrix} \mathbf{\Pi} \\ \mathbf{\Phi} \\ \mathbf{\Theta} \end{bmatrix}, \quad h(\mathbf{Z}) = \begin{bmatrix} -\mathbf{\Theta} \\ \mathbf{\Theta} \\ \mathbf{\Pi} - \mathbf{\Phi} \end{bmatrix}, \quad \text{and} \;\; \mathbf{Z} \in \mathcal{Z} = \mathcal{A} \times \mathcal{B} \times \mathbb{R}^{d\times d}, \tag{5.13}$$

where $\mathbf{\Pi} \in \mathcal{A}$, $\mathbf{\Phi} \in \mathcal{B}$ and $\mathbf{\Theta} \in \mathbb{R}^{d\times d}$ with $\mathcal{A}$ defined in (3.3) and $\mathcal{B}$ defined in (3.4). Meanwhile, let

$$\overline{\mathbf{\Pi}}^{(t)} = 1/t \cdot \sum_{j=1}^{t} \mathbf{\Pi}^{(j)}, \;\; \overline{\mathbf{\Phi}}^{(t)} = 1/t \cdot \sum_{j=1}^{t} \mathbf{\Phi}^{(j)}, \;\; \overline{\mathbf{\Theta}}^{(t)} = 1/t \cdot \sum_{j=1}^{t} \mathbf{\Theta}^{(j)}, \;\; \overline{\mathbf{Z}}^{(t)} = 1/t \cdot \sum_{j=1}^{t} \mathbf{Z}^{(j)}. \tag{5.14}$$

Our proof strategy is as follows.

**Proof Strategy:** Our joint computational and statistical analysis is bridged via the following key quantity

$$V^{(t)}(\mathbf{Z}) = -\langle \widehat{\boldsymbol{\Sigma}}, \overline{\boldsymbol{\Pi}}^{(t)} - \boldsymbol{\Pi} \rangle + \rho \|\overline{\boldsymbol{\Phi}}^{(t)}\|_{1,1} - \rho\|\boldsymbol{\Phi}\|_{1,1} + \langle \overline{\mathbf{Z}}^{(t)} - \mathbf{Z}, h(\overline{\mathbf{Z}}^{(t)}) \rangle.$$

Here $\mathbf{Z}$ is defined in (5.13). The intuition behind this quantity follows from variational inequalities, which are used to characterize the optimality of the solutions to convex optimization problems. In detail, recall that the Lagrangian $L(\boldsymbol{\Pi}, \boldsymbol{\Phi}, \boldsymbol{\Theta})$ of (3.4) is defined in (3.5). Suppose that $\widetilde{\boldsymbol{\Pi}} \in \mathcal{A}$ is the exact minimizer of (3.4). Let the corresponding auxiliary variable be $\widetilde{\boldsymbol{\Phi}} \in \mathcal{B}$ and the Lagrangian multiplier be $\widetilde{\boldsymbol{\Theta}} \in \mathbb{R}^{d\times d}$. Respectively, we define $\widetilde{\mathbf{Z}} \in \mathcal{Z}$ in a similar way as in (5.13). A formulation of the first-order optimality condition for the convex optimization problem in (3.4) is

$$\sup_{\mathbf{Z}\in\mathcal{Z}} \Big\{ -\langle \widehat{\boldsymbol{\Sigma}}, \widetilde{\boldsymbol{\Pi}} - \boldsymbol{\Pi} \rangle + \rho \|\widetilde{\boldsymbol{\Phi}}\|_{1,1} - \rho\|\boldsymbol{\Phi}\|_{1,1} + \langle \widetilde{\mathbf{Z}} - \mathbf{Z}, h(\widetilde{\mathbf{Z}}) \rangle \Big\} \le 0, \tag{5.15}$$

which is called the variational inequality. See He et al. (2014); Kinderlehrer and Stampacchia (2000) for more details. Our optimization analysis proves that, $\sup_{\mathbf{Z}\in\mathcal{Z}} V^{(t)}(\mathbf{Z})$ decreases to zero at the rate of $1/\sqrt{t}$. According to (5.15), if we have $\sup_{\mathbf{Z}\in\mathcal{Z}} V^{(t)}(\mathbf{Z}) \le \epsilon$, where $\epsilon \ge 0$ is sufficiently small, $\boldsymbol{\Pi}^{(t)}$ is a good approximate solution to (3.4). Meanwhile, remind $\boldsymbol{\Pi}^*$ is the projection matrix corresponding to the true principal subspace $\mathcal{U}^*$. We define

$$\mathbf{Z}^* = \begin{bmatrix} \boldsymbol{\Pi}^* \\ \boldsymbol{\Pi}^* \\ \widetilde{\boldsymbol{\Theta}}^{(t)} \end{bmatrix}, \qquad \text{where} \quad \widetilde{\boldsymbol{\Theta}}^{(t)} = -\rho \cdot \mathrm{sign}\big(\overline{\boldsymbol{\Pi}}^{(t)} - \overline{\boldsymbol{\Phi}}^{(t)}\big).$$

Since $\mathbf{Z}^* \in \mathcal{Z}$, we have $V^{(t)}(\mathbf{Z}^*) \le \sup_{\mathbf{Z}\in\mathcal{Z}} V^{(t)}(\mathbf{Z})$. Therefore, $V^{(t)}(\mathbf{Z}^*)$ is also upper bounded by a quantity decaying to zero at the rate of $1/t$. Moreover, our statistical analysis proves that $V^{(t)}(\mathbf{Z}^*)$ can be lower bounded by a quadratic form of $\|\boldsymbol{\Pi}^{(t)} - \boldsymbol{\Pi}^*\|_F$. Combining the upper bound of $V^{(t)}(\mathbf{Z}^*)$ obtained from optimization analysis with this lower bound yields a quadratic inequality. Solving it we obtain an upper bound of $\|\boldsymbol{\Pi}^{(t)} - \boldsymbol{\Pi}^*\|_F$, which further implies the upper bound of $D(\mathcal{U}^{(t)}, \mathcal{U}^*)$ in Theorem 4.3. Note that our whole proof is based on Assumption 5.7, which holds with high probability according to Lemma 5.8. Therefore, our conclusion in Theorem 4.3 holds with high probability. See §C.2 for a detailed proof.

The main advantage of such a variational inequality-based analytic approach is that, it doesn't explicitly involve the exact solution to the convex optimization problem, which is not necessarily unique in our setting, because the objective function in (3.4) is not strongly convex. Also, because it doesn't directly involve objective function value, it is suitable for analyzing the statistical properties of the intermediate solutions of convex optimization algorithms for which the iterative sequence of objective function values is not monotone, e.g., ADMM.

### 5.3 Extensions for Non-Gaussian and Dependent Data

The extensions for non-Gaussian data and dependent data are established by replacing Assumptions 5.1 and 5.7 using the respective modified versions. We will prove that these modified assumptions hold with high probability.

**Non-Gaussian Data:** Recall that $\|\cdot\|_{2,|\mathcal{I}|}$ is defined in (5.1). The following assumption plays the role of Assumption 5.1 in the non-Gaussian setting.

**Assumption 5.9.** Using the notation in §5.1 and §5.2, we assume:

- If $\mathcal{S}^* \subseteq \mathcal{I}$, then

$$D\big[\mathcal{U}^*, \widehat{\mathcal{U}}(\mathcal{I})\big] \leq C \cdot \frac{\sqrt{k} \cdot \lambda_1}{\lambda_k - \lambda_{k+1}} \cdot \sqrt{\frac{|\mathcal{I}| \cdot \log d}{n}}. \tag{5.16}$$

- For any $\mathcal{I} \subseteq \{1, \ldots, d\}$ with $|\mathcal{I}| \leq d/2$,

$$\big\|\widehat{\mathbf{\Sigma}} - \mathbf{\Sigma}\big\|_{2,|\mathcal{I}|} \leq C' \lambda_1 \sqrt{\frac{|\mathcal{I}| \cdot \log d}{n}}. \tag{5.17}$$

The next two lemmas prove that Assumption 5.9 holds with high probability for non-Gaussian data. Remind that the transelliptical distribution $\mathrm{TE}_d(\mathbf{\Sigma}, \Xi; f_1, \ldots, f_d)$ is defined in Definition 4.6, and the rank-based estimator $\widehat{\mathbf{R}}$ for the latent generalized correlation matrix $\mathbf{\Sigma}$ is defined in (4.18).

**Lemma 5.10.** Suppose $\mathbf{x}_1, \ldots, \mathbf{x}_n$ are independent realizations of $\boldsymbol{X} \sim \mathrm{TE}_d(\mathbf{\Sigma}, \Xi; f_1, \ldots, f_d)$, and $\widehat{\mathbf{\Sigma}}$ is set to be $\widehat{\mathbf{R}}$. Under the sign sub-Gaussian condition (see Han and Liu (2013b) for details), we have that, for any $\mathcal{I} \subseteq \{1, \ldots, d\}$ with $|\mathcal{I}| \leq d/2$, (5.17) holds with probability at least $1 - 2/d - 1/d^2$ for a sufficiently large constant $C' > 0$ and $n$.

*Proof.* This lemma is adapted from Theorem 4.10 of Han and Liu (2013b). □

**Lemma 5.11.** Suppose we have $\mathcal{S}^* \subseteq \mathcal{I}$. Under the same condition as in Lemma 5.10, (5.16) holds with the same probability as in Lemma 5.10 for a sufficiently large $C > 0$ and $n$.

*Proof.* See §D.1 for a detailed proof. □

Lemmas 5.10 and 5.11 prove that Assumption 5.9 holds with probability at least $1 - 4/d^2 - 2/d$. The next assumption plays the role of Assumption 5.7 in the non-Gaussian setting.

**Assumption 5.12.** For $\rho^{\mathrm{TE}} = C\sqrt{\log d / n}$ with a sufficiently large constant $C > 0$, we assume

$$\big\|\widehat{\mathbf{\Sigma}} - \mathbf{\Sigma}\big\|_{\infty,\infty} \leq \rho^{\mathrm{TE}}. \tag{5.18}$$

The following lemma proves that Assumption 5.12 holds with high probability.

**Lemma 5.13.** Assume $\mathbf{x}_1, \ldots, \mathbf{x}_n$ are independent realizations of $\boldsymbol{X} \sim \mathrm{TE}_d(\mathbf{\Sigma}, \Xi; f_1, \ldots, f_d)$, and $\widehat{\mathbf{\Sigma}}$ is set to be $\widehat{\mathbf{R}}$. For $n$ sufficiently large, (5.18) holds with probability at least $1 - 4/d^2$.

*Proof.* This lemma is adapted from Theorem 4.1 of Han and Liu (2012). □

Under Assumptions 5.9 and 5.12, we can then establish our results for non-Gaussian data in §4.5 following the proof in §5.1-§5.2. Correspondingly, we need to modify the key quantities as specified in (4.19)-(4.21). Since Lemmas 5.10, 5.11 and 5.13 prove that Assumptions 5.9 and 5.12 hold with high probability, our results for non-Gaussian data also hold with high probability. In the sequel, we establish the results for dependent data in the same way.

**Dependent Data:** The following assumption replaces Assumption 5.1 for dependent data. Recall that $\mathbf{T} \in \mathbb{R}^{d\times d}$ is the transition matrix in the VAR model defined in (4.22) and satisfies $\|\mathbf{T}\|_2 < 1$.

**Assumption 5.14.** Using the notation in §5.1 and §5.2, we assume:

- If $\mathcal{S}^* \subseteq \mathcal{I}$, then

$$D\big[\mathcal{U}^*, \widehat{\mathcal{U}}(\mathcal{I})\big] \le \frac{C}{\sqrt{1-\|\mathbf{T}\|_2}} \cdot \frac{\sqrt{k\cdot\lambda_1}}{\lambda_k - \lambda_{k+1}} \cdot \sqrt{\frac{|\mathcal{I}|\cdot \log d}{n}}. \tag{5.19}$$

- For any $\mathcal{I} \subseteq \{1,\ldots,d\}$ with $|\mathcal{I}| \le d/2$,

$$\big\|\widehat{\mathbf{\Sigma}} - \mathbf{\Sigma}\big\|_{2,|\mathcal{I}|} \le \frac{C'\sqrt{\lambda_1}}{\sqrt{1-\|\mathbf{T}\|_2}} \cdot \sqrt{\frac{|\mathcal{I}|\cdot \log d}{n}}. \tag{5.20}$$

Similarly, we prove that Assumption 5.14 holds with high probability using the following two lemmas.

**Lemma 5.15.** Suppose that $\mathbf{x}_1,\ldots,\mathbf{x}_n$ are the realizations of $\boldsymbol{X}_1,\ldots,\boldsymbol{X}_n$, which are from the VAR process $\{\boldsymbol{X}_i\}_{i=-\infty}^{\infty}$ defined by (4.22), and $\widehat{\mathbf{\Sigma}}$ is the sample covariance matrix. We have that, for any $\mathcal{I} \subseteq \{1,\ldots,d\}$ with $|\mathcal{I}| \le d/2$, (5.20) holds with probability at least $1-1/d$ for a sufficiently large constant $C' > 0$ and $n$.

*Proof.* This lemma is adapted from the proof of Theorem 4.1 in Wang et al. (2013). □

**Lemma 5.16.** Suppose we have $\mathcal{S}^* \subseteq \mathcal{I}$. Under the same condition as in Lemma 5.15, (5.19) holds with the same probability as in Lemma 5.15 for a sufficiently large $C > 0$ and $n$.

*Proof.* The proof is identical to Lemma 5.11. □

Hence, we prove that Assumption 5.14 holds with probability at least $1-2/d$. In the sequel, we lay out the counterpart of Assumption 5.7 for dependent data.

**Assumption 5.17.** For

$$\rho^{\mathrm{VAR}} = \frac{C\lambda_1}{1-\|\mathbf{T}\|_2} \cdot \frac{\max_j(\mathbf{\Sigma}_{j,j})}{\min_j(\mathbf{\Sigma}_{j,j})} \cdot \sqrt{\frac{\log d}{n}}$$

with a sufficiently large constant $C > 0$, we assume

$$\big\|\widehat{\mathbf{\Sigma}} - \mathbf{\Sigma}\big\|_{\infty,\infty} \le \rho^{\mathrm{VAR}}. \tag{5.21}$$

The following lemma proves that Assumption 5.17 holds with high probability.

**Lemma 5.18.** Suppose that $\mathbf{x}_1, \ldots, \mathbf{x}_n$ are the realizations of $\boldsymbol{X}_1, \ldots, \boldsymbol{X}_n$ from the VAR process $\{\boldsymbol{X}_i\}_{i=-\infty}^{\infty}$ defined by (4.22). Assume $\widehat{\boldsymbol{\Sigma}}$ is the sample covariance matrix. We have that (5.21) holds with probability at least $1 - 6/d$.

*Proof.* This Lemma is adapted from Lemma A.1 of Han and Liu (2013a). □

Thus, Assumptions 5.14 and 5.17 both hold with high probability. Following the same argument for non-Gaussian data, we can establish the results for dependent data in §4.5. Correspondingly, we need to modify the key quantities as specified in (4.23)-(4.25).

# 6 Discussion

In the sequel, we discuss two important issues. First, we propose a practical stopping criterion of our framework. Second, we discuss the relationship between our result and the recent result on the computational lower bound for sparse PCA (Berthet and Rigollet, 2013a).

**Practical Stopping Criterion:** Recall that our theory shows that the two stages of our framework require at least $T_{\min}$ and $\widetilde{T}_{\min}$ iterations correspondingly as specified in (4.4). Nevertheless, $T_{\min}$ and $\widetilde{T}_{\min}$ involve the eigenvalues $\lambda_1$, $\lambda_k$ and $\lambda_{k+1}$ of the population covariance matrix $\boldsymbol{\Sigma}$, which are unknown in practice. In the following, we propose a stopping criterion that is useful in practice.

For the "relax" stage, Theorem 4.3 proves that $D\big(\mathcal{U}^*, \mathcal{U}^{(t)}\big) \le \zeta_1 + \zeta_2/\sqrt{t}$ holds with high probability, where $\zeta_1, \zeta_2$ are defined in (4.2) and $\mathcal{U}^{(t)}$ is defined below (4.13). By triangular inequality we have

$$D\big(\mathcal{U}^{(t)}, \mathcal{U}^{(t+1)}\big) \le D\big(\mathcal{U}^{(t)}, \mathcal{U}^*\big) + D\big(\mathcal{U}^*, \mathcal{U}^{(t+1)}\big) < 2\zeta_1 + 2\zeta_2/\sqrt{t}. \tag{6.1}$$

Hence, for a sufficiently large $t$, we have $D\big(\mathcal{U}^{(t)}, \mathcal{U}^{(t+1)}\big) \le C\zeta_1$. Remind $\zeta_1 \propto s^*\sqrt{\log d/n}$, where $s^*$ is assumed to be known in our setting. Thus, we can set the stopping criterion of the "relax" stage to be

$$D\big(\mathcal{U}^{(t)}, \mathcal{U}^{(t+1)}\big) \le \nu_1 \cdot s^* \sqrt{\frac{\log d}{n}},$$

where $\nu_1 > 0$ is a tuning parameter.

For the "tighten" stage, Theorem 4.2 proves that $D\big(\mathcal{U}^*, \mathcal{U}^{(t)}\big) \le \xi_1 + \gamma^{(t-T-1)/4} \cdot \gamma^{-1/2} R$ holds with high probability. Here $\xi_1$, $\gamma$ and $R$ are defined in (4.3), (1.4) and (1.5) respectively. Similar to (6.1), we have

$$D\big(\mathcal{U}^{(t)}, \mathcal{U}^{(t+1)}\big) \le D\big(\mathcal{U}^{(t)}, \mathcal{U}^*\big) + D\big(\mathcal{U}^*, \mathcal{U}^{(t+1)}\big) < 2\xi_1 + 2\gamma^{(t-T-1)/4} \cdot \gamma^{-1/2} R.$$

Since $\zeta_1 \propto \sqrt{s^* \log d/n}$, we can set the stopping criterion of the "tighten" stage to be

$$D\big(\mathcal{U}^{(t)}, \mathcal{U}^{(t+1)}\big) \le \nu_2 \cdot \sqrt{\frac{s^* \log d}{n}},$$

where $\nu_2 > 0$ is a tuning parameter.

**Relationship with Computational Lower Bound:** Berthet and Rigollet (2013a) consider the problem of principal component detection. In detail, their goal is to test whether the eigenvalues of the population covariance matrix $\mathbf{\Sigma}$ are $\{1+\theta, 1, \ldots, 1\}$ or $\{1, 1, \ldots, 1\}$ where $\theta > 0$ is the eigengap. They prove that when $\theta < Cs^*\sqrt{\log d/n}$ such a test reduces to a certain type of planted clique problem that is unlikely to be solved by any randomized polynomial-time algorithm. Moreover, based on this computational lower bound one can easily prove that, if the eigenvalues of $\mathbf{\Sigma}$ are $\{1+\theta, 1, \ldots, 1\}$ with $\theta < Cs^*\sqrt{\log d/n}$, there is no computational tractable estimator $\widetilde{\mathbf{u}}_1$ for the first leading eigenvector $\mathbf{u}_1^*$ of $\mathbf{\Sigma}$ that can attain the minimax-optimal statistical rate of convergence

$$\|\widetilde{\mathbf{u}}_1 - \mathbf{u}_1^*\|_2 \leq C \cdot \frac{\sqrt{\lambda_1\lambda_2}}{\lambda_1 - \lambda_2} \cdot \sqrt{\frac{s^* \log d}{n}}.$$

This is because, if otherwise, we can construct a test with the statistics $\widetilde{\mathbf{u}}_1^T \cdot \widehat{\mathbf{\Sigma}} \cdot \widetilde{\mathbf{u}}_1$ and successfully distinguish the two eigenstructures. Because $\widetilde{\mathbf{u}}_1$ can be computed in polynomial time, such a test is computationally tractable, which contradicts the computational lower bound.

Our result is consistent with such a computational lower bound. Assume that the eigenvalues of $\mathbf{\Sigma}$ are $\{1+\theta, 1, \ldots, 1\}$ with $\theta < Cs^*\sqrt{\log d/n}$ for a sufficiently small $C > 0$. Plugging $k = 1$ and $\lambda_1 - \lambda_2 = \theta$ into (4.2), we have $\zeta_1 \geq C'$ where $C' \propto 1/C$. For a sufficiently large $C'$, the upper bound of $D\big(\mathcal{U}^*, \mathcal{U}^{(t)}\big)$ in Theorem 4.3 is no longer smaller than the radius $R$ of the basin of attraction defined in (1.5) even when $t \to \infty$. Therefore, the iterative sequence within the "relax" stage is no longer guaranteed to enter the basin of attraction of the "tighten" stage, and thus our theoretical guarantees fail in this case. From another point of view, our required minimum number of samples $n_{\min}$ defined in (4.1) indicates that, to make our theory valid, the eigengap $\theta = \lambda_1 - \lambda_2$ should be at least larger than the order of $s^*\sqrt{\log d/n}$, which is obtained by solving $\lambda_1 - \lambda_2$ in (4.1). In other words, our theory is established in a regime where the eigengap is sufficiently large, and therefore doesn't conflict with such a computational lower bound. Nevertheless, we argue that, for real-world applications the settings where the eigengap is sufficiently large are more common than the challenging setting considered in Berthet and Rigollet (2013a). For the more common settings, our framework attains the optimal $\sqrt{s^* \log d/n}$ statistical rate in polynomial time.

# 7 Numerical Results

We provide numerical results illustrating the computational efficiency and statistical accuracy of the proposed two-stage framework. First we present numerical results on synthetic data to back up our theory. Then we apply our framework on real-world datasets to illustrate its effectiveness.

## 7.1 Synthetic Data

First we use two numerical results to back up the statistical and computational characterization in Theorems 4.2 and 4.3. We then illustrate the minimax-optimal statistical rate of convergence of the final subspace estimator attained by our framework, which is characterized by Theorem 4.1. Finally, we compare our framework with several existing sparse PCA methods.

**Illustration of Theorems 4.2 and 4.3:** To show the necessity of a good initial estimator for the "tighten" stage, which is characterized in (4.10) of Theorem 4.2, we demonstrate the evolution of the estimation error $D(\mathcal{U}^*, \mathcal{U}^{(t)})$ in the SOAP algorithm (Algorithm 1) with different initializations. Particularly, we compare between the initial estimator obtained from the "relax" stage and random initialization.

We set $n = 150, d = 500, s^* = 10$ and $k = 5$. The top $k$ leading eigenvectors of $\mathbf{\Sigma}$ are generated as follows. First, we generate an $s^* \times k$ matrix whose entires are i.i.d. Gaussian with unit variance. We standardize this matrix with QR decomposition so that its columns are orthonormal. We then concatenate the $s^* \times k$ orthonormal matrix with a lower $(d - s^*) \times k$ zero matrix to obtain $\mathbf{U}^* \in \mathbb{R}^{d \times k}$. Similarly, the rest $d - k$ eigenvectors of $\mathbf{\Sigma}$ are obtained by first generating a $d \times (d-k)$ matrix with i.i.d. Gaussian entires with unit variance, then projecting its columns onto the subspace orthogonal to the column space of $\mathbf{U}^*$, and finally standardizing it with QR decomposition. To generate $\mathbf{\Sigma}$, we set its first $k$ eigenvalues to be $\{5, \ldots, 5\}$ and the rest of its eigenvalues to be uniformly distributed in $[0.5, 1]$. Then we set $\mathbf{\Sigma} = \sum_{j=1}^{d} \lambda_j \mathbf{u}_j^* (\mathbf{u}_j^*)^T$. Finally, we draw each sample from the $d$-dimensional Gaussian distribution $\mathcal{N}_d(\mathbf{0}, \mathbf{\Sigma})$.

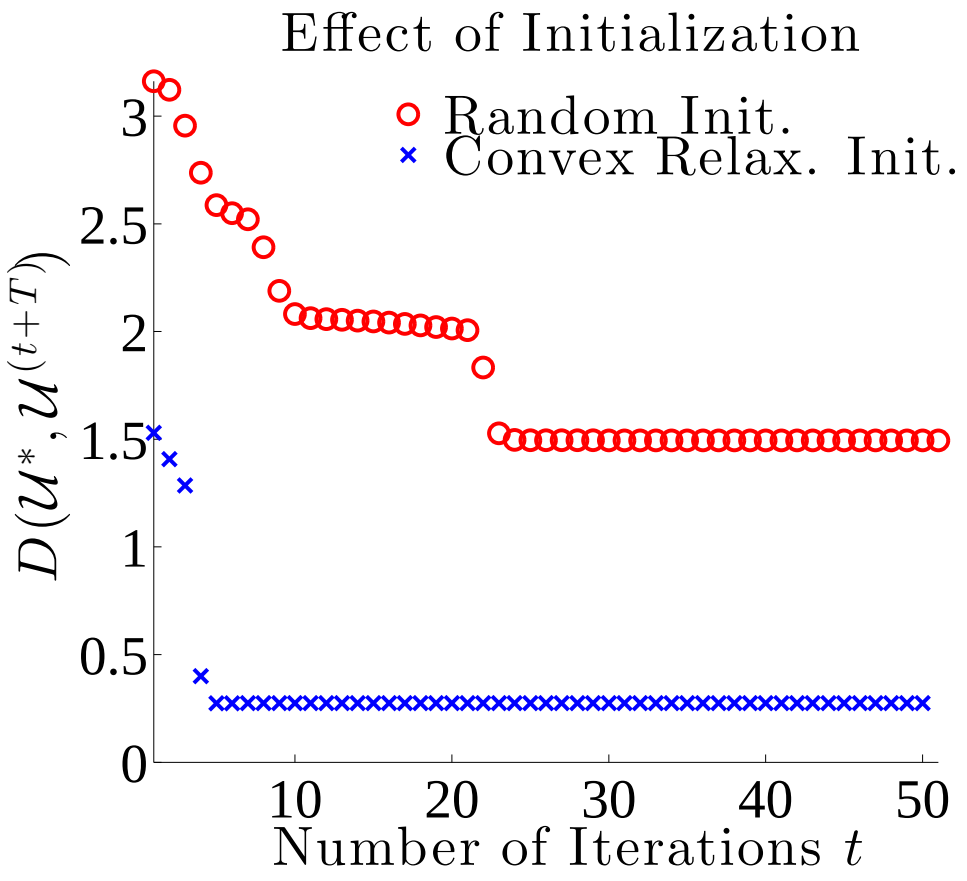

Figure 2: Effect of initialization: Circle and cross plots denote the estimation error at each iteration of SOAP (Algorithm 1). Circle plots correspond to random initialization, and cross plots correspond to the initialization obtained from the "relax" stage. As is shown here, with random initialization, SOAP might be trapped in a bad local solution with low estimation accuracy.

In Figure 2, we illustrate the evolution of $D(\mathcal{U}^*, \mathcal{U}^{(t+T)})$. Recall that $T$ denotes the total number of iterations of the "relax" stage, and thus the iteration index in SOAP is $t+T$ $(t=1,2,\ldots)$. As is shown in Figure 2, with random initialization, SOAP can be trapped in a bad local solution with unsatisfactory statistical performance. In contrast, with the initialization obtained from the "relax" stage, SOAP converges to a good local solution with high estimation accuracy. This suggests that a proper initialization that falls in the basin of attraction of SOAP is crucial for the estimation accuracy of the final estimator.

We demonstrate the results in Theorems 4.2 and 4.3 with the second experiment. The setting is the same as the first experiment, except that $n = 100$, $d = 200$, and the eigenvalues of $\mathbf{\Sigma}$ are set to $\{100, 100, 100, 100, 4, 1, \ldots, 1\}$. In Figure 3.(a) and Figure 3.(b), we illustrate the evolution of the estimation error $D(\mathcal{U}^*, \mathcal{U}^{(t)})$ within each of the two stages. Figure 3.(c) illustrates the evolution of $D\big(\mathcal{U}^*, \mathcal{U}^{(t)}\big) - D\big(\mathcal{U}^*, \mathcal{U}^{(T+\widetilde{T})}\big)$ within the "tighten" stage. In detail, according to (4.11) of Theorem 4.2, when $T + \widetilde{T}$ is sufficiently large, $D\big(\mathcal{U}^*, \mathcal{U}^{(T+\widetilde{T})}\big)$ is roughly the same as statistical error. Hence, $D\big(\mathcal{U}^*, \mathcal{U}^{(t)}\big) - D\big(\mathcal{U}^*, \mathcal{U}^{(T+\widetilde{T})}\big)$ is roughly the optimization error at the "tighten" stage, which decays to zero at a geometric rate as described in Theorem 4.2. Meanwhile, Figure 3.(c) shows that, even with sufficiently many iterations, the iterative sequence within the "relax" stage still converges to a solution with a suboptimal statistical rate of convergence, as illustrated in Figure 1 and described in Theorem 4.3. However, we observe that, the intermediate solution along the solution path of the "relax" stage serves as a good initial estimator, which falls into the basin of attraction of the "tighten" stage.

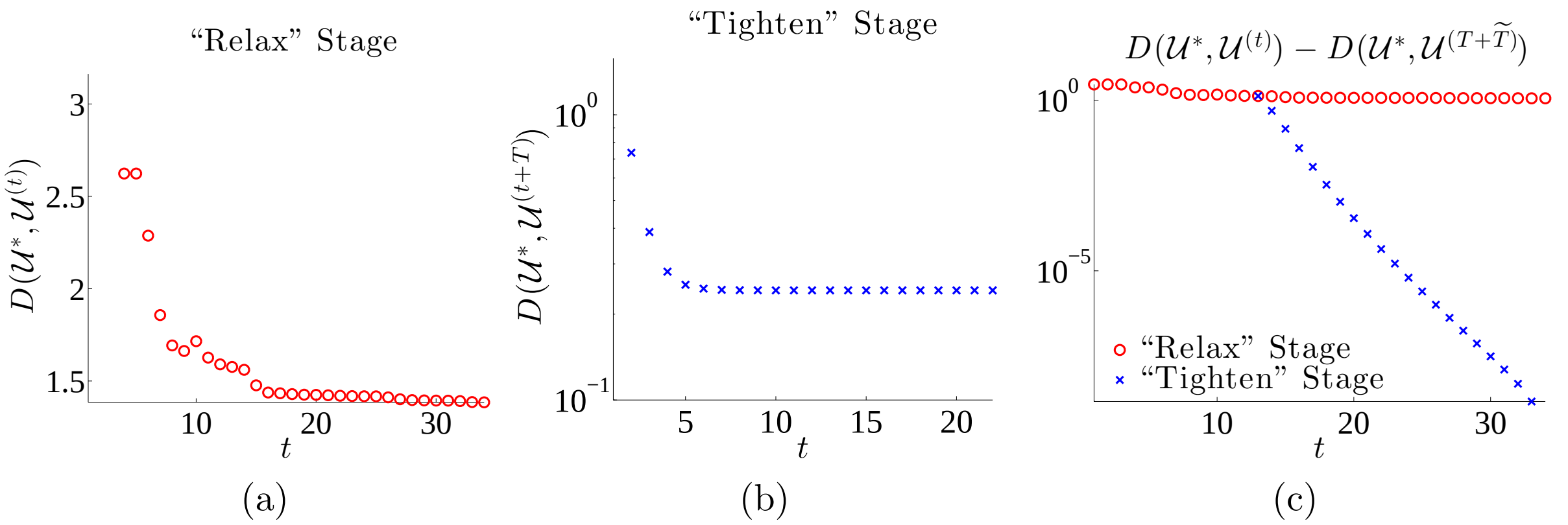

Figure 3: Evolution of estimation error within each stage: The circle plots denote the "relax" stage, while the cross plots denote the "tighten" stage. Here (b) and (c) are in log-scale. In particular, (c) illustrates the geometric decay of optimization error within the "tighten" stage, as well as the basin of attraction phenomenon shown in Figure 1.

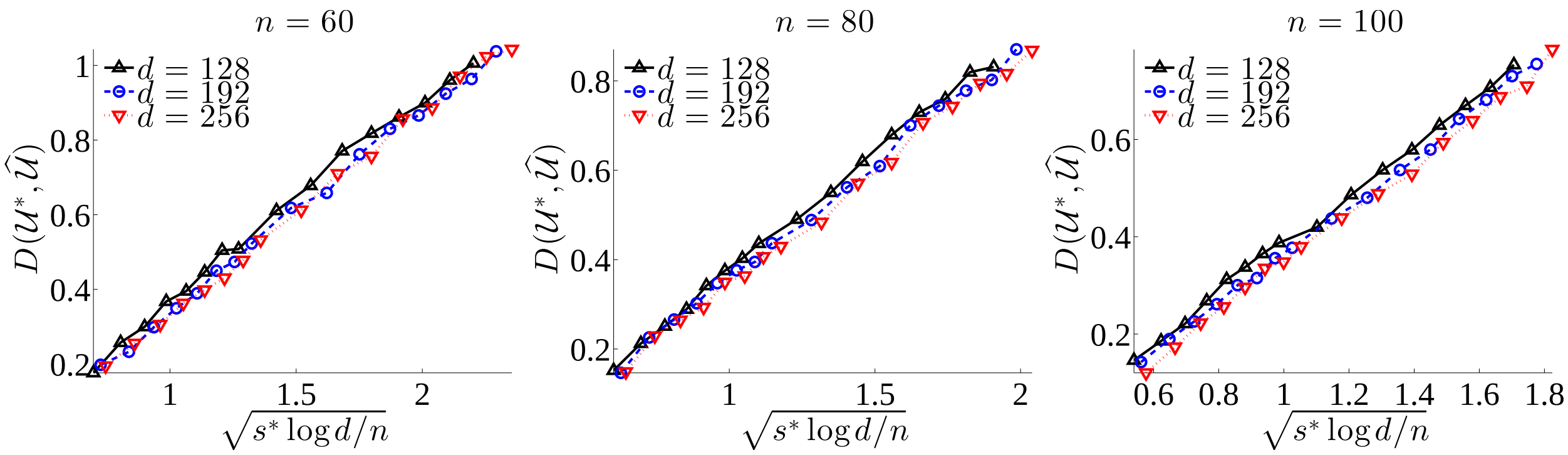

Figure 4: Estimation error $D\big(\mathcal{U}^*, \widehat{\mathcal{U}}\big)$ plotted against $\sqrt{s^* \log d/n}$ with fixed $n = 60, 80, 100$, $d = 128, 192, 256$ and varying $s^*$. The curves with different $d$ align, which suggests that the $\sqrt{\log d}$ part in our statistical rate is sharp. Meanwhile, the curves are almost linear in $\sqrt{s^*}$, which backs up the $\sqrt{s^*}$ dependency in our statistical rate.

**Optimal Statistical Rate of Convergence:** We illustrate how the estimation error $D(\mathcal{U}^*, \widehat{\mathcal{U}})$ of the final estimator $\widehat{\mathcal{U}} = \mathcal{U}^{(T+\widetilde{T})}$ scales with the sparsity level $s^*$, dimension $d$ and sample size $n$. The detailed setting is the same as the first experiment, except that $k = 4$ and the eigenvalues of $\mathbf{\Sigma}$ are set to $\{100, 50, 30, 15, 1, \ldots, 1\}$. We plot $D(\mathcal{U}^*, \widehat{\mathcal{U}})$ against $\sqrt{s^* \log d / n}$ with fixed $n$, $s^*$ and $d$ correspondingly in Figures 4-6, and we summarize these results in Figure 7. It is shown that, the curves of estimation error plotted against the rescaled axes of $\sqrt{s^* \log d / n}$ all stack up. Thus, our optimal statistical rate $\sqrt{s^* \log d / n}$ is sharp, in the sense that the same value of $\sqrt{s^* \log d / n}$ leads to similar estimation error in practice. Particularly, Figure 4 verifies that the statistical rate of our final estimator enjoys the optimal $\sqrt{s^*}$ dependency on the sparsity level. In contrast, the solution to the convex relaxation only has a suboptimal $s^*$ dependency on the sparsity level.

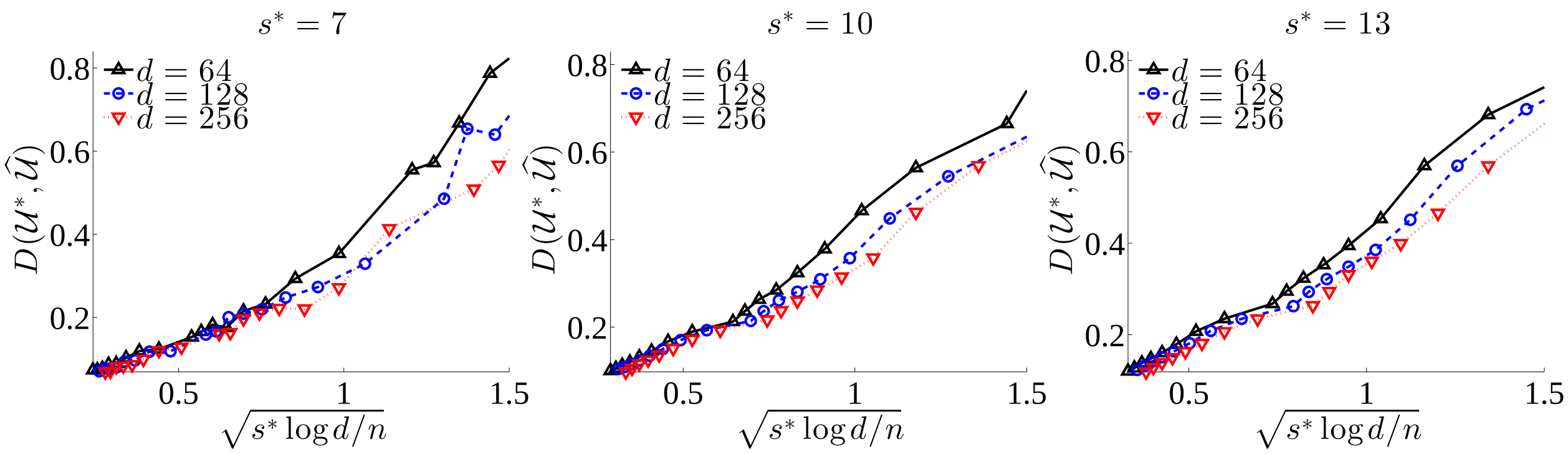

Figure 5: Estimation error $D(\mathcal{U}^*, \widehat{\mathcal{U}})$ plotted against $\sqrt{s^* \log d / n}$ with fixed $s^* = 7, 10, 13$, $d = 64, 128, 256$ and varying $n$. The curves with different $d$ align, which indicates that the $\sqrt{\log d}$ part in our statistical rate is sharp. Meanwhile, the curves are almost linear in $\sqrt{1/n}$, which backs up the $\sqrt{1/n}$ dependency in our statistical rate.

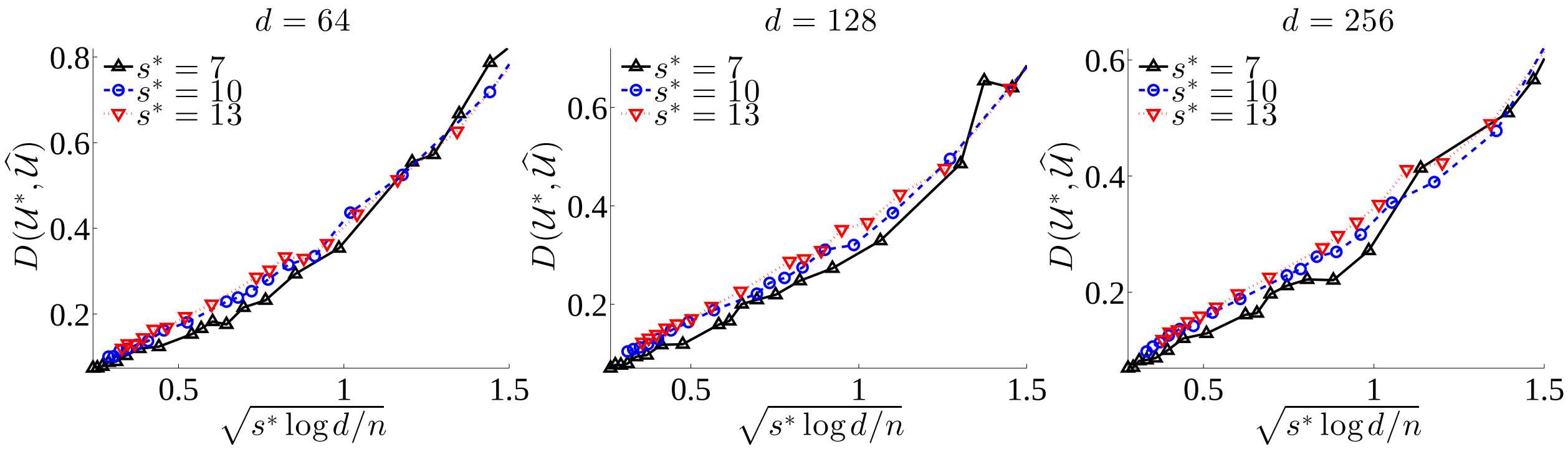

Figure 6: Estimation error $D(\mathcal{U}^*, \widehat{\mathcal{U}})$ plotted against $\sqrt{s^* \log d / n}$ with fixed $s^* = 7, 10, 13$, $d = 64, 128, 256$ and varying $n$. The curves with different $s^*$ stack up, which verifies that the $\sqrt{s^*}$ part in our statistical rate is sharp. Besides, the curves are almost linear in $\sqrt{1/n}$, which backs up the $\sqrt{1/n}$ dependency in our statistical rate.

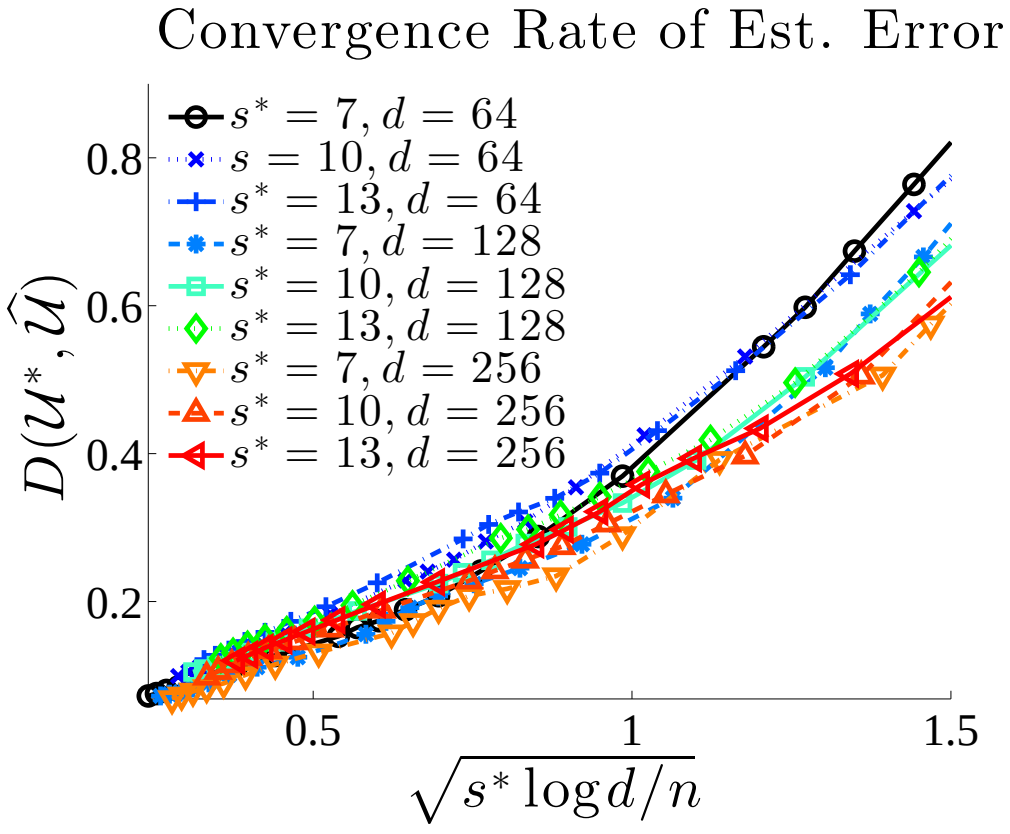

Figure 7: Estimation error $D(\mathcal{U}^*, \widehat{\mathcal{U}})$ plotted against $\sqrt{s^* \log d / n}$. The estimation error with fixed $d = 64, 128, 256$, $s^* = 7, 10, 13$ and varying $n$. The curves with different $s^*$ and $d$ all stack up. Thus the $\sqrt{s^* \log d}$ part in our statistical rate is sharp. Besides, the curves increase almost linearly with $\sqrt{1/n}$, which backs up the $\sqrt{1/n}$ dependency in our statistical rate.

**Comparison with Existing Methods:** We compare the proposed framework with several existing sparse PCA methods on estimation accuracy. These methods include convex relaxation (Vu et al., 2013), greedy method (PathSPCA) (d'Aspremont et al., 2008), truncated power method (TPower) (Yuan and Zhang, 2013) and generalized power method (GPower) (Journée et al., 2010). We employ the deflation method (Mackey, 2009) to enable subspace estimation for the other three methods.

Table 1: Comparison of the subspace estimation error $D(\mathcal{U}^*, \widehat{\mathcal{U}})$ between our framework and several existing sparse PCA methods. The detailed settings are as follows.
Setting (i): $d = 200$, $s^* = 10$, $k = 5$, $n = 50$, eigenvalues of $\boldsymbol{\Sigma}$ are $\{100, 100, 100, 100, 4, 1, \ldots, 1\}$;
Setting (ii): $d = 200$, $s^* = 10$, $k = 5$, $n = 100$, eigenvalues of $\boldsymbol{\Sigma}$ are $\{300, 240, 180, 120, 60, 1, \ldots, 1\}$.
Every experiment is repeated 50 times. The corresponding variance is provided in the parenthesis.

| Method | $D(\mathcal{U}^*, \widehat{\mathcal{U}})$ | |
|---|---|---|
| | Setting (i) | Setting (ii) |
| **Our Framework** | **0.32** (0.0067) | **0.064** (0.00016) |
| Convex Relaxation | 1.62 (0.0398) | 0.57 (0.021) |
| TPower + Deflation | 1.15 (0.1336) | 0.10 (0.00042) |
| GPower + Deflation | 1.84 (0.0226) | 1.75 (0.029) |
| PathSPCA + Deflation | 0.96 (0.0983) | 0.10 (0.00039) |

The detailed experiment settings are provided in Table 1. Note that setting (i) is more challenging than setting (ii). Because the top $k$ eigenvalues of $\boldsymbol{\Sigma}$ in setting (i) are not distinct, and the eigengap and sample size are smaller. As is shown in Table 1, our framework outperforms other methods on estimation accuracy in both settings. In particular, in setting (i) the subspace estimation error of our framework is only 30% of that of PathSPCA, the most accurate one among the rest methods. This is because, for non-distinctive leading eigenvalues the deflation method has both consistency and orthogonality issues. In contrast, our method directly estimates the principal subspace instead of individual eigenvectors, and avoids these problems. Remind that the convex relaxation method also directly estimates subspace. However, it only attains a suboptimal statistical rate and is less accurate in practice.

## 7.2 Real Data

In this section, we present numerical results on both equity data and biological data.

### 7.2.1 Equity Data

We apply the proposed framework on a dataset that contains the daily closing price of 452 stocks in the S&P 500 index between January 1, 2003 and January 1, 2008 from Yahoo Finance. There are totally 1258 samples, forming a matrix $\mathbf{P} = [\mathbf{P}_{j,i}]$, where $\mathbf{P}_{j,i}$ is the closing price of stock $j$ on day $i$. For data preprocessing, we calculate the log-ratio of the price at day $i+1$ to price at day $i$:

$$\mathbf{X}_{j,i} = \log(\mathbf{P}_{j,i+1}/\mathbf{P}_{j,i}), \quad i = 1, \ldots, 1257.$$

The sample covariance is calculated by $\widehat{\boldsymbol{\Sigma}} = 1/1257 \cdot \sum_{i=1}^{1257} \mathbf{X}_{*,i}\mathbf{X}_{*,i}^T$, where $\mathbf{X}_{*,i}$ is the $i$-th column of $\mathbf{X} = [\mathbf{X}_{j,i}]$. These stocks are categorized into 10 Global Industry Classification Standard (GICS) sectors, including Financials (74 stocks), Consumer Discretionary (70 stocks), Information Technology (64 stocks), Industrials (59 stocks), Health Care (46 stocks), Energy (37 stocks), Consumer Staples (35 stocks), Utilities (32 stocks), Materials (29 stocks), and Telecommunications Services (6 stocks). As we can expect, the correlation between stocks within the same sector is higher than the correlation between stocks across different sectors. Thus, the support of the sparse principal subspace tends to contain the stocks from the same sector.

The stocks contained in the support of the estimated principal subspace are shown in Table 2. That is to say, this combination of 36 stocks has the most explanatory power of the variance of the total 452 stocks. Let $\mathrm{Tr}\big(\widehat{\mathbf{U}}^T\widehat{\boldsymbol{\Sigma}}\widehat{\mathbf{U}}\big)$ be the variance explained by the subspace estimator, and $\mathrm{Tr}\big(\widehat{\boldsymbol{\Sigma}}\big)$ be the total variance. Our calculation shows, the estimated principal subspace explains 9.57% of the total variance. Moreover, as shown in Table 2, our framework successfully groups stocks from the same sector (mainly from Information Technology).

### 7.2.2 Biological Data

We evaluate our framework on the Colon Cancer data from Alon et al. (1999) and Lymphobia data from Alizadeh et al. (2000). Following the same experiment setup of Yuan and Zhang (2013) and d'Aspremont et al. (2008), we focus on the top 500 genes with the largest variance. Thus, the sample covariance matrix of both datasets are 500 by 500. In Table 3, we compare between the 20 genes contained in the support of our estimated principal subspace and the genes selected by PathSPCA (d'Aspremont et al., 2008) and TPower (Yuan and Zhang, 2013). Meanwhile, in Figure

Table 2: Equity data: We illustrate the stocks contained in the support of the estimated principal subspace and their sectors. Here $d = 452, n = 1257, \widehat{s} = 36$ and $k = 3$. We observe that the support of the estimated principal subspace mainly consists of stocks from Information Technology and some closely related stocks such as Priceline.com Inc and E-Trade.

| Stock Name | Sector | Full Name |
|---|---|---|
| AMD | Information Technology | Advanced Micro Devices |
| AKAM | Information Technology | Akamai Technologies Inc |
| ALTR | Information Technology | Altera Corp |
| ADI | Information Technology | Analog Devices Inc |
| AMAT | Information Technology | Applied Materials Inc |
| BRCM | Information Technology | Broadcom Corporation |
| FFIV | Information Technology | F5 Networks |
| JDSU | Information Technology | JDS Uniphase Corp. |
| JNPR | Information Technology | Juniper Networks |
| KLAC | Information Technology | KLA-Tencor Corp. |
| LSI | Information Technology | LSI Corporation |
| WFR | Information Technology | MEMC Electronic Materials |
| MCHP | Information Technology | Microchip Technology |
| MU | Information Technology | Micron Technology |
| MSFT | Information Technology | Microsoft Corp. |
| NSM | Information Technology | National Semiconductor |
| NTAP | Information Technology | NetApp |
| NFLX | Information Technology | NetFlix Inc. |
| NVLS | Information Technology | Novellus Systems |
| NVDA | Information Technology | Nvidia Corporation |
| SNDK | Information Technology | SanDisk Corporation |
| TLAB | Information Technology | Tellabs Inc. |
| TER | Information Technology | Teradyne Inc. |
| TXN | Information Technology | Texas Instruments |
| VRSN | Information Technology | Verisign Inc. |
| WDC | Information Technology | Western Digital |
| XLNX | Information Technology | Xilinx Inc |
| GLW | Industrials | Corning Inc. |
| MWW | Industrials | Monster Worldwide |
| PCLN | Industrials | Priceline.com Inc |
| PWR | Industrials | Quanta Services Inc. |
| AKS | Materials | AK Steel Hldg Corp |
| ATI | Materials | Allegheny Technologies Inc |
| TIE | Materials | Titanium Metals Corp |
| ETFC | Financials | E-Trade |
| URBN | Consumer Discretionary | Urban Outfitters |

8 we plot the proportion of the explained variance, i.e., $\mathrm{Tr}\big(\widehat{\mathbf{U}}^T\widehat{\boldsymbol{\Sigma}}\widehat{\mathbf{U}}\big)/\,\mathrm{Tr}\big(\widehat{\boldsymbol{\Sigma}}\big)$, against the cardinality of the estimated subspace's support.

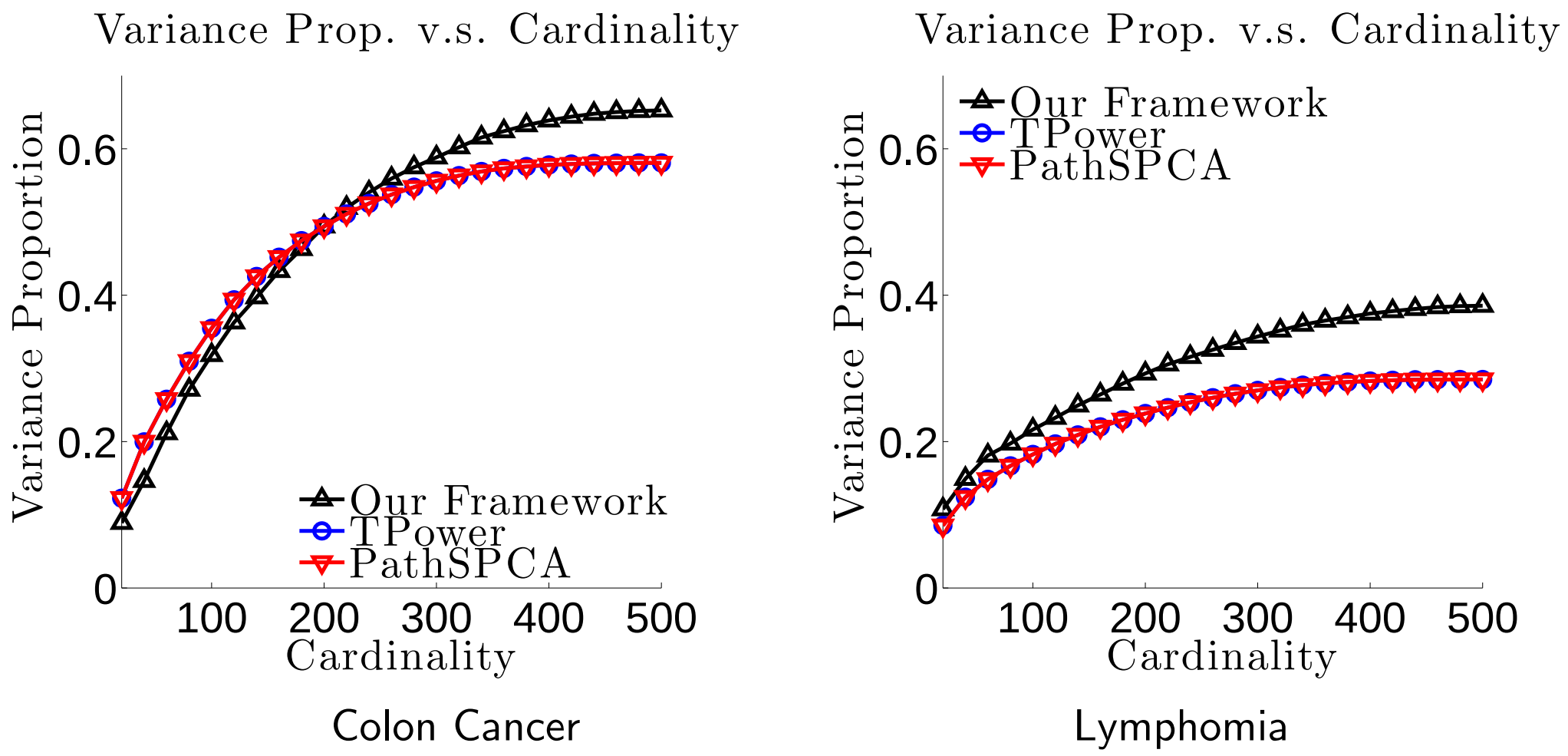

Figure 8: Variance proportion versus cardinality: The variance proportion denotes the proportion of the variance explained by the estimated principal subspace to the total variance of all genes, i.e., $\mathrm{Tr}\big(\widehat{\mathbf{U}}^T\widehat{\boldsymbol{\Sigma}}\widehat{\mathbf{U}}\big)/\,\mathrm{Tr}\big(\widehat{\boldsymbol{\Sigma}}\big)$. Here we have $k = 2$ and $d = 500$.

# 8 Conclusion

In this paper, we propose a two-stage computational framework for sparse PCA. Within the "relax" stage, we approximately solve a convex relaxation of sparse PCA. Within the "tighten" stage, we employ the proposed SOAP algorithm to iteratively refine the initial estimator obtained from the "relax" stage. Our theory proves that the final principal subspace estimator attains the minimax-optimal statistical rate of convergence. Moreover, our analysis sharply characterizes the computational and statistical properties of the solution path within the two stages. In particular, we demonstrate the basin of attraction and blessing of large sample size phenomena under this framework. The "tighten after relax" strategy proposed in this paper motivates a general paradigm for solving the statistical problems that involve nonconvex optimization with provable guarantees.

# Acknowledgement

We thank Tengyu Ma and Philippe Rigollet for very constructive discussions. We are also grateful for Jianqing Fan and Ramon van Handel for providing many helpful feedbacks on this work.

Table 3: Comparison of the top 20 genes selected by our framework, PathSPCA and TPower. The common genes selected by all three methods are bolded. The indices of genes are the column indices in the data matrix of Colon Cancer and Lymphomia datasets provided in d'Aspremont et al. (2008).

| Colon Cancer | | | Lymphoma | | |
|---|---|---|---|---|---|
| **Our Framework** | TPower | PathSPCA | **Our Framework** | TPower | PathSPCA |
| **1** | **1** | **1** | 84 | 46 | 46 |
| **2** | **2** | **2** | **85** | 47 | **85** |
| **3** | **3** | **3** | **86** | **85** | **86** |
| **6** | **6** | **6** | **87** | **86** | **87** |
| **11** | **11** | **11** | **88** | **87** | **88** |
| **15** | 13 | 13 | 90 | **88** | **424** |
| 19 | **15** | **15** | 91 | **424** | **443** |
| 21 | 17 | 17 | 92 | **443** | 444 |
| 28 | 23 | 23 | **424** | 446 | 446 |
| **32** | **32** | **32** | **443** | **447** | **447** |
| **33** | **33** | **33** | **447** | 448 | 448 |
| **38** | **38** | **38** | **452** | **452** | **452** |
| **41** | **41** | **41** | **453** | **453** | **453** |
| 43 | 52 | 52 | **454** | **454** | **454** |
| 61 | **64** | **64** | **455** | **455** | **455** |
| **64** | **82** | **82** | **456** | **456** | **456** |
| **82** | 106 | 106 | **457** | **457** | **457** |
| 83 | 110 | 110 | **458** | **458** | **458** |
| 98 | 123 | 116 | **459** | **459** | **459** |
| 109 | 243 | 123 | **488** | **488** | **488** |

# A Derivations of Algorithms 4 and 5

To obtain $\mathbf{\Pi}^{(t+1)}$ and $\mathbf{\Phi}^{(t+1)}$, we need to solve the two subproblems shown in Line 6 and Line 7 of Algorithm 3. The subproblem in Line 6 can be reformulated as

$$\begin{aligned}\mathbf{\Pi}^{(t+1)} \leftarrow \operatorname{argmin}&\Big\{-\langle\widehat{\mathbf{\Sigma}},\mathbf{\Pi}\rangle+\rho\big\|\mathbf{\Phi}^{(t)}\big\|_{1,1}-\big\langle\mathbf{\Theta}^{(t)},\mathbf{\Pi}-\mathbf{\Phi}^{(t)}\big\rangle+\frac{\beta}{2}\big\|\mathbf{\Pi}-\mathbf{\Phi}^{(t)}\big\|_{\mathrm{F}}^2 \;\Big|\; \mathbf{\Pi}\in\mathcal{A}\Big\}\\ =\operatorname{argmin}&\Big\{-\langle\widehat{\mathbf{\Sigma}},\mathbf{\Pi}\rangle-\langle\mathbf{\Theta}^{(t)},\mathbf{\Pi}\rangle+\frac{\beta}{2}\big\|\mathbf{\Pi}-\mathbf{\Phi}^{(t)}\big\|_{\mathrm{F}}^2 \;\Big|\; \mathbf{\Pi}\in\mathcal{A}\Big\}\\ =\operatorname{argmin}&\Big\{\frac{\beta}{2}\big\|\mathbf{\Pi}-\big(\mathbf{\Phi}^{(t)}+\mathbf{\Theta}^{(t)}/\beta+\widehat{\mathbf{\Sigma}}/\beta\big)\big\|_{\mathrm{F}}^2 \;\Big|\; \mathbf{\Pi}\in\mathcal{A}\Big\}. \end{aligned} \tag{A.1}$$

The subproblem in (A.1) is equivalent to projecting $\mathbf{\Phi}^{(t)}-\mathbf{\Theta}^{(t)}+\widehat{\mathbf{\Sigma}}/\rho$ onto $\mathcal{A}$. Such a projection can be accomplished effectively using Algorithm 4, where the quadratic programming problem in Line 4 has some closed form solution as shown in Lemma 4.1 of Vu et al. (2013).

Meanwhile, the subproblem in Line 7 can be written as

$$\begin{aligned}\mathbf{\Phi}^{(t+1)} &\leftarrow \operatorname{argmin}\left\{-\langle\widehat{\mathbf{\Sigma}},\mathbf{\Pi}^{(t+1)}\rangle+\rho\|\mathbf{\Phi}\|_{1,1}-\langle\mathbf{\Theta}^{(t)},\mathbf{\Pi}^{(t+1)}-\mathbf{\Phi}\rangle+\frac{\beta}{2}\|\mathbf{\Pi}^{(t+1)}-\mathbf{\Phi}\|_{\mathrm{F}}^2 \;\middle|\; \mathbf{\Phi}\in\mathcal{B}\right\}\\ &= \operatorname{argmin}\left\{\rho\|\mathbf{\Phi}\|_{1,1}+\langle\mathbf{\Theta}^{(t)},\mathbf{\Phi}\rangle+\frac{\beta}{2}\|\mathbf{\Pi}^{(t+1)}-\mathbf{\Phi}\|_{\mathrm{F}}^2 \;\middle|\; \mathbf{\Phi}\in\mathbb{R}^{d\times d}\right\}\\ &= \operatorname{argmin}\left\{\rho\|\mathbf{\Phi}\|_{1,1}+\frac{\beta}{2}\|\mathbf{\Pi}^{(t+1)}-\mathbf{\Theta}^{(t)}/\beta-\mathbf{\Phi}\|_{\mathrm{F}}^2 \;\middle|\; \mathbf{\Phi}\in\mathbb{R}^{d\times d}\right\},\end{aligned} \tag{A.2}$$

where the solution to (A.2) can be obtained by entry-wise soft-thresholding as shown in Algorithm 5.

# B Proof for the "Tighten" Stage

We present the detailed proof of Theorem 4.2 and the corresponding auxiliary lemmas.

## B.1 Proof of Lemma 5.2

*Proof.* According to Lemma 4.2 of Vu and Lei (2013), we have

$$D\big[\mathcal{U}^*,\widehat{\mathcal{U}}(\mathcal{I})\big]^2 \le \frac{2\big\langle\mathbf{\Sigma},\mathbf{U}^*\cdot(\mathbf{U}^*)^T-\widehat{\mathbf{U}}(\mathcal{I})\cdot\widehat{\mathbf{U}}(\mathcal{I})^T\big\rangle}{\lambda_k-\lambda_{k+1}}. \tag{B.1}$$

Since the columns of $\widehat{\mathbf{U}}(\mathcal{I})$ are the top $k$ eigenvectors of $\widehat{\mathbf{\Sigma}}_{\mathcal{I}}$, we have

$$\operatorname{Tr}\Big[\widehat{\mathbf{U}}(\mathcal{I})^T\cdot\widehat{\mathbf{\Sigma}}_{\mathcal{I}}\cdot\widehat{\mathbf{U}}(\mathcal{I})\Big] \ge \operatorname{Tr}\Big[(\mathbf{U}^*)^T\cdot\widehat{\mathbf{\Sigma}}_{\mathcal{I}}\cdot\mathbf{U}^*\Big]. \tag{B.2}$$

Hence we obtain

$$\big\langle\widehat{\mathbf{\Sigma}},\widehat{\mathbf{U}}(\mathcal{I})\cdot\widehat{\mathbf{U}}(\mathcal{I})^T\big\rangle = \big\langle\widehat{\mathbf{\Sigma}}_{\mathcal{I}},\widehat{\mathbf{U}}(\mathcal{I})\cdot\widehat{\mathbf{U}}(\mathcal{I})^T\big\rangle \ge \big\langle\widehat{\mathbf{\Sigma}}_{\mathcal{I}},\mathbf{U}^*\cdot(\mathbf{U}^*)^T\big\rangle = \big\langle\widehat{\mathbf{\Sigma}},\mathbf{U}^*\cdot(\mathbf{U}^*)^T\big\rangle, \tag{B.3}$$

where the inequality follows from (B.2). The first equality in (B.3) follows from

$$\big\langle\widehat{\mathbf{\Sigma}}_{\mathcal{I}},\widehat{\mathbf{U}}(\mathcal{I})\cdot\widehat{\mathbf{U}}(\mathcal{I})^T\big\rangle = \operatorname{Tr}\Big[\widehat{\mathbf{U}}(\mathcal{I})^T\cdot\widehat{\mathbf{\Sigma}}_{\mathcal{I}}\cdot\widehat{\mathbf{U}}(\mathcal{I})\Big] = \operatorname{Tr}\Big[\widehat{\mathbf{U}}(\mathcal{I})^T\cdot\widehat{\mathbf{\Sigma}}\cdot\widehat{\mathbf{U}}(\mathcal{I})\Big] = \big\langle\widehat{\mathbf{\Sigma}},\widehat{\mathbf{U}}(\mathcal{I})\cdot\widehat{\mathbf{U}}(\mathcal{I})^T\big\rangle.$$

Here the second equality is due to the row-sparsity of $\widehat{\mathbf{U}}(\mathcal{I})$. The last equality in (B.3) holds for the same reason, since the row-support of $\mathbf{U}^*$ is $\mathcal{S}^*$, which satisfies $\mathcal{S}^*\subseteq\mathcal{I}$. From (B.3) we have

$$0 \le \frac{-\big\langle\widehat{\mathbf{\Sigma}},\mathbf{U}^*\cdot(\mathbf{U}^*)^T-\widehat{\mathbf{U}}(\mathcal{I})\cdot\widehat{\mathbf{U}}(\mathcal{I})^T\big\rangle}{\lambda_k-\lambda_{k+1}}. \tag{B.4}$$

Combining (B.1) and (B.4), we obtain

$$D\big[\mathcal{U}^*,\widehat{\mathcal{U}}(\mathcal{I})\big]^2 \le \frac{2\Big\langle\mathbf{\Sigma}-\widehat{\mathbf{\Sigma}},\mathbf{U}^*\cdot(\mathbf{U}^*)^T-\widehat{\mathbf{U}}(\mathcal{I})\cdot\widehat{\mathbf{U}}(\mathcal{I})^T\Big\rangle}{\lambda_k-\lambda_{k+1}}. \tag{B.5}$$

Now we provide an upper bound for the numerator on the right-hand side of (B.5) following the proof strategy for Theorem 3.4 in Vu and Lei (2013). The major difference is that, the row-sparsity level of $\widehat{\mathbf{U}}(\mathcal{I})$ is $\|\widehat{\mathbf{U}}(\mathcal{I})\|_{2,0} = |\mathcal{I}|$, while their corresponding row-sparsity level is $s^*$. We will sketch the whole proof and point out the detailed differences. We denote the projection matrices $\mathbf{U}^* \cdot (\mathbf{U}^*)^T$ and $\widehat{\mathbf{U}}(\mathcal{I}) \cdot \widehat{\mathbf{U}}(\mathcal{I})^T$ to be $\mathbf{\Pi}^*$ and $\widehat{\mathbf{\Pi}}(\mathcal{I})$. Also, we define $\mathbf{W} = \widehat{\mathbf{\Sigma}} - \mathbf{\Sigma}$. Note that

$$\big\langle \mathbf{\Sigma} - \widehat{\mathbf{\Sigma}}, \mathbf{U}^* \cdot (\mathbf{U}^*)^T - \widehat{\mathbf{U}}(\mathcal{I}) \cdot \widehat{\mathbf{U}}(\mathcal{I})^T \big\rangle = \big\langle \mathbf{W}, \widehat{\mathbf{U}}(\mathcal{I}) \cdot \widehat{\mathbf{U}}(\mathcal{I})^T - \mathbf{U}^* \cdot (\mathbf{U}^*)^T \big\rangle.$$

Then by Proposition C.1 of Vu and Lei (2013), we have

$$\begin{aligned} &\big\langle \mathbf{\Sigma} - \widehat{\mathbf{\Sigma}}, \mathbf{U}^* \cdot (\mathbf{U}^*)^T - \widehat{\mathbf{U}}(\mathcal{I}) \cdot \widehat{\mathbf{U}}(\mathcal{I})^T \big\rangle \\ &= \underbrace{-\big\langle \mathbf{W}, \mathbf{\Pi}^* \cdot \big[\mathbf{I}_d - \widehat{\mathbf{\Pi}}(\mathcal{I})\big] \cdot \mathbf{\Pi}^* \big\rangle}_{\text{(i)}} + 2 \underbrace{\big\langle \mathbf{W}, [\mathbf{I}_d - \mathbf{\Pi}^*] \cdot \widehat{\mathbf{\Pi}}(\mathcal{I}) \cdot \mathbf{\Pi}^* \big\rangle}_{\text{(ii)}} + \underbrace{\big\langle \mathbf{W}, [\mathbf{I}_d - \mathbf{\Pi}^*] \cdot \widehat{\mathbf{\Pi}}(\mathcal{I}) \cdot [\mathbf{I}_d - \mathbf{\Pi}^*] \big\rangle}_{\text{(iii)}}. \end{aligned} \tag{B.6}$$

Now we provide upper bounds for the terms in (B.6) correspondingly:

- For term (i), following exactly the same proof of Theorem 3.4 in Vu and Lei (2013), we obtain

$$\begin{aligned} -\big\langle \mathbf{W}, \mathbf{\Pi}^* \cdot \big[\mathbf{I}_d - \widehat{\mathbf{\Pi}}(\mathcal{I})\big] \cdot \mathbf{\Pi}^* \big\rangle &\le \Big| \big\langle \mathbf{W}, \mathbf{\Pi}^* \cdot \big[\mathbf{I}_d - \widehat{\mathbf{\Pi}}(\mathcal{I})\big] \cdot \mathbf{\Pi}^* \big\rangle \Big| \\ &\le C \cdot \lambda_1 \cdot \log n \cdot \sqrt{\frac{k}{n}} \cdot D\big[\mathcal{U}^*, \widehat{\mathcal{U}}(\mathcal{I})\big]^2 \end{aligned}$$

  with probability at least $1 - (n-1)^{-1}$.

- For term (ii), we have

$$\begin{aligned} \big\langle \mathbf{W}, [\mathbf{I}_d - \mathbf{\Pi}^*] \cdot \widehat{\mathbf{\Pi}}(\mathcal{I}) \cdot \mathbf{\Pi}^* \big\rangle &= \mathrm{Tr}\Big[\mathbf{W} \cdot (\mathbf{I}_d - \mathbf{\Pi}^*) \cdot \widehat{\mathbf{\Pi}}(\mathcal{I}) \cdot \mathbf{\Pi}^*\Big] \\ &= \mathrm{Tr}\Big[\mathbf{\Pi}^* \cdot \mathbf{W} \cdot (\mathbf{I}_d - \mathbf{\Pi}^*) \cdot (\mathbf{I}_d - \mathbf{\Pi}^*) \cdot \widehat{\mathbf{\Pi}}(\mathcal{I})\Big] \\ &= \big\langle (\mathbf{I}_d - \mathbf{\Pi}^*) \cdot \mathbf{W} \cdot \mathbf{\Pi}^*, (\mathbf{I}_d - \mathbf{\Pi}^*) \cdot \widehat{\mathbf{\Pi}}(\mathcal{I}) \big\rangle, \end{aligned} \tag{B.7}$$

  where the second equality follows from $(\mathbf{I}_d - \mathbf{\Pi}^*) \cdot (\mathbf{I}_d - \mathbf{\Pi}^*) = \mathbf{I}_d - \mathbf{\Pi}^*$ since $\mathbf{\Pi}^*$ is a projection matrix. Applying Hölder's inequality to the right-hand side of (B.7), we further obtain

$$\big\langle \mathbf{W}, (\mathbf{I}_d - \mathbf{\Pi}^*) \cdot \widehat{\mathbf{\Pi}}(\mathcal{I}) \cdot \mathbf{\Pi}^* \big\rangle \le \big\| (\mathbf{I}_d - \mathbf{\Pi}^*) \cdot \mathbf{W} \cdot \mathbf{\Pi}^* \big\|_{2,\infty} \big\| (\mathbf{I}_d - \mathbf{\Pi}^*) \cdot \widehat{\mathbf{\Pi}}(\mathcal{I}) \big\|_{2,1}, \tag{B.8}$$

  where $\|\cdot\|_{2,\infty}$ and $\|\cdot\|_{2,1}$ are defined in the notation part of §1. First, for $\|(\mathbf{I}_d - \mathbf{\Pi}^*) \cdot \widehat{\mathbf{\Pi}}(\mathcal{I})\|_{2,1}$ in (B.8) we have

$$\big\| (\mathbf{I}_d - \mathbf{\Pi}^*) \cdot \widehat{\mathbf{\Pi}}(\mathcal{I}) \big\|_{2,1} \le \sqrt{\big\| (\mathbf{I}_d - \mathbf{\Pi}^*) \cdot \widehat{\mathbf{\Pi}}(\mathcal{I}) \big\|_{2,0}} \cdot \big\| (\mathbf{I}_d - \mathbf{\Pi}^*) \cdot \widehat{\mathbf{\Pi}}(\mathcal{I}) \big\|_F,$$

  where $\|(\mathbf{I}_d - \mathbf{\Pi}^*) \cdot \widehat{\mathbf{\Pi}}(\mathcal{I})\|_F = D\big[\mathcal{U}^*, \widehat{\mathcal{U}}(\mathcal{I})\big] / \sqrt{2}$ by Lemma 2.1. For $\|(\mathbf{I}_d - \mathbf{\Pi}^*) \cdot \widehat{\mathbf{\Pi}}(\mathcal{I})\|_{2,0}$ we have

$$\big\| (\mathbf{I}_d - \mathbf{\Pi}^*) \cdot \widehat{\mathbf{\Pi}}(\mathcal{I}) \big\|_{2,0} = \big\| (\mathbf{I}_d - \mathbf{\Pi}^*) \cdot \widehat{\mathbf{U}}(\mathcal{I}) \cdot \widehat{\mathbf{U}}(\mathcal{I})^T \big\|_{2,0} = \big\| (\mathbf{I}_d - \mathbf{\Pi}^*) \cdot \widehat{\mathbf{U}}(\mathcal{I}) \big\|_{2,0}. \tag{B.9}$$

Here in the last equality we use the fact that right-multiplying a matrix with the transpose of an orthonormal matrix doesn't change the $\ell_2$ norm of its each row. From (B.9) we have

$$\big\|(\mathbf{I}_d - \mathbf{\Pi}^*)\cdot\widehat{\mathbf{\Pi}}(\mathcal{I})\big\|_{2,0} = \big\|(\mathbf{I}_d - \mathbf{\Pi}^*)\cdot\widehat{\mathbf{U}}(\mathcal{I})\big\|_{2,0} \le \big\|\widehat{\mathbf{U}}(\mathcal{I})\big\|_{2,0} + \big\|\mathbf{U}^*\cdot(\mathbf{U}^*)^T\cdot\widehat{\mathbf{U}}(\mathcal{I})\big\|_{2,0},$$

where we use the fact that $\mathbf{\Pi}^* = \mathbf{U}^*\cdot(\mathbf{U}^*)^T$. Note that $(\mathbf{U}^*)^T\cdot\widehat{\mathbf{U}}(\mathcal{I})$ is a $k\times k$ matrix. Hence each column of $\mathbf{U}^*\cdot(\mathbf{U}^*)^T\cdot\widehat{\mathbf{U}}(\mathcal{I})$ is a linear combination of the columns of $\mathbf{U}^*$. Therefore, the number of nonzero rows of $\mathbf{U}^*\cdot(\mathbf{U}^*)^T\cdot\widehat{\mathbf{U}}(\mathcal{I})$ is upper bounded by that of $\mathbf{U}^*$, i.e.,

$$\big\|(\mathbf{I}_d - \mathbf{\Pi}^*)\cdot\widehat{\mathbf{\Pi}}(\mathcal{I})\big\|_{2,0} \le \big\|\widehat{\mathbf{U}}(\mathcal{I})\big\|_{2,0} + \big\|\mathbf{U}^*\big\|_{2,0} \le |\mathcal{I}| + s^* \le 2|\mathcal{I}|,$$

where the last inequality follows from $\mathcal{S}^*\subseteq\mathcal{I}$. Hence, we obtain

$$\big\|(\mathbf{I}_d - \mathbf{\Pi}^*)\cdot\widehat{\mathbf{\Pi}}(\mathcal{I})\big\|_{2,1} \le \sqrt{|\mathcal{I}|}\cdot D\big[\mathcal{U}^*,\widehat{\mathcal{U}}(\mathcal{I})\big].$$

Second, for $\big\|(\mathbf{I}_d - \mathbf{\Pi}^*)\cdot\mathbf{W}\cdot\mathbf{\Pi}^*\big\|_{2,\infty}$ in (B.8), following exactly the same proof of Theorem 3.4 in Vu and Lei (2013), we obtain

$$\big\|(\mathbf{I}_d - \mathbf{\Pi}^*)\cdot\mathbf{W}\cdot\mathbf{\Pi}^*\big\|_{2,\infty} \le C\sqrt{\lambda_1\lambda_{k+1}}\cdot\sqrt{\frac{k+\log d}{n}}$$

with probability at least $1-1/d$.

Thus for term (ii) in (B.6), we have

$$\big\langle \mathbf{W}, (\mathbf{I}_d - \mathbf{\Pi}^*)\cdot\widehat{\mathbf{\Pi}}(\mathcal{I})\cdot\mathbf{\Pi}^*\big\rangle \le C'\sqrt{\lambda_1\lambda_{k+1}}\cdot\sqrt{\frac{|\mathcal{I}|\cdot(k+\log d)}{n}}\cdot D\big[\mathcal{U}^*,\widehat{\mathcal{U}}(\mathcal{I})\big]$$

with the same probability.

- For term (iii), we have

$$\big\langle \mathbf{W}, (\mathbf{I}_d - \mathbf{\Pi}^*)\cdot\widehat{\mathbf{\Pi}}(\mathcal{I})\cdot(\mathbf{I}_d - \mathbf{\Pi}^*)\big\rangle = \big\langle \mathbf{W}, (\mathbf{I}_d - \mathbf{\Pi}^*)\cdot\widehat{\mathbf{U}}(\mathcal{I})\cdot\widehat{\mathbf{U}}(\mathcal{I})^T\cdot(\mathbf{I}_d - \mathbf{\Pi}^*)\big\rangle \tag{B.10}$$

because $\widehat{\mathbf{\Pi}}(\mathcal{I}) = \widehat{\mathbf{U}}(\mathcal{I})\cdot\widehat{\mathbf{U}}(\mathcal{I})^T$. It is nontrivial to provide a sharp upper bound for right-hand side. We consider the following quantity

$$\sup_{\mathbf{U}}\big\langle \mathbf{W}, (\mathbf{I}_d - \mathbf{\Pi}^*)\cdot\mathbf{U}\cdot\mathbf{U}^T\cdot(\mathbf{I}_d - \mathbf{\Pi}^*)\big\rangle \tag{B.11}$$

$$\text{subject to } \mathbf{U} \text{ orthonormal},\ \|\mathbf{U}\|_{2,0}\le|\mathcal{I}|,\ D\big[\mathcal{U}^*,\mathcal{U}\big]\le D\big[\mathcal{U}^*,\widehat{\mathcal{U}}(\mathcal{I})\big],$$

which is an upper bound of the right-hand side of (B.10).

Then we invoke the quadratic form empirical process argument in Lemma D.4 of Vu and Lei (2013), and replace the sparsity level $s^*$ therein by $|\mathcal{I}|$. It follows that (B.11) is upper bounded by

$$(\log n)^{5/2}\cdot\lambda_{k+1}\cdot\Bigg(C\sqrt{\frac{|\mathcal{I}|\cdot(k+\log d)}{n}}\cdot D\big[\mathcal{U}^*,\widehat{\mathcal{U}}(\mathcal{I})\big]^2 + C'\frac{|\mathcal{I}|\cdot(k+\log d)}{n}\cdot D\big[\mathcal{U}^*,\widehat{\mathcal{U}}(\mathcal{I})\big] + \bigg[\frac{|\mathcal{I}|\cdot(k+\log d)}{n}\bigg]^2\Bigg)$$

with probability at least $1-6\log n/n-3/n$.

Combining the upper bounds for the three terms on the right-hand side of (B.6) and plugging it back into (B.5) gives a quadratic inequality with respect to $D\big[\mathcal{U}^*,\widehat{\mathcal{U}}(\mathcal{I})\big]$. Suppose $n$ is sufficiently large, by solving such a quadratic inequality we obtain

$$D\big[\mathcal{U}^*,\widehat{\mathcal{U}}(\mathcal{I})\big] \leq C \cdot \frac{\sqrt{\lambda_1\lambda_{k+1}}}{\lambda_k - \lambda_{k+1}} \cdot \sqrt{\frac{|\mathcal{I}|\cdot(k+\log d)}{n}}$$

with probability at least $1 - 4/(n-1) - 1/d - 6\log n/n$, which concludes the proof. □

## B.2 Proof of Lemma 5.4

*Proof.* Because $\mathbf{U}^{(t)}$ is row-sparse and its row-support is a subset of $\mathcal{I}$, the update step in (5.4) is equivalent to

$$\widetilde{\mathbf{V}}^{(t+1)} \leftarrow \widehat{\boldsymbol{\Sigma}}_{\mathcal{I}} \cdot \mathbf{U}^{(t)},$$

which implies

$$\widehat{\boldsymbol{\Sigma}}_{\mathcal{I}} \cdot \mathbf{U}^{(t)} = \widetilde{\mathbf{V}}^{(t+1)} = \mathbf{V}^{(t+1)} \cdot \mathbf{R}_1^{(t+1)}, \tag{B.12}$$

where the second equality follows from the QR decomposition in (5.5).

We denote $\widehat{\mathbf{U}}(\mathcal{I})^{\perp}$ to be the orthonormal matrix whose columns span the subspace corresponding to $\lambda_{k+1}\big(\widehat{\boldsymbol{\Sigma}}_{\mathcal{I}}\big), \ldots, \lambda_d\big(\widehat{\boldsymbol{\Sigma}}_{\mathcal{I}}\big)$. Remind that $\widehat{\boldsymbol{\Sigma}}_{\mathcal{I}} \in \mathbb{R}^{d\times d}$ is the restriction of $\widehat{\boldsymbol{\Sigma}}$ on the rows and columns indexed by $\mathcal{I}$. Hence, both $\widehat{\mathbf{U}}(\mathcal{I})$ and $\widehat{\mathbf{U}}(\mathcal{I})^{\perp}$ are row-sparse with row-support $\mathcal{I}$. For convenience, we define

$$\Lambda_0\big(\widehat{\boldsymbol{\Sigma}}_{\mathcal{I}}\big) = \begin{bmatrix} \lambda_1\big(\widehat{\boldsymbol{\Sigma}}_{\mathcal{I}}\big) & \cdots & 0 \\ \vdots & \ddots & \vdots \\ 0 & \cdots & \lambda_k\big(\widehat{\boldsymbol{\Sigma}}_{\mathcal{I}}\big) \end{bmatrix}, \qquad \Lambda_1\big(\widehat{\boldsymbol{\Sigma}}_{\mathcal{I}}\big) = \begin{bmatrix} \lambda_{k+1}\big(\widehat{\boldsymbol{\Sigma}}_{\mathcal{I}}\big) & \cdots & 0 \\ \vdots & \ddots & \vdots \\ 0 & \cdots & \lambda_d\big(\widehat{\boldsymbol{\Sigma}}_{\mathcal{I}}\big) \end{bmatrix}.$$

Then eigenvalue decomposition gives

$$\widehat{\boldsymbol{\Sigma}}_{\mathcal{I}} = \Big[\widehat{\mathbf{U}}(\mathcal{I}), \widehat{\mathbf{U}}(\mathcal{I})^{\perp}\Big] \cdot \begin{bmatrix} \Lambda_0\big(\widehat{\boldsymbol{\Sigma}}_{\mathcal{I}}\big) & \mathbf{0} \\ \mathbf{0} & \Lambda_1\big(\widehat{\boldsymbol{\Sigma}}_{\mathcal{I}}\big) \end{bmatrix} \cdot \begin{bmatrix} \widehat{\mathbf{U}}(\mathcal{I})^T \\ \big[\widehat{\mathbf{U}}(\mathcal{I})^{\perp}\big]^T \end{bmatrix}.$$

Plugging this into the left-hand side of (B.12), and left-multiplying $\big[\widehat{\mathbf{U}}(\mathcal{I}), \widehat{\mathbf{U}}(\mathcal{I})^{\perp}\big]^T$ on its both sides, we obtain

$$\begin{aligned} &\underbrace{\begin{bmatrix} \widehat{\mathbf{U}}(\mathcal{I})^T \\ \big[\widehat{\mathbf{U}}(\mathcal{I})^{\perp}\big]^T \end{bmatrix} \cdot \Big[\widehat{\mathbf{U}}(\mathcal{I}), \widehat{\mathbf{U}}(\mathcal{I})^{\perp}\Big]}_{=\mathbf{I}_d} \cdot \begin{bmatrix} \Lambda_0\big(\widehat{\boldsymbol{\Sigma}}_{\mathcal{I}}\big) & \mathbf{0} \\ \mathbf{0} & \Lambda_1\big(\widehat{\boldsymbol{\Sigma}}_{\mathcal{I}}\big) \end{bmatrix} \cdot \begin{bmatrix} \widehat{\mathbf{U}}(\mathcal{I})^T\mathbf{U}^{(t)} \\ \big[\widehat{\mathbf{U}}(\mathcal{I})^{\perp}\big]^T\mathbf{U}^{(t)} \end{bmatrix} \\ &\quad = \begin{bmatrix} \Lambda_0\big(\widehat{\boldsymbol{\Sigma}}_{\mathcal{I}}\big) & \mathbf{0} \\ \mathbf{0} & \Lambda_1\big(\widehat{\boldsymbol{\Sigma}}_{\mathcal{I}}\big) \end{bmatrix} \cdot \begin{bmatrix} \widehat{\mathbf{U}}(\mathcal{I})^T\mathbf{U}^{(t)} \\ \big[\widehat{\mathbf{U}}(\mathcal{I})^{\perp}\big]^T\mathbf{U}^{(t)} \end{bmatrix} = \begin{bmatrix} \widehat{\mathbf{U}}(\mathcal{I})^T\mathbf{V}^{(t+1)} \\ \big[\widehat{\mathbf{U}}(\mathcal{I})^{\perp}\big]^T\mathbf{V}^{(t+1)} \end{bmatrix} \cdot \mathbf{R}_1^{(t+1)}. \end{aligned} \tag{B.13}$$

Here the last equality gives

$$\underbrace{\Lambda_0\big(\widehat{\boldsymbol{\Sigma}}_{\mathcal{I}}\big)}_{\text{(i)}} \cdot \underbrace{\widehat{\mathbf{U}}(\mathcal{I})^T \mathbf{U}^{(t)}}_{\text{(ii)}} = \widehat{\mathbf{U}}(\mathcal{I})^T \mathbf{V}^{(t+1)} \cdot \mathbf{R}_1^{(t+1)}. \tag{B.14}$$

Now we prove both term (i) and term (ii) on the left-hand side of (B.14) are nonsingular:

- For term (i), the perturbation theory (Stewart and Sun, 1990) for eigenvalues of symmetric matrices gives

$$\lambda_k\big(\widehat{\boldsymbol{\Sigma}}_{\mathcal{I}}\big) \geq \lambda_k(\boldsymbol{\Sigma}_{\mathcal{I}}) + \lambda_d\big(\widehat{\boldsymbol{\Sigma}}_{\mathcal{I}} - \boldsymbol{\Sigma}_{\mathcal{I}}\big) \geq \lambda_k(\boldsymbol{\Sigma}_{\mathcal{I}}) - \big\|\big(\widehat{\boldsymbol{\Sigma}} - \boldsymbol{\Sigma}\big)_{\mathcal{I}}\big\|_2 \geq \lambda_k(\boldsymbol{\Sigma}_{\mathcal{I}}) - \big\|\widehat{\boldsymbol{\Sigma}} - \boldsymbol{\Sigma}\big\|_{2,|\mathcal{I}|}, \tag{B.15}$$

where the second inequality follows from the fact that

$$\begin{aligned}\lambda_d\big(\widehat{\boldsymbol{\Sigma}}_{\mathcal{I}} - \boldsymbol{\Sigma}_{\mathcal{I}}\big) = \min_{\|\mathbf{v}\|_2=1} \mathbf{v}^T\big(\widehat{\boldsymbol{\Sigma}} - \boldsymbol{\Sigma}\big)_{\mathcal{I}}\mathbf{v} &\geq - \max_{\|\mathbf{v}\|_2=1}\Big|\mathbf{v}^T\big(\widehat{\boldsymbol{\Sigma}} - \boldsymbol{\Sigma}\big)_{\mathcal{I}}\mathbf{v}\Big| \\ &\geq - \max_{\|\mathbf{v}_2\|_2=1}\max_{\|\mathbf{v}_1\|_2=1}\Big|\mathbf{v}_1^T\big(\widehat{\boldsymbol{\Sigma}} - \boldsymbol{\Sigma}\big)_{\mathcal{I}}\mathbf{v}_2\Big| \\ &= - \max_{\|\mathbf{v}_2\|_2=1}\big\|\big(\widehat{\boldsymbol{\Sigma}} - \boldsymbol{\Sigma}\big)_{\mathcal{I}}\mathbf{v}_2\big\|_2 = -\big\|\big(\widehat{\boldsymbol{\Sigma}} - \boldsymbol{\Sigma}\big)_{\mathcal{I}}\big\|_2.\end{aligned}$$

In (B.15), the last inequality follows from

$$\begin{aligned}\big\|\big(\widehat{\boldsymbol{\Sigma}} - \boldsymbol{\Sigma}\big)_{\mathcal{I}}\big\|_2 = \max_{\|\mathbf{v}\|_2=1}\big\|\big(\widehat{\boldsymbol{\Sigma}} - \boldsymbol{\Sigma}\big)_{\mathcal{I}}\mathbf{v}\big\|_2 &= \max_{\substack{\operatorname{supp}(\mathbf{v})\subseteq\mathcal{I} \\ \|\mathbf{v}\|_2=1}}\big\|\big(\widehat{\boldsymbol{\Sigma}} - \boldsymbol{\Sigma}\big)\mathbf{v}\big\|_2 \\ &\leq \max_{\substack{|\operatorname{supp}(\mathbf{v})|\leq|\mathcal{I}| \\ \|\mathbf{v}\|_2=1}}\big\|\big(\widehat{\boldsymbol{\Sigma}} - \boldsymbol{\Sigma}\big)\mathbf{v}\big\|_2 = \big\|\widehat{\boldsymbol{\Sigma}} - \boldsymbol{\Sigma}\big\|_{2,|\mathcal{I}|}.\end{aligned}$$

Note that each column vector of $\mathbf{U}^*$, namely, $\mathbf{U}^*_{*,j} \in \mathbb{R}^d$ $(j = 1, \ldots, k)$, satisfies $\operatorname{supp}(\mathbf{U}^*_{*,j}) \subseteq \mathcal{S}^* \subseteq \mathcal{I}$. Hence we have

$$\boldsymbol{\Sigma}_{\mathcal{I}} \cdot \mathbf{U}^*_{*,j} = \boldsymbol{\Sigma} \cdot \mathbf{U}^*_{*,j} = \lambda_j \cdot \mathbf{U}^*_{*,j},$$

where the last equality is because $\mathbf{U}^*_{*,j}$ is the eigenvector of $\boldsymbol{\Sigma}$ corresponding to eigenvalue $\lambda_j$. This indicates that $\mathbf{U}^*_{*,j}$ is also an eigenvector of $\boldsymbol{\Sigma}_{\mathcal{I}}$ with eigenvalue

$$\lambda_j(\boldsymbol{\Sigma}_{\mathcal{I}}) = \lambda_j, \qquad j = 1, \ldots, k. \tag{B.16}$$

Hence, on the right-hand side of (B.15), we have

$$\begin{aligned}\lambda_k(\boldsymbol{\Sigma}_{\mathcal{I}}) - \big\|\widehat{\boldsymbol{\Sigma}} - \boldsymbol{\Sigma}\big\|_{2,|\mathcal{I}|} = \lambda_k(\boldsymbol{\Sigma}) - \big\|\widehat{\boldsymbol{\Sigma}} - \boldsymbol{\Sigma}\big\|_{2,|\mathcal{I}|} &\geq \lambda_k(\boldsymbol{\Sigma}) - \big[\lambda_k(\boldsymbol{\Sigma}) - \lambda_{k+1}(\boldsymbol{\Sigma})\big]/4 \\ &\geq \big[3\lambda_k(\boldsymbol{\Sigma}) + \lambda_{k+1}(\boldsymbol{\Sigma})\big]/4 > 0,\end{aligned}$$

where the equality follows from (B.16), and the first inequality follows from (5.6). Hence, we have $\lambda_k\big(\widehat{\boldsymbol{\Sigma}}_{\mathcal{I}}\big) > 0$ and thus $\Lambda_0\big(\widehat{\boldsymbol{\Sigma}}_{\mathcal{I}}\big)$ is nonsingular.

- For term (ii), we invoke the Cosine-Sine decomposition theorem (see Theorem 2.5.2 of Golub and Van Loan (2012)), which states

$$1 = \sigma_k\Big[\widehat{\mathbf{U}}(\mathcal{I})^T\mathbf{U}^{(t)}\Big]^2 + \sigma_1\Big[\big[\widehat{\mathbf{U}}(\mathcal{I})^{\perp}\big]^T\mathbf{U}^{(t)}\Big]^2. \tag{B.17}$$

  Since we assume $D\big[\mathcal{U}^{(t)},\widehat{\mathcal{U}}(\mathcal{I})\big]<\sqrt{2}$, which implies

$$\sigma_1\Big[\big[\widehat{\mathbf{U}}(\mathcal{I})^{\perp}\big]^T\mathbf{U}^{(t)}\Big] = \Big\|\big[\widehat{\mathbf{U}}(\mathcal{I})^{\perp}\big]^T\mathbf{U}^{(t)}\Big\|_2 \le \Big\|\big[\widehat{\mathbf{U}}(\mathcal{I})^{\perp}\big]^T\mathbf{U}^{(t)}\Big\|_{\mathrm{F}} = D\big[\mathcal{U}^{(t)},\widehat{\mathcal{U}}(\mathcal{I})\big]/\sqrt{2} < 1. \tag{B.18}$$

  Here the second equality is from Lemma 2.1 and the last equality follows from our assumption. Thus, by (B.17) we obtain $\sigma_k\big[\widehat{\mathbf{U}}(\mathcal{I})^T\mathbf{U}^{(t)}\big] > 0$ and thus $\widehat{\mathbf{U}}(\mathcal{I})^T\mathbf{U}^{(t)} \in \mathbb{R}^{k\times k}$ is nonsingular.

Therefore, term (i) and term (ii) on the left-hand side of (B.14) are square and nonsingular. On its right-hand side, $\widehat{\mathbf{U}}(\mathcal{I})^T\mathbf{V}^{(t+1)}$ and $\mathbf{R}_1^{(t+1)}$ are square. Hence, both $\widehat{\mathbf{U}}(\mathcal{I})^T\mathbf{V}^{(t+1)}$ and $\mathbf{R}_1^{(t+1)}$ are also nonsingular. Note that from (B.13) we also have

$$\Lambda_1\big(\widehat{\mathbf{\Sigma}}_{\mathcal{I}}\big)\cdot\big[\widehat{\mathbf{U}}(\mathcal{I})^{\perp}\big]^T\mathbf{U}^{(t)} = \big[\widehat{\mathbf{U}}(\mathcal{I})^{\perp}\big]^T\mathbf{V}^{(t+1)}\cdot\mathbf{R}_1^{(t+1)}.$$

Since $\mathbf{R}_1^{(t+1)}$ is nonsingular, it follows

$$\Lambda_1\big(\widehat{\mathbf{\Sigma}}_{\mathcal{I}}\big)\cdot\big[\widehat{\mathbf{U}}(\mathcal{I})^{\perp}\big]^T\mathbf{U}^{(t)}\cdot\big[\mathbf{R}_1^{(t+1)}\big]^{-1} = \big[\widehat{\mathbf{U}}(\mathcal{I})^{\perp}\big]^T\mathbf{V}^{(t+1)}. \tag{B.19}$$

Moreover, since term (i) and term (ii) on the left-hand side of (B.14) are nonsingular, we obtain

$$\big[\mathbf{R}_1^{(t+1)}\big]^{-1} = \Big[\Lambda_0\big(\widehat{\mathbf{\Sigma}}_{\mathcal{I}}\big)\cdot\widehat{\mathbf{U}}(\mathcal{I})^T\mathbf{U}^{(t)}\Big]^{-1}\widehat{\mathbf{U}}(\mathcal{I})^T\mathbf{V}^{(t+1)}.$$

Plugging this into the left-hand side of (B.19), we obtain

$$\begin{aligned}\big[\widehat{\mathbf{U}}(\mathcal{I})^{\perp}\big]^T\mathbf{V}^{(t+1)} &= \Lambda_1\big(\widehat{\mathbf{\Sigma}}_{\mathcal{I}}\big)\cdot\big[\widehat{\mathbf{U}}(\mathcal{I})^{\perp}\big]^T\mathbf{U}^{(t)}\cdot\Big[\Lambda_0\big(\widehat{\mathbf{\Sigma}}_{\mathcal{I}}\big)\cdot\widehat{\mathbf{U}}(\mathcal{I})^T\mathbf{U}^{(t)}\Big]^{-1}\widehat{\mathbf{U}}(\mathcal{I})^T\mathbf{V}^{(t+1)}\\ &= \Lambda_1\big(\widehat{\mathbf{\Sigma}}_{\mathcal{I}}\big)\cdot\big[\widehat{\mathbf{U}}(\mathcal{I})^{\perp}\big]^T\mathbf{U}^{(t)}\cdot\Big[\widehat{\mathbf{U}}(\mathcal{I})^T\mathbf{U}^{(t)}\Big]^{-1}\cdot\big[\Lambda_0\big(\widehat{\mathbf{\Sigma}}_{\mathcal{I}}\big)\big]^{-1}\cdot\widehat{\mathbf{U}}(\mathcal{I})^T\mathbf{V}^{(t+1)}.\end{aligned}$$

Now we take Frobenius norm on both sides. On the left-hand side, we have

$$D\big[\widehat{\mathcal{U}}(\mathcal{I}),\mathcal{V}^{(t+1)}\big]/\sqrt{2} = \Big\|\big[\widehat{\mathbf{U}}(\mathcal{I})^{\perp}\big]^T\mathbf{V}^{(t+1)}\Big\|_{\mathrm{F}} \tag{B.20}$$

by Lemma 2.1. Meanwhile, the Frobenius norm of the right-hand side is upper bounded by

$$\underbrace{\big\|\Lambda_1\big(\widehat{\mathbf{\Sigma}}_{\mathcal{I}}\big)\big\|_2}_{\text{(i)}}\cdot\underbrace{\Big\|\big[\widehat{\mathbf{U}}(\mathcal{I})^{\perp}\big]^T\mathbf{U}^{(t)}\Big\|_{\mathrm{F}}}_{\text{(ii)}}\cdot\underbrace{\Big\|\big[\widehat{\mathbf{U}}(\mathcal{I})^T\mathbf{U}^{(t)}\big]^{-1}\Big\|_2}_{\text{(iii)}}\cdot\underbrace{\Big\|\big[\Lambda_0\big(\widehat{\mathbf{\Sigma}}_{\mathcal{I}}\big)\big]^{-1}\Big\|_2}_{\text{(iv)}}\cdot\underbrace{\Big\|\widehat{\mathbf{U}}(\mathcal{I})^T\mathbf{V}^{(t+1)}\Big\|_2}_{\text{(v)}}. \tag{B.21}$$

Now we provide upper bounds for each term in (B.21):

- For term (i), similar to (B.15) we can prove

$$\lambda_{k+1}\big(\widehat{\mathbf{\Sigma}}_{\mathcal{I}}\big) \leq \lambda_{k+1}(\mathbf{\Sigma}_{\mathcal{I}}) + \big\|\widehat{\mathbf{\Sigma}} - \mathbf{\Sigma}\big\|_{2,|\mathcal{I}|} \leq \lambda_{k+1}(\mathbf{\Sigma}_{\mathcal{I}}) + \big\|\widehat{\mathbf{\Sigma}} - \mathbf{\Sigma}\big\|_{2,|\mathcal{I}|},$$
$$\lambda_{d}\big(\widehat{\mathbf{\Sigma}}_{\mathcal{I}}\big) \geq \lambda_{d}(\mathbf{\Sigma}_{\mathcal{I}}) - \big\|\widehat{\mathbf{\Sigma}} - \mathbf{\Sigma}\big\|_{2,|\mathcal{I}|} \geq \lambda_{d}(\mathbf{\Sigma}_{\mathcal{I}}) - \big\|\widehat{\mathbf{\Sigma}} - \mathbf{\Sigma}\big\|_{2,|\mathcal{I}|},$$

which implies $\big|\lambda_{k+1}\big(\widehat{\mathbf{\Sigma}}_{\mathcal{I}}\big)\big| \geq \big|\lambda_{d}\big(\widehat{\mathbf{\Sigma}}_{\mathcal{I}}\big)\big|$ since $\lambda_{k+1}(\mathbf{\Sigma}_{\mathcal{I}}) \geq \lambda_{d}(\mathbf{\Sigma}_{\mathcal{I}}) \geq 0$. Therefore, $\big\|\Lambda_1\big(\widehat{\mathbf{\Sigma}}_{\mathcal{I}}\big)\big\|_2 = \lambda_{k+1}\big(\widehat{\mathbf{\Sigma}}_{\mathcal{I}}\big)$. Consequently, for term (i) in (B.21) we have

$$\big\|\Lambda_1\big(\widehat{\mathbf{\Sigma}}_{\mathcal{I}}\big)\big\|_2 = \lambda_{k+1}\big(\widehat{\mathbf{\Sigma}}_{\mathcal{I}}\big) \leq \lambda_{k+1}(\mathbf{\Sigma}_{\mathcal{I}}) + \big\|\widehat{\mathbf{\Sigma}} - \mathbf{\Sigma}\big\|_{2,|\mathcal{I}|}.$$

- For terms (ii) and (iii) in (B.21), we have

$$\Big\|\big[\widehat{\mathbf{U}}(\mathcal{I})^{\perp}\big]^T \mathbf{U}^{(t)}\Big\|_{\mathrm{F}} = D\big[\widehat{\mathcal{U}}(\mathcal{I}), \mathcal{U}^{(t)}\big]/\sqrt{2}$$

by Lemma 2.1, and

$$\Big\|\big[\widehat{\mathbf{U}}(\mathcal{I})^T \mathbf{U}^{(t)}\big]^{-1}\Big\|_2 = \frac{1}{\sigma_k\Big[\widehat{\mathbf{U}}(\mathcal{I})^T \mathbf{U}^{(t)}\Big]} = \frac{1}{\sqrt{1 - \sigma_1\Big[\big[\widehat{\mathbf{U}}(\mathcal{I})^{\perp}\big]^T \mathbf{U}^{(t)}\Big]^2}}$$
$$\leq \frac{1}{\sqrt{1 - D\big[\widehat{\mathcal{U}}(\mathcal{I}), \mathcal{U}^{(t)}\big]^2/(2k)}},$$

where the second equality follows from (B.17) and the inequality follows from

$$\sigma_1\Big[\big[\widehat{\mathbf{U}}(\mathcal{I})^{\perp}\big]^T \mathbf{U}^{(t)}\Big] = \Big\|\big[\widehat{\mathbf{U}}(\mathcal{I})^{\perp}\big]^T \mathbf{U}^{(t)}\Big\|_2 \geq \Big\|\big[\widehat{\mathbf{U}}(\mathcal{I})^{\perp}\big]^T \mathbf{U}^{(t)}\Big\|_{\mathrm{F}}/\sqrt{k} = D\big[\widehat{\mathcal{U}}(\mathcal{I}), \mathcal{U}^{(t)}\big]/\sqrt{2k}$$

by Lemma 2.1.

- For term (iv) in (B.21), similar to (B.15) we have

$$\lambda_k\big(\widehat{\mathbf{\Sigma}}_{\mathcal{I}}\big) \geq \lambda_k(\mathbf{\Sigma}_{\mathcal{I}}) - \big\|\widehat{\mathbf{\Sigma}} - \mathbf{\Sigma}\big\|_{2,|\mathcal{I}|} = \lambda_k(\mathbf{\Sigma}) - \big\|\widehat{\mathbf{\Sigma}} - \mathbf{\Sigma}\big\|_{2,|\mathcal{I}|} > 0,$$

where the equality follows from (B.16), and the last inequality is from our assumption that $n$ is sufficiently large such that $\big\|\widehat{\mathbf{\Sigma}} - \mathbf{\Sigma}\big\|_{2,|\mathcal{I}|} \leq \big[\lambda_k - \lambda_{k+1}\big]/4$. Hence we obtain

$$\Big\|\big[\Lambda_0\big(\widehat{\mathbf{\Sigma}}_{\mathcal{I}}\big)\big]^{-1}\Big\|_2 = \frac{1}{\lambda_k\big(\widehat{\mathbf{\Sigma}}_{\mathcal{I}}\big)} \leq \frac{1}{\lambda_k(\mathbf{\Sigma}_{\mathcal{I}}) - \big\|\widehat{\mathbf{\Sigma}} - \mathbf{\Sigma}\big\|_{2,|\mathcal{I}|}}.$$

- For term (v) in (B.21), we have

$$\Big\|\widehat{\mathbf{U}}(\mathcal{I})^T \mathbf{V}^{(t+1)}\Big\|_2 \leq \big\|\widehat{\mathbf{U}}(\mathcal{I})\big\|_2 \cdot \big\|\mathbf{V}^{(t+1)}\big\|_2 = 1,$$

where the in last equality we use the fact that both $\widehat{\mathbf{U}}(\mathcal{I})$ and $\mathbf{V}^{(t+1)}$ are orthonormal matrices, which implies $\big\|\widehat{\mathbf{U}}(\mathcal{I})\big\|_2 = \big\|\mathbf{V}^{(t+1)}\big\|_2 = 1$.

Combining these upper bounds of the terms in (B.21) and combining (B.20), we obtain

$$\begin{aligned} D\big[\widehat{\mathcal{U}}(\mathcal{I}),\mathcal{V}^{(t+1)}\big] &\le \frac{D\big[\widehat{\mathcal{U}}(\mathcal{I}),\mathcal{U}^{(t)}\big]}{\sqrt{1-D\big[\widehat{\mathcal{U}}(\mathcal{I}),\mathcal{U}^{(t)}\big]^2/(2k)}}\cdot\frac{\lambda_{k+1}(\boldsymbol{\Sigma})+\big\|\widehat{\boldsymbol{\Sigma}}-\boldsymbol{\Sigma}\big\|_{2,|\mathcal{I}|}}{\lambda_k(\boldsymbol{\Sigma})-\big\|\widehat{\boldsymbol{\Sigma}}-\boldsymbol{\Sigma}\big\|_{2,|\mathcal{I}|}} \\ &\le \frac{D\big[\widehat{\mathcal{U}}(\mathcal{I}),\mathcal{U}^{(t)}\big]}{\sqrt{1-D\big[\widehat{\mathcal{U}}(\mathcal{I}),\mathcal{U}^{(t)}\big]^2/(2k)}}\cdot\gamma. \end{aligned}$$

Here the second inequality follows from (5.6) and the definition $\gamma=(3\lambda_{k+1}+\lambda_k)/(\lambda_{k+1}+3\lambda_k)$. Thus we conclude the proof. □

### B.3 Proof of Lemma 5.5

*Proof.* First we consider the relationship between $\big\|(\mathbf{U}^*)^T\mathbf{V}^{(t+1)}\big\|_{\mathrm{F}}$ and $\big\|(\mathbf{U}^*)^T\widetilde{\mathbf{U}}^{(t+1)}\big\|_{\mathrm{F}}$. Note that

$$\begin{aligned} \big\|(\mathbf{U}^*)^T\widetilde{\mathbf{U}}^{(t+1)}\big\|_{\mathrm{F}} &= \big\|(\mathbf{U}^*)^T\mathbf{V}^{(t+1)}+(\mathbf{U}^*)^T\widetilde{\mathbf{U}}^{(t+1)}-(\mathbf{U}^*)^T\mathbf{V}^{(t+1)}\big\|_{\mathrm{F}} \\ &\ge \big\|(\mathbf{U}^*)^T\mathbf{V}^{(t+1)}\big\|_{\mathrm{F}}-\big\|(\mathbf{U}^*)^T\widetilde{\mathbf{U}}^{(t+1)}-(\mathbf{U}^*)^T\mathbf{V}^{(t+1)}\big\|_{\mathrm{F}}, \end{aligned} \tag{B.22}$$

where the last inequality follows from triangular inequality. In Algorithm 2 (Line 2), $\mathcal{I}_{\widehat{s}}$ is the set of row index $i$'s corresponding to the top $\widehat{s}$ largest row $\ell_2$ norm $\big\|\mathbf{V}^{(t+1)}_{i,*}\big\|_2$'s. Recall $\mathcal{S}^*$ is the index set of the nonzero rows of $\mathbf{U}^*$, whose columns are the top $k$ leading eigenvectors of $\boldsymbol{\Sigma}$.

Let $\mathcal{I}_1=\mathcal{S}^*\backslash\mathcal{I}_{\widehat{s}}$, $\mathcal{I}_2=\mathcal{S}^*\cap\mathcal{I}_{\widehat{s}}$ and $\mathcal{I}_3=\mathcal{I}_{\widehat{s}}\backslash\mathcal{S}^*$. By the definition of the truncation function in Algorithm 2, for the second term in (B.22) we have

$$\big\|(\mathbf{U}^*)^T\widetilde{\mathbf{U}}^{(t+1)}-(\mathbf{U}^*)^T\mathbf{V}^{(t+1)}\big\|_{\mathrm{F}}=\big\|(\mathbf{U}^*)^T\big(\widetilde{\mathbf{U}}^{(t+1)}-\mathbf{V}^{(t+1)}\big)\big\|_{\mathrm{F}}=\big\|(\mathbf{U}^*_{\mathcal{I}_1,\cdot})^T\mathbf{V}^{(t+1)}_{\mathcal{I}_1,\cdot}\big\|_{\mathrm{F}}. \tag{B.23}$$

The last equality is because the row-support of $\mathbf{U}^*$ is $\mathcal{S}^*$, while the row-support of $\widetilde{\mathbf{U}}^{(t+1)}-\mathbf{V}^{(t+1)}$ is $\{1,\ldots,d\}\backslash\mathcal{I}_{\widehat{s}}$, since $\widetilde{\mathbf{U}}^{(t+1)}=\mathbf{V}^{(t+1)}_{\mathcal{I}_{\widehat{s}},\cdot}$. Note that $\big\|(\mathbf{U}^*_{\mathcal{I}_1,\cdot})^T\mathbf{V}^{(t+1)}_{\mathcal{I}_1,\cdot}\big\|_{\mathrm{F}}\le\big\|\mathbf{U}^*_{\mathcal{I}_1,\cdot}\big\|_2\cdot\big\|\mathbf{V}^{(t+1)}_{\mathcal{I}_1,\cdot}\big\|_{\mathrm{F}}$. Thus, plugging (B.23) into the right-hand side of (B.22), we obtain

$$\big\|(\mathbf{U}^*)^T\widetilde{\mathbf{U}}^{(t+1)}\big\|_{\mathrm{F}}\ge\big\|(\mathbf{U}^*)^T\mathbf{V}^{(t+1)}\big\|_{\mathrm{F}}-\big\|\mathbf{V}^{(t+1)}_{\mathcal{I}_1,\cdot}\big\|_{\mathrm{F}}\cdot\big\|\mathbf{U}^*_{\mathcal{I}_1,\cdot}\big\|_2. \tag{B.24}$$

Now we derive the relationship between $\big\|(\mathbf{U}^*)^T\widetilde{\mathbf{U}}^{(t+1)}\big\|_{\mathrm{F}}$ and $\big\|(\mathbf{U}^*)^T\mathbf{U}^{(t+1)}\big\|_{\mathrm{F}}$. The QR decomposition in (5.9) gives $\mathbf{U}^{(t+1)}\cdot\mathbf{R}_2^{(t+1)}=\widetilde{\mathbf{U}}^{(t+1)}$, which implies

$$\big\|(\mathbf{U}^*)^T\widetilde{\mathbf{U}}^{(t+1)}\big\|_{\mathrm{F}}=\big\|(\mathbf{U}^*)^T\mathbf{U}^{(t+1)}\cdot\mathbf{R}_2^{(t+1)}\big\|_{\mathrm{F}}\le\big\|(\mathbf{U}^*)^T\mathbf{U}^{(t+1)}\big\|_{\mathrm{F}}\cdot\big\|\mathbf{R}_2^{(t+1)}\big\|_2. \tag{B.25}$$

Meanwhile, since $\mathbf{U}^{(t+1)}$ is orthonormal, we have

$$\big\|\mathbf{R}_2^{(t+1)}\big\|_2=\big\|\mathbf{U}^{(t+1)}\mathbf{R}_2^{(t+1)}\big\|_2=\big\|\widetilde{\mathbf{U}}^{(t+1)}\big\|_2=\big\|\mathbf{V}^{(t+1)}_{\mathcal{I}_{\widehat{s}},\cdot}\big\|_2\le\big\|\mathbf{V}^{(t+1)}\big\|_2=1, \tag{B.26}$$

where the second last equality follows from the definition of the truncation function in Algorithm 2, and the last inequality follows from the fact that

$$\big\|\mathbf{V}^{(t+1)}_{\mathcal{I}_{\widehat{s}},\cdot}\big\|_2=\max_{\|\mathbf{v}\|_2=1}\big\|\mathbf{v}^T\mathbf{V}^{(t+1)}_{\mathcal{I}_{\widehat{s}},\cdot}\big\|_2\le\max_{\|\mathbf{v}\|_2=1}\big\|\mathbf{v}^T\mathbf{V}^{(t+1)}\big\|_2=\big\|\mathbf{V}^{(t+1)}\big\|_2,$$

while the last equality is because $\mathbf{V}^{(t+1)}$ is orthonormal. Plugging (B.26) into the right-hand side of (B.25), we obtain

$$\big\|(\mathbf{U}^*)^T\widetilde{\mathbf{U}}^{(t+1)}\big\|_{\mathrm{F}} \le \big\|(\mathbf{U}^*)^T\mathbf{U}^{(t+1)}\big\|_{\mathrm{F}} \cdot \big\|\mathbf{R}_2^{(t+1)}\big\|_2 \le \big\|(\mathbf{U}^*)^T\mathbf{U}^{(t+1)}\big\|_{\mathrm{F}}.$$

Together with (B.24), we have

$$\big\|(\mathbf{U}^*)^T\mathbf{U}^{(t+1)}\big\|_{\mathrm{F}} \ge \big\|(\mathbf{U}^*)^T\widetilde{\mathbf{U}}^{(t+1)}\big\|_{\mathrm{F}} \ge \big\|(\mathbf{U}^*)^T\mathbf{V}^{(t+1)}\big\|_{\mathrm{F}} - \big\|\mathbf{V}^{(t+1)}_{\mathcal{I}_1,\cdot}\big\|_{\mathrm{F}} \cdot \big\|\mathbf{U}^*_{\mathcal{I}_1,\cdot}\big\|_2. \tag{B.27}$$

In the rest of the proof, we focus on providing upper bounds for $\big\|\mathbf{V}^{(t+1)}_{\mathcal{I}_1,\cdot}\big\|_{\mathrm{F}}$ and $\big\|\mathbf{U}^*_{\mathcal{I}_1,\cdot}\big\|_2$.

**Upper Bound for $\big\|\mathbf{V}^{(t+1)}_{\mathcal{I}_1,\cdot}\big\|_{\mathrm{F}}$:** By the definition of the truncation function, we have

$$\frac{1}{|\mathcal{I}_1|}\sum_{i\in\mathcal{I}_1}\big\|\mathbf{V}^{(t+1)}_{i,*}\big\|_2^2 \le \frac{1}{|\mathcal{I}_3|}\sum_{i\in\mathcal{I}_3}\big\|\mathbf{V}^{(t+1)}_{i,*}\big\|_2^2,$$

which implies

$$\frac{1}{|\mathcal{I}_1|}\cdot\big\|\mathbf{V}^{(t+1)}_{\mathcal{I}_1,\cdot}\big\|_{\mathrm{F}}^2 = \frac{1}{|\mathcal{I}_1|}\sum_{i\in\mathcal{I}_1}\big\|\mathbf{V}^{(t+1)}_{i,*}\big\|_2^2 \le \frac{1}{|\mathcal{I}_3|}\sum_{i\in\mathcal{I}_3}\big\|\mathbf{V}^{(t+1)}_{i,*}\big\|_2^2 = \frac{1}{|\mathcal{I}_3|}\cdot\big\|\mathbf{V}^{(t+1)}_{\mathcal{I}_3,\cdot}\big\|_{\mathrm{F}}^2.$$

Here remind that $\mathbf{V}^{(t+1)}_{i,*} \in \mathbb{R}^k$ denotes the $i$-th row vector of $\mathbf{V}^{(t+1)}$. Hence, we obtain

$$\big\|\mathbf{V}^{(t+1)}_{\mathcal{I}_3,\cdot}\big\|_{\mathrm{F}}^2 \ge \frac{|\mathcal{I}_3|}{|\mathcal{I}_1|}\cdot\big\|\mathbf{V}^{(t+1)}_{\mathcal{I}_1,\cdot}\big\|_{\mathrm{F}}^2 \ge \frac{|\mathcal{I}_3|+|\mathcal{I}_2|}{|\mathcal{I}_1|+|\mathcal{I}_2|}\cdot\big\|\mathbf{V}^{(t+1)}_{\mathcal{I}_1,\cdot}\big\|_{\mathrm{F}}^2 = \frac{\widehat{s}}{s^*}\cdot\big\|\mathbf{V}^{(t+1)}_{\mathcal{I}_1,\cdot}\big\|_{\mathrm{F}}^2.$$

Here the second inequality follows from $a/b \ge (a+c)/(b+c)$ for $a \ge b > 0$, and $|\mathcal{I}_3| = \widehat{s} - |\mathcal{I}_2| \ge s^* - |\mathcal{I}_2| = |\mathcal{I}_1|$, which is from our assumption that $\widehat{s} \ge s^*$. The last equality is because $\mathcal{I}_1 \cup \mathcal{I}_2 = \mathcal{S}^*$ and $\mathcal{I}_2 \cup \mathcal{I}_3 = \mathcal{I}_{\widehat{s}}$. Then it follows that

$$k - \frac{\widehat{s}}{s^*}\cdot\big\|\mathbf{V}^{(t+1)}_{\mathcal{I}_1,\cdot}\big\|_{\mathrm{F}}^2 \ge k - \big\|\mathbf{V}^{(t+1)}_{\mathcal{I}_3,\cdot}\big\|_{\mathrm{F}}^2. \tag{B.28}$$

Now we provide a lower bound for its right-hand side. Note that

$$\begin{aligned}\big\|(\mathbf{U}^*)^T\mathbf{V}^{(t+1)}\big\|_{\mathrm{F}} = \big\|(\mathbf{U}^*)^T\mathbf{V}^{(t+1)}_{\mathcal{S}^*,\cdot}\big\|_{\mathrm{F}} = \big\|(\mathbf{U}^*)^T\mathbf{V}^{(t+1)}_{\mathcal{I}_1\cup\mathcal{I}_2,\cdot}\big\|_{\mathrm{F}} &\le \|\mathbf{U}^*\|_2\cdot\big\|\mathbf{V}^{(t+1)}_{\mathcal{I}_1\cup\mathcal{I}_2,\cdot}\big\|_{\mathrm{F}}\\ &\le \big\|\mathbf{V}^{(t+1)}_{\mathcal{I}_1\cup\mathcal{I}_2,\cdot}\big\|_{\mathrm{F}},\end{aligned} \tag{B.29}$$

where we use $\|\mathbf{U}^*\|_2 = 1$ in the last inequality because $\mathbf{U}^*$ is orthonormal. Meanwhile, we have

$$\big\|\mathbf{V}^{(t+1)}_{\mathcal{I}_1\cup\mathcal{I}_2,\cdot}\big\|_{\mathrm{F}}^2 + \big\|\mathbf{V}^{(t+1)}_{\mathcal{I}_3,\cdot}\big\|_{\mathrm{F}}^2 = \big\|\mathbf{V}^{(t+1)}_{\mathcal{I}_1\cup\mathcal{I}_2\cup\mathcal{I}_3,\cdot}\big\|_{\mathrm{F}}^2 \le \big\|\mathbf{V}^{(t+1)}\big\|_{\mathrm{F}}^2 \le k\cdot\big\|\mathbf{V}^{(t+1)}\big\|_2^2 = k, \tag{B.30}$$

where we use $\big\|\mathbf{V}^{(t+1)}\big\|_2 = 1$ in the last equality, since $\mathbf{V}^{(t+1)}$ is also orthonormal. Combining (B.29) and (B.30), we obtain

$$\big\|(\mathbf{U}^*)^T\mathbf{V}^{(t+1)}\big\|_{\mathrm{F}}^2 \le \big\|\mathbf{V}^{(t+1)}_{\mathcal{I}_1\cup\mathcal{I}_2,\cdot}\big\|_{\mathrm{F}}^2 \le k - \big\|\mathbf{V}^{(t+1)}_{\mathcal{I}_3,\cdot}\big\|_{\mathrm{F}}^2.$$

Plugging this into the right-hand side of (B.28) and we obtain

$$k - \frac{\widehat{s}}{s^*} \cdot \big\|\mathbf{V}^{(t+1)}_{\mathcal{I}_1,\cdot}\big\|_{\mathrm{F}}^2 \geq \big\|(\mathbf{U}^*)^T \mathbf{V}^{(t+1)}\big\|_{\mathrm{F}}^2,$$

which implies

$$\big\|\mathbf{V}^{(t+1)}_{\mathcal{I}_1,\cdot}\big\|_{\mathrm{F}} \leq \sqrt{\frac{s^*}{\widehat{s}}} \cdot \sqrt{k - \big\|(\mathbf{U}^*)^T \mathbf{V}^{(t+1)}\big\|_{\mathrm{F}}^2} = \sqrt{\frac{s^*}{\widehat{s}}} \cdot D\big[\mathcal{U}^*, \mathcal{U}^{(t+1)}\big]/\sqrt{2}, \tag{B.31}$$

where the last inequality follows from Lemma 2.1.

**Upper Bound for $\big\|\mathbf{U}^*_{\mathcal{I}_1,\cdot}\big\|_2$:** Note that

$$\big\|\mathbf{U}^*_{\mathcal{I}_1,\cdot}\big\|_{\mathrm{F}}^2 + \big\|\mathbf{U}^*_{\mathcal{I}_2,\cdot}\big\|_{\mathrm{F}}^2 = \big\|\mathbf{U}^*_{\mathcal{I}_1 \cup \mathcal{I}_2,\cdot}\big\|_{\mathrm{F}}^2 = \big\|\mathbf{U}^*_{\mathcal{S}^*,\cdot}\big\|_{\mathrm{F}}^2 = \|\mathbf{U}^*\|_{\mathrm{F}}^2 \leq k \cdot \|\mathbf{U}^*\|_2^2 = k,$$

where the second equality is because $\mathcal{I}_1 \cup \mathcal{I}_2 = \mathcal{S}^*$, the third equality is because the row-support of $\mathbf{U}^*$ is $\mathcal{S}^*$, and the last equality follows from $\|\mathbf{U}^*\|_2 = 1$ since $\mathbf{U}^*$ is orthonormal. Hence, we have

$$\big\|\mathbf{U}^*_{\mathcal{I}_1,\cdot}\big\|_{\mathrm{F}}^2 \leq k - \big\|\mathbf{U}^*_{\mathcal{I}_2,\cdot}\big\|_{\mathrm{F}}^2. \tag{B.32}$$

Moreover, note that

$$\big\|(\mathbf{U}^*)^T \mathbf{U}^{(t+1)}\big\|_{\mathrm{F}} = \big\|(\mathbf{U}^*_{\mathcal{S}^* \cap \mathcal{I}_{\widehat{s}},\cdot})^T \mathbf{U}^{(t+1)}\big\|_{\mathrm{F}} = \big\|(\mathbf{U}^*_{\mathcal{I}_2,\cdot})^T \mathbf{U}^{(t+1)}\big\|_{\mathrm{F}}, \tag{B.33}$$

where the first equality is because the row-support of $\mathbf{U}^{(t+1)}$ is $\mathcal{I}_{\widehat{s}}$ (since the row-support of $\widetilde{\mathbf{U}}^{(t+1)}$ is $\mathcal{I}_{\widehat{s}}$ by the definition of truncation function, and the QR decomposition in (5.9) doesn't change its row-support), while the second is because $\mathcal{S}^* \cap \mathcal{I}_{\widehat{s}} = \mathcal{I}_2$ by our definition. Meanwhile, we have

$$\big\|(\mathbf{U}^*_{\mathcal{I}_2,\cdot})^T \mathbf{U}^{(t+1)}\big\|_{\mathrm{F}} \leq \big\|\mathbf{U}^*_{\mathcal{I}_2,\cdot}\big\|_{\mathrm{F}} \cdot \big\|\mathbf{U}^{(t+1)}\big\|_2 = \big\|\mathbf{U}^*_{\mathcal{I}_2,\cdot}\big\|_{\mathrm{F}}, \tag{B.34}$$

where the last equality is because $\big\|\mathbf{U}^{(t+1)}\big\|_2 = 1$, since $\mathbf{U}^{(t+1)}$ is the output of the QR decomposition in (5.9) and is thus orthonormal. Plugging (B.34) into the right-hand side of (B.33), we obtain

$$\big\|(\mathbf{U}^*)^T \mathbf{U}^{(t+1)}\big\|_{\mathrm{F}} \leq \big\|\mathbf{U}^*_{\mathcal{I}_2,\cdot}\big\|_{\mathrm{F}},$$

which together with (B.32) implies

$$\big\|\mathbf{U}^*_{\mathcal{I}_1,\cdot}\big\|_{\mathrm{F}}^2 \leq k - \big\|\mathbf{U}^*_{\mathcal{I}_2,\cdot}\big\|_{\mathrm{F}}^2 \leq k - \big\|(\mathbf{U}^*)^T \mathbf{U}^{(t+1)}\big\|_{\mathrm{F}}^2 = D\big[\mathcal{U}^*, \mathcal{U}^{(t+1)}\big]^2/2,$$

where the last equality follows from Lemma 2.1. Finally we have

$$\big\|\mathbf{U}^*_{\mathcal{I}_1,\cdot}\big\|_2 \leq \big\|\mathbf{U}^*_{\mathcal{I}_1,\cdot}\big\|_{\mathrm{F}} \leq D\big[\mathcal{U}^*, \mathcal{U}^{(t+1)}\big]/\sqrt{2}. \tag{B.35}$$

**Derivation of (5.10):** So far we obtain the upper bounds for $\big\|\mathbf{V}^{(t+1)}_{\mathcal{I}_1,\cdot}\big\|_{\mathrm{F}}$ and $\big\|\mathbf{U}^*_{\mathcal{I}_1,\cdot}\big\|_2$ respectively in (B.31) and (B.35). Furthermore, plugging those upper bounds into the right-hand side of (B.27) gives

$$\begin{aligned} \big\|(\mathbf{U}^*)^T \mathbf{U}^{(t+1)}\big\|_{\mathrm{F}} &\geq \big\|(\mathbf{U}^*)^T \mathbf{V}^{(t+1)}\big\|_{\mathrm{F}} - \big\|\mathbf{V}^{(t+1)}_{\mathcal{I}_1,\cdot}\big\|_{\mathrm{F}} \cdot \big\|\mathbf{U}^*_{\mathcal{I}_1,\cdot}\big\|_2 \\ &\geq \big\|(\mathbf{U}^*)^T \mathbf{V}^{(t+1)}\big\|_{\mathrm{F}} - \sqrt{\frac{s^*}{\widehat{s}}} \cdot D\big[\mathcal{U}^*, \mathcal{V}^{(t+1)}\big] \cdot D\big[\mathcal{U}^*, \mathcal{U}^{(t+1)}\big]/2. \end{aligned} \tag{B.36}$$

Now we prove that the right-hand side of (B.36) is nonnegative. By Lemma 2.1, we have

$$\big\|(\mathbf{U}^*)^T\mathbf{V}^{(t+1)}\big\|_{\mathrm{F}} = \sqrt{k - D\big[\mathcal{U}^*,\mathcal{V}^{(t+1)}\big]^2/2}, \quad D\big[\mathcal{U}^*,\mathcal{U}^{(t+1)}\big] \le \sqrt{2k}.$$

By our assumption, we have $\sqrt{s^*/\widehat{s}} \le 1$ and $D\big[\mathcal{U}^*,\mathcal{V}^{(t+1)}\big] \le 1$. Thus, the right-hand side of (B.36) is lower bounded by

$$\sqrt{k - D\big[\mathcal{U}^*,\mathcal{V}^{(t+1)}\big]^2/2} - D\big[\mathcal{U}^*,\mathcal{U}^{(t+1)}\big]/2 \ge \sqrt{k-1/2} - \sqrt{k/2} \ge 0,$$

where the last inequality is because $k \ge 1$. Taking square of the both sides of (B.36) gives

$$\big\|(\mathbf{U}^*)^T\mathbf{U}^{(t+1)}\big\|_{\mathrm{F}}^2 \ge \big\|(\mathbf{U}^*)^T\mathbf{V}^{(t+1)}\big\|_{\mathrm{F}}^2 - \big\|(\mathbf{U}^*)^T\mathbf{V}^{(t+1)}\big\|_{\mathrm{F}} \cdot \sqrt{\frac{s^*}{\widehat{s}}} \cdot D\big[\mathcal{U}^*,\mathcal{V}^{(t+1)}\big] \cdot D\big[\mathcal{U}^*,\mathcal{U}^{(t+1)}\big]. \tag{B.37}$$

By Lemma 2.1 we have

$$D\big[\mathcal{U}^*,\mathcal{U}^{(t+1)}\big]^2/2 = k - \big\|(\mathbf{U}^*)^T\mathbf{U}^{(t+1)}\big\|_{\mathrm{F}}^2.$$

Plugging (B.37) into the right-hand side of the above quality, we obtain

$$\begin{aligned} D\big[\mathcal{U}^*,\mathcal{U}^{(t+1)}\big]^2/2 &\le k - \big\|(\mathbf{U}^*)^T\mathbf{V}^{(t+1)}\big\|_{\mathrm{F}}^2 + \big\|(\mathbf{U}^*)^T\mathbf{V}^{(t+1)}\big\|_{\mathrm{F}} \cdot \sqrt{\frac{s^*}{\widehat{s}}} \cdot D\big[\mathcal{U}^*,\mathcal{V}^{(t+1)}\big] \cdot D\big[\mathcal{U}^*,\mathcal{U}^{(t+1)}\big] \\ &\le D\big[\mathcal{U}^*,\mathcal{V}^{(t+1)}\big]^2/2 + \sqrt{\frac{k \cdot s^*}{\widehat{s}}} \cdot D\big[\mathcal{U}^*,\mathcal{V}^{(t+1)}\big] \cdot D\big[\mathcal{U}^*,\mathcal{U}^{(t+1)}\big], \end{aligned} \tag{B.38}$$

where the last inequality follows from the fact that

$$\big\|(\mathbf{U}^*)^T\mathbf{V}^{(t+1)}\big\|_{\mathrm{F}} = \sqrt{k - D\big[\mathcal{U}^*,\mathcal{V}^{(t+1)}\big]^2/2} \le \sqrt{k}.$$

Solving for $D\big[\mathcal{U}^*,\mathcal{U}^{(t+1)}\big]$ in the quadratic inequality (B.38), we obtain

$$\begin{aligned} D\big[\mathcal{U}^*,\mathcal{U}^{(t+1)}\big] &\le \sqrt{\frac{k \cdot s^*}{\widehat{s}}} \cdot D\big[\mathcal{U}^*,\mathcal{V}^{(t+1)}\big] + \sqrt{\frac{k \cdot s^*}{\widehat{s}} \cdot D\big[\mathcal{U}^*,\mathcal{V}^{(t+1)}\big]^2 + D\big[\mathcal{U}^*,\mathcal{V}^{(t+1)}\big]^2} \\ &\le \left(1 + 2\sqrt{\frac{k \cdot s^*}{\widehat{s}}}\right) \cdot D\big[\mathcal{U}^*,\mathcal{V}^{(t+1)}\big], \end{aligned}$$

which concludes the proof. □

## B.4 Proof of Lemma 5.6

*Proof.* We define $\mathcal{I}$ as the union of the row-supports of $\mathbf{U}^*$, $\mathbf{U}^{(t)}$ in the rest of the proof. Since we assume $\widehat{s} \ge s^*$, and by our algorithm the row-sparsity level of $\mathbf{U}^{(t)}$ is $\widehat{s}$, we have $|\mathcal{I}| \le s^* + \widehat{s} \le 2\widehat{s}$.

Remind we assume $D\big[\mathcal{U}^*,\mathcal{U}^{(t)}\big] \le \sqrt{2}/2$. Meanwhile, by Assumption 5.1, for a sufficiently large $n$, we have $D\big[\mathcal{U}^*,\widehat{\mathcal{U}}(\mathcal{I})\big] \le \sqrt{2}/2$. By triangular inequality we have

$$D\big[\widehat{\mathcal{U}}(\mathcal{I}),\mathcal{U}^{(t)}\big] \le D\big[\mathcal{U}^*,\mathcal{U}^{(t)}\big] + D\big[\mathcal{U}^*,\widehat{\mathcal{U}}(\mathcal{I})\big] \le \sqrt{2}.$$

Then we can apply Lemma 5.4 and obtain

$$D\big[\mathcal{V}^{(t+1)},\widehat{\mathcal{U}}(\mathcal{I})\big] \le \frac{D\big[\mathcal{U}^{(t)},\widehat{\mathcal{U}}(\mathcal{I})\big]}{\sqrt{1-D\big[\mathcal{U}^{(t)},\widehat{\mathcal{U}}(\mathcal{I})\big]^2/(2k)}}\cdot\gamma.$$

Meanwhile, by triangular inequality we have

$$\Big|D\big[\mathcal{U}^{(t+1)},\mathcal{U}^*\big] - D\big[\mathcal{U}^{(t+1)},\widehat{\mathcal{U}}(\mathcal{I})\big]\Big| \le D\big[\mathcal{U}^*,\widehat{\mathcal{U}}(\mathcal{I})\big], \tag{B.39}$$

$$\Big|D\big[\mathcal{V}^{(t+1)},\mathcal{U}^*\big] - D\big[\mathcal{V}^{(t+1)},\widehat{\mathcal{U}}(\mathcal{I})\big]\Big| \le D\big[\mathcal{U}^*,\widehat{\mathcal{U}}(\mathcal{I})\big]. \tag{B.40}$$

By Assumption 5.1 we have

$$D\big[\mathcal{U}^*,\widehat{\mathcal{U}}(\mathcal{I})\big] \le \Delta(|\mathcal{I}|), \tag{B.41}$$

where $\Delta(\cdot)$ is defined in (5.11). Hence, from (B.40) and (B.41) we obtain

$$D\big[\mathcal{V}^{(t+1)},\mathcal{U}^*\big] \le D\big[\mathcal{V}^{(t+1)},\widehat{\mathcal{U}}(\mathcal{I})\big] + \Delta(|\mathcal{I}|) \le \frac{D\big[\mathcal{U}^{(t)},\widehat{\mathcal{U}}(\mathcal{I})\big]}{\sqrt{1-D\big[\mathcal{U}^{(t)},\widehat{\mathcal{U}}(\mathcal{I})\big]^2/(2k)}}\cdot\gamma + \Delta(|\mathcal{I}|). \tag{B.42}$$

Consider the univariate function $f(z) = z/\sqrt{1-z^2/(2k)}$, whose derivative is $f'(z){=}\big[1{-}z^2/(2k)\big]^{-\frac{3}{2}}$. According to the convexity of $f(z)$ within its domain $\big(-\sqrt{2k},\sqrt{2k}\big)$, we have

$$f(z_2) - f(z_1) \le f'(z_2)\cdot(z_2 - z_1).$$

Setting $z_2 = D\big[\mathcal{U}^{(t)},\widehat{\mathcal{U}}(\mathcal{I})\big]$ and $z_1 = D\big[\mathcal{U}^{(t)},\mathcal{U}^*\big]$ yields

$$\frac{D\big[\mathcal{U}^{(t)},\widehat{\mathcal{U}}(\mathcal{I})\big]}{\sqrt{1-D\big[\mathcal{U}^{(t)},\widehat{\mathcal{U}}(\mathcal{I})\big]^2/(2k)}} \le \frac{D\big[\mathcal{U}^{(t)},\mathcal{U}^*\big]}{\sqrt{1-D\big[\mathcal{U}^{(t)},\mathcal{U}^*\big]^2/(2k)}} + \Delta(|\mathcal{I}|)\cdot\Big[1-D\big[\mathcal{U}^{(t)},\widehat{\mathcal{U}}(\mathcal{I})\big]^2/(2k)\Big]^{-3/2}. \tag{B.43}$$

Here we use the fact that

$$\Big|D\big[\mathcal{U}^{(t)},\mathcal{U}^*\big] - D\big[\mathcal{U}^{(t)},\widehat{\mathcal{U}}(\mathcal{I})\big]\Big| \le D\big[\mathcal{U}^*,\widehat{\mathcal{U}}(\mathcal{I})\big] \le \Delta(|\mathcal{I}|),$$

where the last inequality follows from (B.41). For the second term on the right-hand side of (B.43), by Assumption 5.1, for a sufficiently large $n$ we have $D\big[\mathcal{U}^{(t)},\mathcal{U}^*\big] \le \sqrt{2}/2$. Meanwhile, remind that we assume $\Delta(2\widehat{s}) \le 1/24$. Hence, we obtain

$$D\big[\mathcal{U}^{(t)},\widehat{\mathcal{U}}(\mathcal{I})\big] \le D\big[\mathcal{U}^{(t)},\mathcal{U}^*\big] + \Delta(|\mathcal{I}|) \le \sqrt{2}/2 + \Delta(2\widehat{s}) < 0.75 < 1,$$

where we use the fact that $|\mathcal{I}| \le 2\widehat{s}$. Thus, on the right-hand side of (B.43) we obtain

$$\Big[1-D\big[\mathcal{U}^{(t)},\widehat{\mathcal{U}}(\mathcal{I})\big]^2/(2k)\Big]^{-3/2} \le [1-1/(2k)]^{-3/2} \le (1-1/2)^{-3/2} = 2\sqrt{2} < 3,$$

since the function $(1-z)^{-3/2}$ is increasing. Plugging this into the right-hand side of (B.43), we have

$$\frac{D\big[\mathcal{U}^{(t)},\widehat{\mathcal{U}}(\mathcal{I})\big]}{\sqrt{1-D\big[\mathcal{U}^{(t)},\widehat{\mathcal{U}}(\mathcal{I})\big]^2/(2k)}}\cdot\gamma \le \frac{D\big[\mathcal{U}^{(t)},\mathcal{U}^*\big]}{\sqrt{1-D\big[\mathcal{U}^{(t)},\mathcal{U}^*\big]^2/(2k)}}\cdot\gamma + 3\Delta(|\mathcal{I}|)\cdot\gamma.$$

Finally, plugging this into the right-hand side of (B.42), we obtain

$$D\big[\mathcal{V}^{(t+1)},\mathcal{U}^*\big] \le \frac{D\big[\mathcal{U}^{(t)},\mathcal{U}^*\big]}{\sqrt{1-D\big[\mathcal{U}^{(t)},\mathcal{U}^*\big]^2/(2k)}}\cdot\gamma + 3\Delta(|\mathcal{I}|)\cdot\gamma. \tag{B.44}$$

Moreover, since we assume $D\big[\mathcal{U}^{(t)},\mathcal{U}^*\big]\le\sqrt{2}/2$ and $\Delta(|\mathcal{I}|)\le\Delta(2\widehat{s})\le 1/24$, we have

$$\frac{D\big[\mathcal{U}^{(t)},\mathcal{U}^*\big]}{\sqrt{1-D\big[\mathcal{U}^{(t)},\mathcal{U}^*\big]^2/(2k)}}\cdot\gamma + 3\Delta(|\mathcal{I}|)\cdot\gamma \le \frac{\sqrt{2}/2}{\sqrt{1-1/4}} + 3\cdot 1/24 < 0.95 < 1,$$

where we use the fact that $\gamma<1$. By (B.44), it then follows that $D\big[\mathcal{V}^{(t+1)},\mathcal{U}^*\big]<1$. Since $s^*/\widehat{s}\le 1$ by assumption, from Lemma 5.5 we have

$$D\big[\mathcal{U}^*,\mathcal{U}^{(t+1)}\big] \le \left(1+2\sqrt{\frac{k\cdot s^*}{\widehat{s}}}\right)\cdot D\big[\mathcal{U}^*,\mathcal{V}^{(t+1)}\big].$$

Plugging (B.44) into its right-hand side, we obtain

$$D\big[\mathcal{U}^*,\mathcal{U}^{(t+1)}\big] \le \left(1+2\sqrt{\frac{k\cdot s^*}{\widehat{s}}}\right)\cdot\gamma\cdot\frac{D\big[\mathcal{U}^*,\mathcal{U}^{(t)}\big]}{\sqrt{1-D\big[\mathcal{U}^*,\mathcal{U}^{(t)}\big]^2/(2k)}} + \left(1+2\sqrt{\frac{k\cdot s^*}{\widehat{s}}}\right)\cdot 3\Delta(|\mathcal{I}|)\cdot\gamma. \tag{B.45}$$

By our assumption of this lemma we have

$$\widehat{s} = C\left[\frac{4k}{(\gamma^{-1/2}-1)^2}\right]\cdot s^*, \qquad D\big[\mathcal{U}^*,\mathcal{U}^{(t)}\big]\le\sqrt{2k(1-\gamma^{1/2})}$$

with $C\ge 1$, which correspondingly imply

$$1+2\sqrt{\frac{k\cdot s^*}{\widehat{s}}}\le\gamma^{-1/2}, \qquad \frac{1}{\sqrt{1-D\big[\mathcal{U}^*,\mathcal{U}^{(t)}\big]^2/(2k)}}\le\gamma^{-1/4}.$$

Therefore, from (B.45) we obtain

$$D\big[\mathcal{U}^*,\mathcal{U}^{(t+1)}\big] \le \gamma^{1/4}\cdot D\big[\mathcal{U}^*,\mathcal{U}^{(t)}\big] + 3\gamma^{1/2}\cdot\Delta(2\widehat{s}),$$

where we use the fact that $\Delta(|\mathcal{I}|)\le\Delta(2\widehat{s})$ since $|\mathcal{I}|\le 2\widehat{s}$. Thus we conclude the proof. $\square$

## B.5 Proof of Theorem 4.2

*Proof.* Note that we assume the initialization $\mathcal{U}^{\mathrm{init}}$ satisfies

$$D(\mathcal{U}^*,\mathcal{U}^{\mathrm{init}}) \leq \min\left\{\sqrt{\frac{k\gamma(1-\gamma^{1/2})}{2}}, \frac{\sqrt{2\gamma}}{4}\right\} < 1.$$

Recall that in Algorithm 1, we first truncate the initialization $\mathbf{U}^{\mathrm{init}}$ (Line 4) and then obtain $\mathbf{U}^{(T+1)}$. Following the same proof of Lemma 5.5, we have

$$D(\mathcal{U}^*,\mathcal{U}^{(T+1)}) \leq \left(1+2\sqrt{\frac{k\cdot s^*}{\widehat{s}}}\right)\cdot D(\mathcal{U}^*,\mathcal{U}^{\mathrm{init}}). \tag{B.46}$$

Since we take

$$\widehat{s} = C\max\left\{\left\lceil\frac{4k}{(\gamma^{-1/2}-1)^2}\right\rceil, 1\right\}\cdot s^*$$

with $C \geq 1$, on the right-hand side of (B.46) we have

$$1+2\sqrt{\frac{k\cdot s^*}{\widehat{s}}} \leq \gamma^{-1/2},$$

which implies

$$D(\mathcal{U}^*,\mathcal{U}^{(T+1)}) \leq \gamma^{-1/2}\cdot D(\mathcal{U}^*,\mathcal{U}^{\mathrm{init}}) \leq \min\left\{\sqrt{\frac{k(1-\gamma^{1/2})}{2}}, \frac{\sqrt{2}}{4}\right\}. \tag{B.47}$$

Now we prove by mathematical induction that, for $t \geq T+1$,

$$D(\mathcal{U}^*,\mathcal{U}^{(t)}) \leq \gamma^{(t-T-1)/4}\cdot D(\mathcal{U}^*,\mathcal{U}^{(T+1)}) + \frac{3\gamma^{1/2}}{1-\gamma^{1/4}}\cdot\Delta(2\widehat{s}). \tag{B.48}$$

For $t = T+1$, by (B.47) we have

$$D(\mathcal{U}^*,\mathcal{U}^{(T+1)}) \leq 1/2\cdot\min\left\{\sqrt{2k\big[1-\gamma^{1/2}\big]}, \sqrt{2}/2\right\} < \min\left\{\sqrt{2k\big[1-\gamma^{1/2}\big]}, \sqrt{2}/2\right\}.$$

Meanwhile, since $n$ is sufficiently large, we have $\Delta(2\widehat{s}) \leq 1/24$. Hence, the conditions of Lemma 5.6 are satisfied. Therefore we obtain

$$D(\mathcal{U}^*,\mathcal{U}^{(T+2)}) \leq \gamma^{1/4}\cdot D(\mathcal{U}^*,\mathcal{U}^{(T+1)}) + 3\gamma^{1/2}\cdot\Delta(2\widehat{s}) \leq \gamma^{1/4}\cdot D(\mathcal{U}^*,\mathcal{U}^{(T+1)}) + \frac{3\gamma^{1/2}}{1-\gamma^{1/4}}\cdot\Delta(2\widehat{s}),$$

where the last inequality is because $\gamma \in (0,1)$. Suppose that (B.48) holds for $t > T+1$, then on the right-hand side of (B.48), by (B.47) we have

$$\gamma^{(t-T-1)/4}\cdot D(\mathcal{U}^*,\mathcal{U}^{(T+1)}) \leq D(\mathcal{U}^*,\mathcal{U}^{(T+1)}) \leq 1/2\cdot\min\left\{\sqrt{2k\big[1-\gamma^{1/2}\big]}, \sqrt{2}/2\right\},$$

and

$$\frac{3\gamma^{1/2}}{1-\gamma^{1/4}} \cdot \Delta(2\widehat{s}) \le 1/2 \cdot \min\left\{\sqrt{2k\big[1-\gamma^{1/2}\big]}, \sqrt{2}/2\right\},$$

because $\Delta(\cdot)$ is defined in (5.11) and we assume $n$ is sufficiently large. Thus, from (B.48) we have

$$D\big(\mathcal{U}^*, \mathcal{U}^{(t)}\big) \le \min\left\{\sqrt{2k\big[1-\gamma^{1/2}\big]}, \sqrt{2}/2\right\}. \tag{B.49}$$

By Lemma 5.6 we have

$$D\big(\mathcal{U}^*, \mathcal{U}^{(t+1)}\big) \le \gamma^{1/4} \cdot D\big(\mathcal{U}^*, \mathcal{U}^{(t)}\big) + 3\gamma^{1/2} \cdot \Delta(2\widehat{s}).$$

Plugging (B.48) into its right-hand side, we obtain

$$\begin{aligned} D\big(\mathcal{U}^*, \mathcal{U}^{(t+1)}\big) &\le \gamma^{1/4} \cdot \left[\gamma^{(t-T-1)/4} \cdot D\big(\mathcal{U}^*, \mathcal{U}^{(T+1)}\big) + \frac{3\gamma^{1/2}}{1-\gamma^{1/4}} \cdot \Delta(2\widehat{s})\right] + 3\gamma^{1/2} \cdot \Delta(2\widehat{s}) \\ &\le \gamma^{(t+1-T-1)/4} \cdot D\big(\mathcal{U}^*, \mathcal{U}^{(T+1)}\big) + \frac{3\gamma^{1/2}}{1-\gamma^{1/4}} \cdot \Delta(2\widehat{s}). \end{aligned}$$

By mathematical induction, we have that (B.48) holds for $t = T+1, T+2, \ldots$.

Meanwhile, since we take

$$\widehat{s} = C \max\left\{\left\lceil \frac{4k}{(\gamma^{-1/2}-1)^2}\right\rceil, 1\right\} \cdot s^*$$

with $C \ge 1$, plugging the definition of $\Delta(\cdot)$ in (5.11) into (B.48), we obtain

$$D\big(\mathcal{U}^*, \mathcal{U}^{(t)}\big) \le \gamma^{(t-T-1)/4} \cdot D\big(\mathcal{U}^*, \mathcal{U}^{(T+1)}\big) + \underbrace{\frac{C}{(1-\gamma^{1/4})(1-\gamma^{1/2})}}_{\text{(i)}} \cdot \frac{\sqrt{\lambda_1 \lambda_{k+1}}}{\lambda_k - \lambda_{k+1}} \cdot \sqrt{\frac{k \cdot s^* \cdot (k + \log d)}{n}}, \tag{B.50}$$

where we use the fact that $\gamma < 1$. For term (i), we have

$$(1-\gamma^{1/4})(1-\gamma^{1/2}) \ge (1-\gamma^{1/4})^2 \ge [(1-\gamma)/4]^2.$$

Plugging the definition of $\gamma$ in (1.4) into the right-hand side of the above inequality, we obtain

$$(1-\gamma^{1/4})(1-\gamma^{1/2}) \ge C \cdot \left(\frac{\lambda_k - \lambda_{k+1}}{\lambda_k}\right)^2. \tag{B.51}$$

Plugging (B.51) into (B.50) and the upper bound of $D\big(\mathcal{U}^*, \mathcal{U}^{(T+1)}\big)$ in (B.47), we obtain

$$\begin{aligned} &D\big(\mathcal{U}^*, \mathcal{U}^{(t)}\big) \\ &\le \gamma^{(t-T-1)/4} \cdot \min\left\{\sqrt{\frac{k(1-\gamma^{1/2})}{2}}, \sqrt{2}/4\right\} + C\left(\frac{\lambda_k - \lambda_{k+1}}{\lambda_k}\right)^2 \cdot \frac{\sqrt{\lambda_1 \lambda_{k+1}}}{\lambda_k - \lambda_{k+1}} \cdot \sqrt{\frac{k \cdot s^* \cdot (k + \log d)}{n}}. \end{aligned}$$

Note that this result is obtained under Assumption 5.1, which is shown by Lemma 5.2 and Lemma 5.3 to hold with probability at least $1 - 4/(n-1) - 1/d - 6\log n/n - 1/C'$, where $C'$ is the sufficiently large constant in Lemma 5.3. Hence, this result holds with the same probability, which concludes the proof. □

# C Proof for the "Relax" Stage

We present the detailed proof of Theorem 4.3 and the corresponding auxiliary lemmas.

## C.1 An Auxiliary Lemma

Before we present the proof of Theorem 4.3, we first lay out a lemma that gives a useful property of $\mathbf{Z}$ and $h(\mathbf{Z})$ defined in (5.13).

**Lemma C.1.** For $\mathbf{Z}_1$ and $\mathbf{Z}_2$, we have

$$\big\langle \mathbf{Z}_1 - \mathbf{Z}_2, h(\mathbf{Z}_1)\big\rangle = \big\langle \mathbf{Z}_1 - \mathbf{Z}_2, h(\mathbf{Z}_2)\big\rangle. \tag{C.1}$$

*Proof.* By definition we have

$$\begin{aligned}\big\langle \mathbf{Z}_1 - \mathbf{Z}_2, h(\mathbf{Z}_1)\big\rangle &= -\big\langle \mathbf{\Pi}_1 - \mathbf{\Pi}_2, \mathbf{\Theta}_1\big\rangle + \big\langle \mathbf{\Phi}_1 - \mathbf{\Phi}_2, \mathbf{\Theta}_1\big\rangle + \big\langle \mathbf{\Theta}_1 - \mathbf{\Theta}_2, \mathbf{\Pi}_1 - \mathbf{\Phi}_1\big\rangle \\ &= \big\langle \mathbf{\Pi}_2, \mathbf{\Theta}_1\big\rangle - \big\langle \mathbf{\Phi}_2, \mathbf{\Theta}_1\big\rangle - \big\langle \mathbf{\Pi}_1, \mathbf{\Theta}_2\big\rangle + \big\langle \mathbf{\Phi}_1, \mathbf{\Theta}_2\big\rangle,\end{aligned}$$

while

$$\begin{aligned}\big\langle \mathbf{Z}_1 - \mathbf{Z}_2, h(\mathbf{Z}_2)\big\rangle &= -\big\langle \mathbf{\Pi}_1 - \mathbf{\Pi}_2, \mathbf{\Theta}_2\big\rangle + \big\langle \mathbf{\Phi}_1 - \mathbf{\Phi}_2, \mathbf{\Theta}_2\big\rangle + \big\langle \mathbf{\Theta}_1 - \mathbf{\Theta}_2, \mathbf{\Pi}_2 - \mathbf{\Phi}_2\big\rangle \\ &= -\big\langle \mathbf{\Pi}_1, \mathbf{\Theta}_2\big\rangle + \big\langle \mathbf{\Phi}_1, \mathbf{\Theta}_2\big\rangle - \big\langle \mathbf{\Phi}_2, \mathbf{\Theta}_1\big\rangle + \big\langle \mathbf{\Pi}_2, \mathbf{\Theta}_1\big\rangle.\end{aligned}$$

Thus we conclude the proof. □

## C.2 Proof of Theorem 4.3

*Proof.* Remind that $\mathbf{\Pi}^*$ denotes the projection matrix corresponding to the true principal subspace $\mathcal{U}^*$. For notational simplicity, we define

$$\mathbf{Z}^* = \begin{bmatrix} \mathbf{\Pi}^* \\ \mathbf{\Pi}^* \\ \widetilde{\mathbf{\Theta}}^{(t)} \end{bmatrix}, \qquad \text{where} \;\; \widetilde{\mathbf{\Theta}}^{(t)} = -\rho \cdot \operatorname{sign}\big(\overline{\mathbf{\Pi}}^{(t)} - \overline{\mathbf{\Phi}}^{(t)}\big). \tag{C.2}$$

We consider the following key quantity throughout this proof

$$-\big\langle \widehat{\mathbf{\Sigma}}, \overline{\mathbf{\Pi}}^{(t)} - \mathbf{\Pi}^*\big\rangle + \rho\big\|\overline{\mathbf{\Phi}}^{(t)}\big\|_{1,1} - \rho\|\mathbf{\Pi}^*\|_{1,1} + \big\langle \overline{\mathbf{Z}}^{(t)} - \mathbf{Z}^*, h\big(\overline{\mathbf{Z}}^{(t)}\big)\big\rangle.$$

By the definition in (5.14) and the convexity of $\|\cdot\|_{1,1}$, we have

$$\begin{aligned}& -\big\langle \widehat{\mathbf{\Sigma}}, \overline{\mathbf{\Pi}}^{(t)} - \mathbf{\Pi}^*\big\rangle + \rho\big\|\overline{\mathbf{\Phi}}^{(t)}\big\|_{1,1} - \rho\|\mathbf{\Pi}^*\|_{1,1} + \big\langle \overline{\mathbf{Z}}^{(t)} - \mathbf{Z}^*, h\big(\overline{\mathbf{Z}}^{(t)}\big)\big\rangle \\ & \leq \frac{1}{t}\sum_{j=1}^{t}\Big\{-\big\langle \widehat{\mathbf{\Sigma}}, \mathbf{\Pi}^{(j)} - \mathbf{\Pi}^*\big\rangle + \rho\big\|\mathbf{\Phi}^{(j)}\big\|_{1,1} - \rho\|\mathbf{\Pi}^*\|_{1,1} + \big\langle \mathbf{Z}^{(j)} - \mathbf{Z}^*, h\big(\mathbf{Z}^{(j)}\big)\big\rangle\Big\}.\end{aligned} \tag{C.3}$$

Plugging the definition of $h(\cdot)$ in (5.13) into (C.3), on the right-hand side of (C.3) we have

$$
\begin{aligned}
&-\big\langle \widehat{\mathbf{\Sigma}}, \mathbf{\Pi}^{(j)} - \mathbf{\Pi}^{*}\big\rangle + \rho\big\|\mathbf{\Phi}^{(j)}\big\|_{1,1} - \rho\|\mathbf{\Pi}^{*}\|_{1,1} + \big\langle \mathbf{Z}^{(j)} - \mathbf{Z}^{*}, h\big(\mathbf{Z}^{(j)}\big)\big\rangle \\
&= \overbrace{-\big\langle \widehat{\mathbf{\Sigma}}, \mathbf{\Pi}^{(j)} - \mathbf{\Pi}^{*}\big\rangle + \big\langle \mathbf{\Pi}^{(j)} - \mathbf{\Pi}^{*}, -\mathbf{\Theta}^{(j)}\big\rangle}^{\text{(i)}} + \overbrace{\rho\big\|\mathbf{\Phi}^{(j)}\big\|_{1,1} - \rho\|\mathbf{\Pi}^{*}\|_{1,1} + \big\langle \mathbf{\Phi}^{(j)} - \mathbf{\Pi}^{*}, \mathbf{\Theta}^{(j)}\big\rangle}^{\text{(ii)}} \\
&\quad + \underbrace{\big\langle \mathbf{\Theta}^{(j)} - \widetilde{\mathbf{\Theta}}^{(t)}, \mathbf{\Pi}^{(j)} - \mathbf{\Phi}^{(j)}\big\rangle}_{\text{(iii)}}.
\end{aligned} \tag{C.4}
$$

Now we provide upper bounds for the terms in (C.4) respectively:

- For term (i) in (C.4), we have

$$
\begin{aligned}
&-\big\langle \widehat{\mathbf{\Sigma}}, \mathbf{\Pi}^{(j)} - \mathbf{\Pi}^{*}\big\rangle + \big\langle \mathbf{\Pi}^{(j)} - \mathbf{\Pi}^{*}, -\mathbf{\Theta}^{(j)}\big\rangle \\
&= -\big\langle \widehat{\mathbf{\Sigma}}, \mathbf{\Pi}^{(j)} - \mathbf{\Pi}^{*}\big\rangle + \big\langle \mathbf{\Pi}^{(j)} - \mathbf{\Pi}^{*}, \beta\big(\mathbf{\Pi}^{(j)} - \mathbf{\Phi}^{(j-1)}\big) - \mathbf{\Theta}^{(j-1)}\big\rangle + \big\langle \mathbf{\Pi}^{(j)} - \mathbf{\Pi}^{*}, \beta\big(\mathbf{\Phi}^{(j-1)} - \mathbf{\Phi}^{(j)}\big)\big\rangle, \\
&= \underbrace{\big\langle -\widehat{\mathbf{\Sigma}} - \mathbf{\Theta}^{(j-1)} + \beta\big(\mathbf{\Pi}^{(j)} - \mathbf{\Phi}^{(j-1)}\big), \mathbf{\Pi}^{(j)} - \mathbf{\Pi}^{*}\big\rangle}_{\text{(i).a}} + \big\langle \mathbf{\Pi}^{(j)} - \mathbf{\Pi}^{*}, \beta\big(\mathbf{\Phi}^{(j-1)} - \mathbf{\Phi}^{(j)}\big)\big\rangle.
\end{aligned} \tag{C.5}
$$

  Here for the first equality we utilize the fact that $\mathbf{\Theta}^{(j)} = \mathbf{\Theta}^{(j-1)} - \beta\big(\mathbf{\Pi}^{(j)} - \mathbf{\Phi}^{(j)}\big)$, according to Line 8 of Algorithm 3. For term (i).a, from the first-order optimality condition for Line 6 (Algorithm 3), we have that, for the $j$-th iteration of Algorithm 3,

$$
\big\langle -\widehat{\mathbf{\Sigma}} - \mathbf{\Theta}^{(j-1)} + \beta\big(\mathbf{\Pi}^{(j)} - \mathbf{\Phi}^{(j-1)}\big), \mathbf{\Pi} - \mathbf{\Pi}^{(j)}\big\rangle \geq 0, \quad \text{for all } \ \mathbf{\Pi} \in \mathcal{A}. \tag{C.6}
$$

  Plugging $\mathbf{\Pi} = \mathbf{\Pi}^{*}$ into (C.6), for term (i).a in (C.5), we obtain

$$
\big\langle -\widehat{\mathbf{\Sigma}} - \mathbf{\Theta}^{(j-1)} + \beta\big(\mathbf{\Pi}^{(j)} - \mathbf{\Phi}^{(j-1)}\big), \mathbf{\Pi}^{(j)} - \mathbf{\Pi}^{*}\big\rangle \leq 0.
$$

  Therefore from (C.5) we obtain

$$
-\big\langle \widehat{\mathbf{\Sigma}}, \mathbf{\Pi}^{(j)} - \mathbf{\Pi}^{*}\big\rangle + \big\langle \mathbf{\Pi}^{(j)} - \mathbf{\Pi}^{*}, -\mathbf{\Theta}^{(j)}\big\rangle \leq \big\langle \mathbf{\Pi}^{(j)} - \mathbf{\Pi}^{*}, \beta\big(\mathbf{\Phi}^{(j-1)} - \mathbf{\Phi}^{(j)}\big)\big\rangle. \tag{C.7}
$$

  Now we construct an upper bound for the right-hand side of (C.7). Note that

$$
\begin{aligned}
&\big\langle \mathbf{\Pi}^{(j)} - \mathbf{\Pi}^{*}, \beta\big(\mathbf{\Phi}^{(j-1)} - \mathbf{\Phi}^{(j)}\big)\big\rangle \\
&= \frac{\beta}{2}\Big[\big\|\mathbf{\Pi}^{*} - \mathbf{\Phi}^{(j-1)}\big\|_{\mathrm{F}}^{2} - \big\|\mathbf{\Pi}^{*} - \mathbf{\Phi}^{(j)}\big\|_{\mathrm{F}}^{2}\Big] + \frac{\beta}{2}\Big[\big\|\mathbf{\Pi}^{(j)} - \mathbf{\Phi}^{(j)}\big\|_{\mathrm{F}}^{2} - \big\|\mathbf{\Pi}^{(j)} - \mathbf{\Phi}^{(j-1)}\big\|_{\mathrm{F}}^{2}\Big] \\
&\leq \frac{\beta}{2}\Big[\big\|\mathbf{\Pi}^{*} - \mathbf{\Phi}^{(j-1)}\big\|_{\mathrm{F}}^{2} - \big\|\mathbf{\Pi}^{*} - \mathbf{\Phi}^{(j)}\big\|_{\mathrm{F}}^{2}\Big] + \frac{1}{2\beta}\big\|\mathbf{\Theta}^{(j)} - \mathbf{\Theta}^{(j-1)}\big\|_{\mathrm{F}}^{2}.
\end{aligned}
$$

  Here the equality could be easily verified. For the inequality we use the fact $\mathbf{\Theta}^{(j-1)} - \mathbf{\Theta}^{(j)} = \beta\big(\mathbf{\Pi}^{(j)} - \mathbf{\Phi}^{(j)}\big)$, which follows from Line 8 of Algorithm 3.

- For term (ii) in (C.4), from the first-order optimality condition for Line 7 of Algorithm 3, we have

$$\rho\|\mathbf{\Phi}\|_{1,1} - \rho\big\|\mathbf{\Phi}^{(j)}\big\|_{1,1} + \big\langle \mathbf{\Phi} - \mathbf{\Phi}^{(j)}, \mathbf{\Theta}^{(j)} \big\rangle \geq 0, \quad \text{for all} \;\; \mathbf{\Phi} \in \mathcal{B}. \tag{C.8}$$

Plugging $\mathbf{\Phi} = \mathbf{\Pi}^*$ into (C.8) and negating both sides, we obtain

$$\rho\big\|\mathbf{\Phi}^{(j)}\big\|_{1,1} - \rho\|\mathbf{\Pi}^*\|_{1,1} + \big\langle \mathbf{\Phi}^{(j)} - \mathbf{\Pi}^*, \mathbf{\Theta}^{(j)} \big\rangle \leq 0. \tag{C.9}$$

- For term (iii) in (C.4), we have

$$\begin{aligned}\big\langle \mathbf{\Theta}^{(j)} - \widetilde{\mathbf{\Theta}}^{(t)}, \mathbf{\Pi}^{(j)} - \mathbf{\Phi}^{(j)} \big\rangle &= \frac{1}{\beta}\big\langle \mathbf{\Theta}^{(j)} - \widetilde{\mathbf{\Theta}}^{(t)}, \mathbf{\Theta}^{(j-1)} - \mathbf{\Theta}^{(j)} \big\rangle \\ &= \frac{1}{2\beta}\Big[\big\|\widetilde{\mathbf{\Theta}}^{(t)} - \mathbf{\Theta}^{(j-1)}\big\|_{\mathrm{F}}^2 - \big\|\widetilde{\mathbf{\Theta}}^{(t)} - \mathbf{\Theta}^{(j)}\big\|_{\mathrm{F}}^2 - \big\|\mathbf{\Theta}^{(j)} - \mathbf{\Theta}^{(j-1)}\big\|_{\mathrm{F}}^2\Big],\end{aligned}$$

  where for the first equality we use the fact that $\mathbf{\Theta}^{(j-1)} - \mathbf{\Theta}^{(j)} = \beta\big(\mathbf{\Pi}^{(j)} - \mathbf{\Phi}^{(j)}\big)$.

Plugging the above upper bounds into the right-hand side of (C.4), and then plugging (C.4) into (C.3), we obtain

$$\begin{aligned}&\overbrace{-\big\langle \widehat{\mathbf{\Sigma}}, \mathbf{\Pi}^* - \overline{\mathbf{\Pi}}^{(t)} \big\rangle}^{\text{(i)}} + \overbrace{\rho\|\mathbf{\Pi}^*\|_1 - \rho\big\|\overline{\mathbf{\Phi}}^{(t)}\big\|_1}^{\text{(ii)}} + \overbrace{\big\langle \mathbf{Z}^* - \overline{\mathbf{Z}}^{(t)}, h\big(\overline{\mathbf{Z}}^{(t)}\big)\big\rangle}^{\text{(iii)}} \\ &\geq -\frac{1}{t}\left[\frac{\beta}{2}\big\|\mathbf{\Pi}^* - \mathbf{\Phi}^{(0)}\big\|_{\mathrm{F}}^2 + \frac{1}{2\beta}\big\|\widetilde{\mathbf{\Theta}}^{(t)} - \mathbf{\Theta}^{(0)}\big\|_{\mathrm{F}}^2\right] = -\frac{1}{t}\left[\frac{\beta}{2}\|\mathbf{\Pi}^*\|_{\mathrm{F}}^2 + \frac{1}{2\beta}\big\|\widetilde{\mathbf{\Theta}}^{(t)}\big\|_{\mathrm{F}}^2\right].\end{aligned} \tag{C.10}$$

Here we use the fact that $\mathbf{\Phi}^{(0)} = \mathbf{\Theta}^{(0)} = \mathbf{0}$. Note that here we negate the both sides of (C.3) for the convenience of manipulation. Now we provide upper bounds for the terms in (C.10):

- For term (i) in (C.10), we have

$$-\big\langle \widehat{\mathbf{\Sigma}}, \mathbf{\Pi}^* - \overline{\mathbf{\Pi}}^{(t)} \big\rangle = \underbrace{-\big\langle \mathbf{\Sigma}, \mathbf{\Pi}^* - \overline{\mathbf{\Pi}}^{(t)} \big\rangle}_{\text{(i).a}} + \underbrace{\big\langle \mathbf{\Sigma} - \widehat{\mathbf{\Sigma}}, \mathbf{\Pi}^* - \overline{\mathbf{\Pi}}^{(t)} \big\rangle}_{\text{(i).b}}.$$

  According to the curvature property in Lemma 3.1 of Vu et al. (2013), term (i).a satisfies

$$-\big\langle \mathbf{\Sigma}, \mathbf{\Pi}^* - \overline{\mathbf{\Pi}}^{(t)} \big\rangle \leq -\frac{\lambda_k - \lambda_{k+1}}{2}\big\|\mathbf{\Pi}^* - \overline{\mathbf{\Pi}}^{(t)}\big\|_{\mathrm{F}}^2. \tag{C.11}$$

  By Cauchy-Schwarz inequality, term (ii).a satisfies

$$\big\langle \mathbf{\Sigma} - \widehat{\mathbf{\Sigma}}, \mathbf{\Pi}^* - \overline{\mathbf{\Pi}}^{(t)} \big\rangle = \big\|\mathbf{\Sigma} - \widehat{\mathbf{\Sigma}}\big\|_{\infty,\infty} \cdot \big\|\mathbf{\Pi}^* - \overline{\mathbf{\Pi}}^{(t)}\big\|_{1,1} \leq \rho\big\|\mathbf{\Pi}^* - \overline{\mathbf{\Pi}}^{(t)}\big\|_{1,1}. \tag{C.12}$$

  Here $\|\cdot\|_{\infty,\infty}$ gives the maximum absolute value of all entries. Here the last inequality follows from Assumption 5.7. Adding (C.11) and (C.12), we obtain

$$-\big\langle \widehat{\mathbf{\Sigma}}, \mathbf{\Pi}^* - \overline{\mathbf{\Pi}}^{(t)} \big\rangle \leq -\frac{\lambda_k - \lambda_{k+1}}{2}\big\|\mathbf{\Pi}^* - \overline{\mathbf{\Pi}}^{(t)}\big\|_{\mathrm{F}}^2 + \rho\big\|\mathbf{\Pi}^* - \overline{\mathbf{\Pi}}^{(t)}\big\|_{1,1}. \tag{C.13}$$

- For term (ii) in (C.10), we have

$$\begin{aligned}\rho\|\mathbf{\Pi}^*\|_{1,1} - \rho\|\overline{\mathbf{\Phi}}^{(t)}\|_{1,1} &= \rho\|\mathbf{\Pi}^*\|_{1,1} - \rho\|\overline{\mathbf{\Pi}}^{(t)} + \overline{\mathbf{\Phi}}^{(t)} - \overline{\mathbf{\Pi}}^{(t)}\|_{1,1} \\ &\le \rho\|\mathbf{\Pi}^*\|_{1,1} - \rho\|\overline{\mathbf{\Pi}}^{(t)}\|_{1,1} + \rho\|\overline{\mathbf{\Phi}}^{(t)} - \overline{\mathbf{\Pi}}^{(t)}\|_{1,1}. \end{aligned} \tag{C.14}$$

- For term (iii) in (C.10), we have

$$\begin{aligned}\langle \mathbf{Z}^* - \overline{\mathbf{Z}}^{(t)}, h(\overline{\mathbf{Z}}^{(t)})\rangle &= -\langle \mathbf{\Pi}^* - \overline{\mathbf{\Pi}}^{(t)}, \overline{\mathbf{\Theta}}^{(t)}\rangle + \langle \mathbf{\Pi}^* - \overline{\mathbf{\Phi}}^{(t)}, \overline{\mathbf{\Theta}}^{(t)}\rangle + \langle \widetilde{\mathbf{\Theta}}^{(t)} - \overline{\mathbf{\Theta}}^{(t)}, \overline{\mathbf{\Pi}}^{(t)} - \overline{\mathbf{\Phi}}^{(t)}\rangle \\ &= \langle \widetilde{\mathbf{\Theta}}^{(t)}, \overline{\mathbf{\Pi}}^{(t)} - \overline{\mathbf{\Phi}}^{(t)}\rangle \\ &= -\rho\|\overline{\mathbf{\Pi}}^{(t)} - \overline{\mathbf{\Phi}}^{(t)}\|_{1,1}. \end{aligned} \tag{C.15}$$

  Here the last equality follows from the definition of $\widetilde{\mathbf{\Theta}}^{(t)}$ in (C.2).

Plugging (C.13)-(C.15) into the left-hand side of (C.10), we obtain

$$\begin{aligned}&-\frac{\lambda_k - \lambda_{k+1}}{2}\|\mathbf{\Pi}^* - \overline{\mathbf{\Pi}}^{(t)}\|_{\mathrm{F}}^2 + \overbrace{\rho\|\mathbf{\Pi}^* - \overline{\mathbf{\Pi}}^{(t)}\|_{1,1}}^{\text{(i)}} + \overbrace{\rho\|\mathbf{\Pi}^*\|_{1,1}}^{\text{(ii)}} - \overbrace{\rho\|\overline{\mathbf{\Pi}}^{(t)}\|_{1,1}}^{\text{(iii)}} \\ &\ge -\frac{1}{t}\left[\frac{\beta}{2}\|\mathbf{\Pi}^*\|_{\mathrm{F}}^2 + \frac{1}{2\beta}\|\widetilde{\mathbf{\Theta}}^{(t)}\|_{\mathrm{F}}^2\right]. \end{aligned} \tag{C.16}$$

Now we derive upper bounds for the terms in (C.16). We define $\mathcal{D}$ to be the index set of nonzero entries of $\mathbf{\Pi}^*$, i.e.,

$$\mathcal{D} = \left\{(i,j) : \mathbf{\Pi}^*_{i,j} \neq 0\right\}.$$

For a matrix $\mathbf{\Pi} \in \mathbb{R}^{d\times d}$, we define $\mathbf{\Pi}(\mathcal{D}) \in \mathbb{R}^{d\times d}$ and $\mathbf{\Pi}(\overline{\mathcal{D}}) \in \mathbb{R}^{d\times d}$ as

$$\left[\mathbf{\Pi}(\mathcal{D})\right]_{i,j} = \mathbb{1}\left[(i,j)\in\mathcal{D}\right]\cdot\mathbf{\Pi}_{i,j}, \quad \left[\mathbf{\Pi}(\overline{\mathcal{D}})\right]_{i,j} = \mathbb{1}\left[(i,j)\notin\mathcal{D}\right]\cdot\mathbf{\Pi}_{i,j}, \quad \text{for } \ i,j = 1,\ldots,d,$$

where $\mathbb{1}[\cdot]$ is the indicator function. Then we have:

- For term (i) in (C.16), we have

$$\begin{aligned}\rho\|\mathbf{\Pi}^* - \overline{\mathbf{\Pi}}^{(t)}\|_{1,1} &= \rho\|\mathbf{\Pi}^*(\mathcal{D}) - \overline{\mathbf{\Pi}}^{(t)}(\mathcal{D})\|_{1,1} + \rho\|\mathbf{\Pi}^*(\overline{\mathcal{D}}) - \overline{\mathbf{\Pi}}^{(t)}(\overline{\mathcal{D}})\|_{1,1} \\ &= \rho\|\mathbf{\Pi}^*(\mathcal{D}) - \overline{\mathbf{\Pi}}^{(t)}(\mathcal{D})\|_{1,1} + \rho\|\overline{\mathbf{\Pi}}^{(t)}(\overline{\mathcal{D}})\|_{1,1}, \end{aligned}$$

  where we use the fact that $\mathbf{\Pi}^*(\overline{\mathcal{D}}) = \mathbf{0}$.

- For term (ii) in (C.16), we have

$$\rho\|\mathbf{\Pi}^*\|_{1,1} = \rho\|\mathbf{\Pi}^*(\mathcal{D})\|_{1,1}.$$

- For term (iii) in (C.16), we have

$$\rho\|\overline{\mathbf{\Pi}}^{(t)}\|_{1,1} = \rho\|\overline{\mathbf{\Pi}}^{(t)}(\mathcal{D})\|_{1,1} + \rho\|\overline{\mathbf{\Pi}}^{(t)}(\overline{\mathcal{D}})\|_{1,1}.$$

Combining the above equalities, on the left-hand side of (C.16) we have

$$\begin{aligned}&\rho\big\|\boldsymbol{\Pi}^* - \overline{\boldsymbol{\Pi}}^{(t)}\big\|_{1,1} + \rho\big\|\boldsymbol{\Pi}^*\big\|_{1,1} - \rho\big\|\overline{\boldsymbol{\Pi}}^{(t)}\big\|_{1,1}\\ &= \rho\big\|\boldsymbol{\Pi}^*(\mathcal{D}) - \overline{\boldsymbol{\Pi}}^{(t)}(\mathcal{D})\big\|_{1,1} + \rho\big\|\boldsymbol{\Pi}^*(\mathcal{D})\big\|_{1,1} - \rho\big\|\overline{\boldsymbol{\Pi}}^{(t)}(\mathcal{D})\big\|_{1,1}.\end{aligned}$$

Applying triangular inequality to the last two terms on the right-hand side of the above equality, we obtain

$$\begin{aligned}\rho\big\|\boldsymbol{\Pi}^* - \overline{\boldsymbol{\Pi}}^{(t)}\big\|_{1,1} + \rho\big\|\boldsymbol{\Pi}^*\big\|_{1,1} - \rho\big\|\overline{\boldsymbol{\Pi}}^{(t)}\big\|_{1,1} &\leq 2\rho\big\|\boldsymbol{\Pi}^*(\mathcal{D}) - \overline{\boldsymbol{\Pi}}^{(t)}(\mathcal{D})\big\|_{1,1}\\ &\leq 2\sqrt{|\mathcal{D}|}\cdot\rho\big\|\boldsymbol{\Pi}^*(\mathcal{D}) - \overline{\boldsymbol{\Pi}}^{(t)}(\mathcal{D})\big\|_{F}. \end{aligned} \tag{C.17}$$

where on the right-hand side we have $|\mathcal{D}| \leq (s^*)^2$ according to our discussion on subspace sparsity in §2. Moreover, we have $\big\|\boldsymbol{\Pi}^*(\mathcal{D}) - \overline{\boldsymbol{\Pi}}^{(t)}(\mathcal{D})\big\|_F \leq \big\|\boldsymbol{\Pi}^* - \overline{\boldsymbol{\Pi}}^{(t)}\big\|_F$. Plugging it into (C.17), and then plugging (C.17) into (C.16), we obtain

$$-\frac{\lambda_k - \lambda_{k+1}}{2}\big\|\boldsymbol{\Pi}^* - \overline{\boldsymbol{\Pi}}^{(t)}\big\|_{\mathrm{F}}^2 + 2s^*\rho\big\|\boldsymbol{\Pi}^* - \overline{\boldsymbol{\Pi}}^{(t)}\big\|_F \geq -\frac{1}{t}\left[\frac{\beta}{2}\big\|\boldsymbol{\Pi}^*\big\|_{\mathrm{F}}^2 + \frac{1}{2\beta}\big\|\widetilde{\boldsymbol{\Theta}}^{(t)}\big\|_{\mathrm{F}}^2\right]. \tag{C.18}$$

Solving for $\big\|\boldsymbol{\Pi}^* - \overline{\boldsymbol{\Pi}}^{(t)}\big\|_{\mathrm{F}}$ in the quadratic inequality in (C.18), with further manipulation we obtain

$$\big\|\boldsymbol{\Pi}^* - \overline{\boldsymbol{\Pi}}^{(t)}\big\|_{\mathrm{F}} \leq \frac{4s^*\rho}{\lambda_k - \lambda_{k+1}} + \frac{\sqrt{\beta}\cdot\|\boldsymbol{\Pi}^*\|_{\mathrm{F}} + \|\widetilde{\boldsymbol{\Theta}}^{(t)}\|_{\mathrm{F}}/\sqrt{\beta}}{\sqrt{\lambda_k - \lambda_{k+1}}}\cdot\frac{1}{\sqrt{t}}. \tag{C.19}$$

By the definition of $\widetilde{\boldsymbol{\Theta}}^{(t)}$ in (C.2), we have $\|\widetilde{\boldsymbol{\Theta}}^{(t)}\|_{\mathrm{F}} \leq \rho\cdot d$. Also, since $\boldsymbol{\Pi}^*$ has $k$ nonzero eigenvalues of one, we have $\|\boldsymbol{\Pi}^*\|_{\mathrm{F}} = \sqrt{k}$. Plugging these into the right-hand side of (C.19), and then optimizing it with respect to the penalty parameter $\beta > 0$, we obtain

$$\big\|\boldsymbol{\Pi}^* - \overline{\boldsymbol{\Pi}}^{(t)}\big\|_{\mathrm{F}} \leq \frac{C\lambda_1}{\lambda_k - \lambda_{k+1}}\cdot s^*\sqrt{\frac{\log d}{n}} + \frac{C'\sqrt{\lambda_1}}{\sqrt{\lambda_k - \lambda_{k+1}}}\cdot\left(\frac{k\cdot d^2\log d}{n}\right)^{1/4}\cdot\frac{1}{\sqrt{t}} \tag{C.20}$$

where we use $\rho = C''\lambda_1\sqrt{\log d/n}$. Finally, by Corollary 3.2 of Vu et al. (2013), we have

$$D(\mathcal{U}^*, \mathcal{U}^{(t)}) \leq 2\big\|\boldsymbol{\Pi}^* - \overline{\boldsymbol{\Pi}}^{(t)}\big\|_{\mathrm{F}},$$

and therefore $D(\mathcal{U}^*, \mathcal{U}^{(t)})$ is upper bounded by the right-hand side of (C.20) up to constant. Remind that the entire proof is relies on Assumption 5.7, and Lemma 5.8 shows that Assumption 5.7 holds with high probability. Thus, we reach the conclusion. □

# D Proof for Non-Gaussian and Dependent Data

We present the detailed proof of an auxiliary lemma in §5.3.

### D.1 Proof of Lemma 5.11

*Proof.* Remind we abbreviate $\lambda_j(\boldsymbol{\Sigma})$ to be $\lambda_j$ ($j = 1, \ldots, d$). Let $(\mathbf{U}^*)^{\perp}$ be the orthonormal matrix whose columns are the eigenvectors corresponding to $\lambda_{k+1}, \ldots, \lambda_d$ and $\widehat{\mathbf{U}}(\mathcal{I})^{\perp}$ be the orthonormal matrix whose columns are the eigenvectors corresponding to $\lambda_{k+1}\big(\widehat{\boldsymbol{\Sigma}}_{\mathcal{I}}\big), \ldots, \lambda_d\big(\widehat{\boldsymbol{\Sigma}}_{\mathcal{I}}\big)$.

From the definition of $\mathbf{U}^*$ and $\widehat{\mathbf{U}}(\mathcal{I})^{\perp}$, we have

$$\boldsymbol{\Sigma} \cdot \mathbf{U}^* = \mathbf{U}^* \cdot \Lambda_0(\boldsymbol{\Sigma}), \qquad \widehat{\boldsymbol{\Sigma}}_{\mathcal{I}} \cdot \widehat{\mathbf{U}}(\mathcal{I})^{\perp} = \widehat{\mathbf{U}}(\mathcal{I})^{\perp} \cdot \Lambda_1\big(\widehat{\boldsymbol{\Sigma}}_{\mathcal{I}}\big), \tag{D.1}$$

where $\Lambda_0(\boldsymbol{\Sigma})$ and $\Lambda_1\big(\widehat{\boldsymbol{\Sigma}}\big)$ are defined as

$$\Lambda_0(\boldsymbol{\Sigma}) = \begin{bmatrix} \lambda_1 & \cdots & 0 \\ \vdots & \ddots & \vdots \\ 0 & \cdots & \lambda_k \end{bmatrix}, \qquad \Lambda_1\big(\widehat{\boldsymbol{\Sigma}}_{\mathcal{I}}\big) = \begin{bmatrix} \lambda_{k+1}\big(\widehat{\boldsymbol{\Sigma}}_{\mathcal{I}}\big) & \cdots & 0 \\ \vdots & \ddots & \vdots \\ 0 & \cdots & \lambda_d\big(\widehat{\boldsymbol{\Sigma}}_{\mathcal{I}}\big) \end{bmatrix}.$$

Hence we have

$$\begin{aligned} (\mathbf{U}^*)^T\big(\widehat{\boldsymbol{\Sigma}} - \boldsymbol{\Sigma}\big)\widehat{\mathbf{U}}(\mathcal{I})^{\perp} &= (\mathbf{U}^*)^T\widehat{\boldsymbol{\Sigma}} \cdot \widehat{\mathbf{U}}(\mathcal{I})^{\perp} - (\boldsymbol{\Sigma} \cdot \mathbf{U}^*)^T\widehat{\mathbf{U}}(\mathcal{I})^{\perp} \\ &= (\mathbf{U}^*)^T\widehat{\mathbf{U}}(\mathcal{I})^{\perp} \cdot \Lambda_1\big(\widehat{\boldsymbol{\Sigma}}_{\mathcal{I}}\big) - \Lambda_0(\boldsymbol{\Sigma}) \cdot (\mathbf{U}^*)^T\widehat{\mathbf{U}}(\mathcal{I})^{\perp}, \end{aligned} \tag{D.2}$$

where the second equality follows from (D.1). Consequently, we have

$$\begin{aligned} \big\|(\mathbf{U}^*)^T\big(\widehat{\boldsymbol{\Sigma}} - \boldsymbol{\Sigma}\big)\widehat{\mathbf{U}}(\mathcal{I})^{\perp}\big\|_2 &= \big\|\Lambda_0(\boldsymbol{\Sigma}) \cdot (\mathbf{U}^*)^T\widehat{\mathbf{U}}(\mathcal{I})^{\perp} - (\mathbf{U}^*)^T\widehat{\mathbf{U}}(\mathcal{I})^{\perp} \cdot \Lambda_1\big(\widehat{\boldsymbol{\Sigma}}_{\mathcal{I}}\big)\big\|_2 \\ &\geq \underbrace{\big\|\Lambda_0(\boldsymbol{\Sigma}) \cdot (\mathbf{U}^*)^T\widehat{\mathbf{U}}(\mathcal{I})^{\perp}\big\|_2}_{\text{(i)}} - \underbrace{\big\|(\mathbf{U}^*)^T\widehat{\mathbf{U}}(\mathcal{I})^{\perp} \cdot \Lambda_1\big(\widehat{\boldsymbol{\Sigma}}_{\mathcal{I}}\big)\big\|_2}_{\text{(ii)}}. \end{aligned} \tag{D.3}$$

For term (i) in (D.3), note that $\lambda_k > \lambda_{k+1} \geq 0$. Hence we have

$$\begin{aligned} \big\|(\mathbf{U}^*)^T\widehat{\mathbf{U}}(\mathcal{I})^{\perp}\big\|_2 = \big\|\Lambda_0(\boldsymbol{\Sigma})^{-1} \cdot \Lambda_0(\boldsymbol{\Sigma}) \cdot (\mathbf{U}^*)^T\widehat{\mathbf{U}}(\mathcal{I})^{\perp}\big\|_2 &\leq \big\|\Lambda_0(\boldsymbol{\Sigma})^{-1}\big\|_2 \cdot \big\|\Lambda_0(\boldsymbol{\Sigma}) \cdot (\mathbf{U}^*)^T\widehat{\mathbf{U}}(\mathcal{I})^{\perp}\big\|_2 \\ &= \frac{1}{\lambda_k} \cdot \big\|\Lambda_0(\boldsymbol{\Sigma}) \cdot (\mathbf{U}^*)^T\widehat{\mathbf{U}}(\mathcal{I})^{\perp}\big\|_2, \end{aligned}$$

which implies

$$\big\|\Lambda_0(\boldsymbol{\Sigma}) \cdot (\mathbf{U}^*)^T\widehat{\mathbf{U}}(\mathcal{I})^{\perp}\big\|_2 \geq \lambda_k \cdot \big\|(\mathbf{U}^*)^T\widehat{\mathbf{U}}(\mathcal{I})^{\perp}\big\|_2.$$

For term (ii) in (D.3), we have

$$\big\|(\mathbf{U}^*)^T\widehat{\mathbf{U}}(\mathcal{I})^{\perp} \cdot \Lambda_1\big(\widehat{\boldsymbol{\Sigma}}_{\mathcal{I}}\big)\big\|_2 \leq \big\|(\mathbf{U}^*)^T\widehat{\mathbf{U}}(\mathcal{I})^{\perp}\big\|_2 \cdot \big\|\Lambda_1\big(\widehat{\boldsymbol{\Sigma}}_{\mathcal{I}}\big)\big\|_2 = \lambda_{k+1}\big(\widehat{\boldsymbol{\Sigma}}_{\mathcal{I}}\big) \cdot \big\|(\mathbf{U}^*)^T\widehat{\mathbf{U}}(\mathcal{I})^{\perp}\big\|_2.$$

Therefore, from (D.3) we have

$$\begin{aligned} \overbrace{\big\|(\mathbf{U}^*)^T\big(\widehat{\boldsymbol{\Sigma}} - \boldsymbol{\Sigma}\big)\widehat{\mathbf{U}}(\mathcal{I})^{\perp}\big\|_2}^{\text{(i)}} &\geq \lambda_k \cdot \big\|(\mathbf{U}^*)^T\widehat{\mathbf{U}}(\mathcal{I})^{\perp}\big\|_2 - \lambda_{k+1}\big(\widehat{\boldsymbol{\Sigma}}_{\mathcal{I}}\big) \cdot \big\|(\mathbf{U}^*)^T\widehat{\mathbf{U}}(\mathcal{I})^{\perp}\big\|_2 \\ &= \underbrace{\Big[\lambda_k - \lambda_{k+1}\big(\widehat{\boldsymbol{\Sigma}}_{\mathcal{I}}\big)\Big]}_{\text{(ii)}} \cdot \big\|(\mathbf{U}^*)^T\widehat{\mathbf{U}}(\mathcal{I})^{\perp}\big\|_2. \end{aligned} \tag{D.4}$$

For term (i) in (D.4), remind each column of $\widehat{\mathbf{U}}(\mathcal{I})^{\perp}$ is an eigenvector of $\widehat{\boldsymbol{\Sigma}}_{\mathcal{I}}$, and thus its support is $\mathcal{I}$. In other words, $\big[\widehat{\mathbf{U}}(\mathcal{I})^{\perp}\big]_{i,*} = \mathbf{0}$ for all $i \notin \mathcal{I}$. Meanwhile, $\mathbf{U}^*$ is also row-sparse with row-support $\mathcal{S}^* \subseteq \mathcal{I}$. Hence we have

$$(\mathbf{U}^*)^T(\widehat{\boldsymbol{\Sigma}} - \boldsymbol{\Sigma})\widehat{\mathbf{U}}(\mathcal{I})^{\perp} = (\mathbf{U}^*)^T(\widehat{\boldsymbol{\Sigma}} - \boldsymbol{\Sigma})_{\mathcal{I}}\widehat{\mathbf{U}}(\mathcal{I})^{\perp}.$$

Thus we obtain

$$\big\|(\mathbf{U}^*)^T(\widehat{\boldsymbol{\Sigma}} - \boldsymbol{\Sigma})\widehat{\mathbf{U}}(\mathcal{I})^{\perp}\big\|_2 = \big\|(\mathbf{U}^*)^T(\widehat{\boldsymbol{\Sigma}} - \boldsymbol{\Sigma})_{\mathcal{I}}\widehat{\mathbf{U}}(\mathcal{I})^{\perp}\big\|_2 \le \|\mathbf{U}^*\|_2 \cdot \big\|(\widehat{\boldsymbol{\Sigma}} - \boldsymbol{\Sigma})_{\mathcal{I}}\big\|_2 \cdot \big\|\widehat{\mathbf{U}}(\mathcal{I})^{\perp}\big\|_2.$$

Because $\mathbf{U}^*$ and $\widehat{\mathbf{U}}(\mathcal{I})^{\perp}$ are both orthonormal matrices, they satisfy $\|\mathbf{U}^*\|_2 = \big\|\widehat{\mathbf{U}}(\mathcal{I})^{\perp}\big\|_2 = 1$. Thus, we obtain

$$\big\|(\mathbf{U}^*)^T(\widehat{\boldsymbol{\Sigma}} - \boldsymbol{\Sigma})\widehat{\mathbf{U}}(\mathcal{I})^{\perp}\big\|_2 \le \big\|(\widehat{\boldsymbol{\Sigma}} - \boldsymbol{\Sigma})_{\mathcal{I}}\big\|_2.$$

Meanwhile, we have

$$\begin{aligned}\big\|(\widehat{\boldsymbol{\Sigma}} - \boldsymbol{\Sigma})_{\mathcal{I}}\big\|_2 = \max_{\|\mathbf{v}\|_2=1}\big\|(\widehat{\boldsymbol{\Sigma}} - \boldsymbol{\Sigma})_{\mathcal{I}}\mathbf{v}\big\|_2 &= \max_{\substack{\operatorname{supp}(\mathbf{v})\subseteq\mathcal{I}\\ \|\mathbf{v}\|_2=1}}\big\|(\widehat{\boldsymbol{\Sigma}} - \boldsymbol{\Sigma})\mathbf{v}\big\|_2\\ &\le \max_{\substack{|\operatorname{supp}(\mathbf{v})|\le|\mathcal{I}|\\ \|\mathbf{v}\|_2=1}}\big\|(\widehat{\boldsymbol{\Sigma}} - \boldsymbol{\Sigma})\mathbf{v}\big\|_2 = \big\|\widehat{\boldsymbol{\Sigma}} - \boldsymbol{\Sigma}\big\|_{2,|\mathcal{I}|}.\end{aligned} \tag{D.5}$$

Thus, for term (i) in (D.4) we obtain

$$\big\|(\mathbf{U}^*)^T(\widehat{\boldsymbol{\Sigma}} - \boldsymbol{\Sigma})\widehat{\mathbf{U}}(\mathcal{I})^{\perp}\big\|_2 \le \big\|(\widehat{\boldsymbol{\Sigma}} - \boldsymbol{\Sigma})_{\mathcal{I}}\big\|_2 \le \big\|\widehat{\boldsymbol{\Sigma}} - \boldsymbol{\Sigma}\big\|_{2,|\mathcal{I}|}. \tag{D.6}$$

For term (ii) in (D.4), the perturbation theory (Stewart and Sun, 1990) for eigenvalues of symmetric matrices gives

$$\begin{aligned}\lambda_{k+1}(\widehat{\boldsymbol{\Sigma}}_{\mathcal{I}}) \le \lambda_{k+1}(\boldsymbol{\Sigma}_{\mathcal{I}}) + \lambda_1(\widehat{\boldsymbol{\Sigma}}_{\mathcal{I}} - \boldsymbol{\Sigma}_{\mathcal{I}}) &\le \lambda_{k+1}(\boldsymbol{\Sigma}_{\mathcal{I}}) + \big\|(\widehat{\boldsymbol{\Sigma}} - \boldsymbol{\Sigma})_{\mathcal{I}}\big\|_2\\ &\le \lambda_{k+1}(\boldsymbol{\Sigma}_{\mathcal{I}}) + \big\|\widehat{\boldsymbol{\Sigma}} - \boldsymbol{\Sigma}\big\|_{2,|\mathcal{I}|},\end{aligned} \tag{D.7}$$

where the second inequality follows from the fact that

$$\begin{aligned}\lambda_1(\widehat{\boldsymbol{\Sigma}}_{\mathcal{I}} - \boldsymbol{\Sigma}_{\mathcal{I}}) = \max_{\|\mathbf{v}\|_2=1}\mathbf{v}^T(\widehat{\boldsymbol{\Sigma}} - \boldsymbol{\Sigma})_{\mathcal{I}}\mathbf{v} &\le \max_{\|\mathbf{v}_2\|_2=1}\max_{\|\mathbf{v}_1\|_2=1}\mathbf{v}_1^T(\widehat{\boldsymbol{\Sigma}} - \boldsymbol{\Sigma})_{\mathcal{I}}\mathbf{v}_2\\ &= \max_{\|\mathbf{v}_2\|_2=1}\big\|(\widehat{\boldsymbol{\Sigma}} - \boldsymbol{\Sigma})_{\mathcal{I}}\mathbf{v}_2\big\|_2 = \big\|(\widehat{\boldsymbol{\Sigma}} - \boldsymbol{\Sigma})_{\mathcal{I}}\big\|_2.\end{aligned}$$

The last inequality in (D.7) is from (D.5). Thus we have $\lambda_k - \lambda_{k+1}(\widehat{\boldsymbol{\Sigma}}_{\mathcal{I}}) \ge \lambda_k - \lambda_{k+1}(\boldsymbol{\Sigma}_{\mathcal{I}}) - \big\|\widehat{\boldsymbol{\Sigma}} - \boldsymbol{\Sigma}\big\|_{2,|\mathcal{I}|}$. Meanwhile, since each column of $\mathbf{U}^*$, i.e., $\mathbf{U}^*_{*,j}$ $(j = 1, \ldots, k)$, satisfies $\operatorname{supp}(\mathbf{U}^*_{*,j}) \subseteq \mathcal{S}^* \subseteq \mathcal{I}$, we have

$$\boldsymbol{\Sigma}_{\mathcal{I}} \cdot \mathbf{U}^*_{*,j} = \boldsymbol{\Sigma} \cdot \mathbf{U}^*_{*,j} = \lambda_j \cdot \mathbf{U}^*_{*,j},$$

where the last equality uses the fact that $\mathbf{U}^*_{*,j}$ is the eigenvector of $\boldsymbol{\Sigma}$ corresponding to eigenvalue $\lambda_j$. This indicates that $\mathbf{U}^*_{*,j}$ is also an eigenvector of $\boldsymbol{\Sigma}_{\mathcal{I}}$ with eigenvalue

$$\lambda_j(\boldsymbol{\Sigma}_{\mathcal{I}}) = \lambda_j, \qquad j = 1, \ldots, k. \tag{D.8}$$

Since the $(k+1)$-th largest eigenvalue of $\mathbf{\Sigma}_{\mathcal{I}}$ can be written as

$$\lambda_{k+1}(\mathbf{\Sigma}_{\mathcal{I}}) = \max_{\|\mathbf{v}\|_2=1} \mathbf{v}^T \left[ \mathbf{\Sigma}_{\mathcal{I}} - \sum_{j=1}^{k} \lambda_j(\mathbf{\Sigma}_{\mathcal{I}}) \cdot \mathbf{U}^*_{*,j} \cdot (\mathbf{U}^*_{*,j})^T \right] \mathbf{v},$$

we have

$$\lambda_{k+1}(\mathbf{\Sigma}_{\mathcal{I}}) = \max_{\|\mathbf{v}\|_2=1} \mathbf{v}^T \left[ \mathbf{\Sigma}_{\mathcal{I}} - \sum_{j=1}^{k} \lambda_j \cdot \mathbf{U}^*_{*,j} \cdot (\mathbf{U}^*_{*,j})^T \right] \mathbf{v} = \max_{\|\mathbf{v}\|_2=1} \mathbf{v}^T \left[ \mathbf{\Sigma}_{\mathcal{I}} - \mathbf{U}^* \cdot \Lambda_0(\mathbf{\Sigma}) \cdot (\mathbf{U}^*)^T \right] \mathbf{v}.$$

Note that $\mathbf{U}^*$ is row-sparse with row-support $\mathcal{S}^* \subseteq \mathcal{I}$. Hence we have

$$\begin{aligned} \lambda_{k+1}(\mathbf{\Sigma}_{\mathcal{I}}) = \max_{\|\mathbf{v}\|_2=1} \mathbf{v}^T \left[ \mathbf{\Sigma}_{\mathcal{I}} - \mathbf{U}^* \cdot \Lambda_0(\mathbf{\Sigma}) \cdot (\mathbf{U}^*)^T \right] \mathbf{v} &= \max_{\|\mathbf{v}\|_2=1} \mathbf{v}^T \left[ \mathbf{\Sigma}_{\mathcal{I}} - \mathbf{U}^*_{\mathcal{I},\cdot} \cdot \Lambda_0(\mathbf{\Sigma}) \cdot (\mathbf{U}^*_{\mathcal{I},\cdot})^T \right] \mathbf{v} \\ &= \max_{\substack{\mathrm{supp}(\mathbf{v}) \subseteq \mathcal{I} \\ \|\mathbf{v}\|_2=1}} \mathbf{v}^T \left[ \mathbf{\Sigma} - \mathbf{U}^* \cdot \Lambda_0(\mathbf{\Sigma}) \cdot (\mathbf{U}^*)^T \right] \mathbf{v}. \end{aligned}$$

Meanwhile, recall that $\mathbf{U}^*_{*,j}$'s $(j = 1, \ldots, k)$ are the eigenvectors of $\mathbf{\Sigma}$. Hence we have

$$\begin{aligned} \lambda_{k+1} = \max_{\|\mathbf{v}\|_2=1} \mathbf{v}^T \left[ \mathbf{\Sigma} - \sum_{j=1}^{k} \lambda_j \cdot \mathbf{U}^*_{*,j} \cdot (\mathbf{U}^*_{*,j})^T \right] \mathbf{v} &= \max_{\|\mathbf{v}\|_2=1} \mathbf{v}^T \left[ \mathbf{\Sigma} - \mathbf{U}^* \cdot \Lambda_0(\mathbf{\Sigma}) \cdot (\mathbf{U}^*)^T \right] \mathbf{v} \\ &\geq \max_{\substack{\mathrm{supp}(\mathbf{v}) \subseteq \mathcal{I} \\ \|\mathbf{v}\|_2=1}} \mathbf{v}^T \left[ \mathbf{\Sigma} - \mathbf{U}^* \cdot \Lambda_0(\mathbf{\Sigma}) \cdot (\mathbf{U}^*)^T \right] \mathbf{v} \\ &= \lambda_{k+1}(\mathbf{\Sigma}_{\mathcal{I}}). \end{aligned}$$

Thus, we obtain $\lambda_k > \lambda_{k+1} \geq \lambda_{k+1}(\mathbf{\Sigma}_{\mathcal{I}})$, which implies

$$\begin{aligned} \lambda_k - \lambda_{k+1}\big(\widehat{\mathbf{\Sigma}}_{\mathcal{I}}\big) &\geq \lambda_k - \lambda_{k+1}(\mathbf{\Sigma}_{\mathcal{I}}) - \big\|\widehat{\mathbf{\Sigma}} - \mathbf{\Sigma}\big\|_{2,|\mathcal{I}|} \\ &\geq \lambda_k - \lambda_{k+1} - \big\|\widehat{\mathbf{\Sigma}} - \mathbf{\Sigma}\big\|_{2,|\mathcal{I}|}, \end{aligned}$$

where the first inequality follows from (D.7). By Lemma 5.10, we have $\big\|\widehat{\mathbf{\Sigma}} - \mathbf{\Sigma}\big\|_{2,|\mathcal{I}|} \leq \big[\lambda_k - \lambda_{k+1}\big]/2$ with probability at least $1 - 2/d - 1/d^2$ for a sufficiently large $n$. Therefore, for term (ii) in (D.4) we obtain

$$\lambda_k - \lambda_{k+1}\big(\widehat{\mathbf{\Sigma}}_{\mathcal{I}}\big) \geq \frac{\lambda_k - \lambda_{k+1}}{2}. \tag{D.9}$$

Plugging (D.6) and (D.9) into (D.4), we obtain

$$\begin{aligned} \big\|\widehat{\mathbf{\Sigma}} - \mathbf{\Sigma}\big\|_{2,|\mathcal{I}|} \geq \big\|(\mathbf{U}^*)^T \big(\widehat{\mathbf{\Sigma}} - \mathbf{\Sigma}\big) \widehat{\mathbf{U}}(\mathcal{I})^{\perp}\big\|_2 &\geq \Big[\lambda_k - \lambda_{k+1}\big(\widehat{\mathbf{\Sigma}}_{\mathcal{I}}\big)\Big] \cdot \big\|(\mathbf{U}^*)^T \widehat{\mathbf{U}}(\mathcal{I})^{\perp}\big\|_2 \\ &\geq \frac{\lambda_k - \lambda_{k+1}}{2} \cdot \big\|(\mathbf{U}^*)^T \widehat{\mathbf{U}}(\mathcal{I})^{\perp}\big\|_2, \end{aligned}$$

which implies

$$D\big[\mathcal{U}^*, \widehat{\mathcal{U}}(\mathcal{I})\big] = \sqrt{2} \cdot \big\|(\mathbf{U}^*)^T \widehat{\mathbf{U}}(\mathcal{I})^{\perp}\big\|_F \leq \sqrt{2k} \cdot \big\|(\mathbf{U}^*)^T \widehat{\mathbf{U}}(\mathcal{I})^{\perp}\big\|_2 \leq \frac{2\sqrt{2k} \cdot \big\|\widehat{\mathbf{\Sigma}} - \mathbf{\Sigma}\big\|_{2,|\mathcal{I}|}}{\lambda_k - \lambda_{k+1}}$$

since $\lambda_k - \lambda_{k+1} > 0$, where the first equality follows from Lemma 2.1. Plugging in the upper bound of $\big\|\widehat{\mathbf{\Sigma}} - \mathbf{\Sigma}\big\|_{2,|\mathcal{I}|}$ in Lemma 5.10, we conclude the proof. □